\documentclass[]{fairmeta}

\usepackage[english]{babel}
\usepackage{amsmath, amssymb, mathtools, bm, amsthm}
\usepackage{physics}
\usepackage{enumitem}
\usepackage{algorithm}
\usepackage{algpseudocode}
\usepackage{graphicx}
\usepackage{booktabs}
\usepackage{xspace}
\usepackage{adjustbox}
\usepackage{ragged2e}
\usepackage{multirow}
\usepackage{float}
\usepackage{wrapfig}

\newcommand{\multilinecomment}[1]{}

\newcommand{\R}{\mathbb{R}}
\newcommand{\E}{\mathbb{E}}

\newcommand{\Proj}{\operatorname{Proj}}

\newcommand{\blendr}{\textsc{BLenDeR}\xspace}

\newcommand{\inner}[2]{\langle #1, #2 \rangle}

\newcommand{\opprompt}{\textsc{TA}\xspace}
\newcommand{\opembmix}{\textsc{TEI}\xspace}
\newcommand{\opblendrop}{\textsc{RSO}\xspace}
\newcommand{\opblendr}{$\blendr$\xspace}

\title{\blendr: Blended Text Embeddings and Diffusion Residuals for Intra-Class Image Synthesis in Deep Metric Learning}

\author[ 2,3,*]{Jan Niklas Kolf}
\author[1]{Ozan Tezcan}
\author[1]{Justin Theiss}
\author[1]{Hyung Jun Kim}
\author[1]{Wentao Bao}
\author[1]{Bhargav Bhushanam}
\author[1]{Khushi Gupta}
\author[1]{Arun Kejariwal}
\author[2,3]{Naser Damer}
\author[2]{Fadi Boutros}

\affiliation[1]{Meta Reality Labs}
\affiliation[2]{Fraunhofer IGD}
\affiliation[3]{Technical University of Darmstadt}

\contribution[*]{Work done while interning at Meta Reality Labs}

\abstract{The rise of Deep Generative Models (DGM) has enabled the generation of high-quality synthetic data. When used to augment authentic data in Deep Metric Learning (DML), these synthetic samples enhance intra-class diversity and improve the performance of downstream DML tasks.
We introduce BLenDeR, a diffusion sampling method designed to increase intra-class diversity for DML in a controllable way by leveraging set-theory inspired union and intersection operations on denoising residuals.
The union operation encourages any attribute present across multiple prompts, while the intersection extracts the common direction through a principal component surrogate.
These operations enable controlled synthesis of diverse attribute combinations within each class, addressing key limitations of existing generative approaches.
Experiments on standard DML benchmarks demonstrate that BLenDeR consistently outperforms state-of-the-art baselines across multiple datasets and backbones.
Specifically, BLenDeR achieves $3.7\%$ increase in Recall@1 on CUB-200 and a $1.8\%$ increase on Cars-196, compared to state-of-the-art baselines under standard experimental settings.}

\date{\today}
\correspondence{\email{jan.niklas.kolf@igd.fraunhofer.de}, \email{ozantezcan@meta.com}}
\metadata[Code]{\url{https://github.com/facebookresearch/meta_ai_blender}}

\begin{document}

\maketitle

\begin{figure}[htbp]
    \centering
    \includegraphics[width=0.5\linewidth]{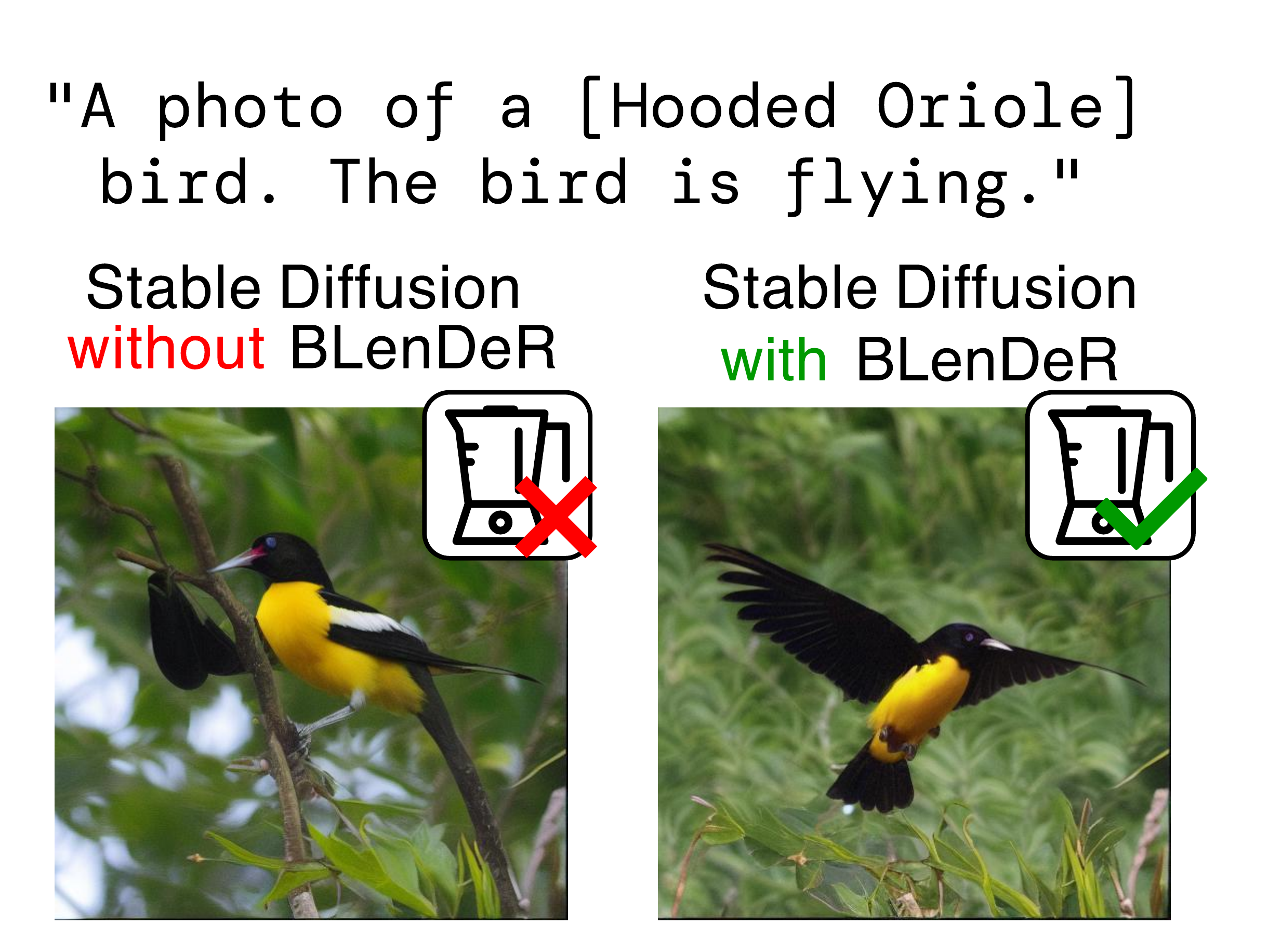}
    \caption{
        Motivated by the observation that Stable Diffusion personalized with LoRA and Textual Inversion struggles to generate learned concepts with novel attributes, e.g., $[\texttt{Hooded Oriole}]$ with the attribute \texttt{flying}, we propose \blendr, a novel diffusion sampling method that steers the personalized model to generate target concepts with novel and challenging target attributes.
    }
    \label{fig:blendr_teaser}
\end{figure}
\section{Introduction}
Deep Metric Learning (DML) learns an embedding function that maps input samples into a feature space where semantically similar samples are closer together than dissimilar ones according to a chosen distance metric \cite{potential_fields, DBLP:conf/eccv/MusgraveBL20, DBLP:conf/icml/RothMSGOC20}.
The goal is to cluster samples from the same class while enforcing margins to separate different classes, which powers image retrieval \cite{DBLP:conf/iccv/Movshovitz-Attias17}, re-identification \cite{DBLP:journals/corr/HermansBL17, DBLP:conf/cvpr/ChenCZH17}, and open-world scenarios \cite{DBLP:conf/eccv/MusgraveBL20}.
The final performance of a DML model does not only depend on the choice of the used loss function that governs the learning of the embedding space, but also on the availability of diverse and informative training samples that capture intra- and inter-class variations \cite{DBLP:conf/icml/RothMSGOC20, DBLP:conf/iccv/BoutrosGKD23}.
Collecting such a breadth of diverse data, e.g., diverse poses and backgrounds, is costly.

Deep Generative Models (DGM) can help to reduce the impact of these issues \cite{wang_cvpr2024_diffmix, wang2025inversion}.
The rapid advance in DGM, especially in text-to-image (T2I) and image-to-image (I2I) diffusion models, allows controlled synthesis of novel variations using natural language prompts \cite{DBLP:conf/iclr/SongME21, DBLP:conf/nips/HoJA20, DBLP:conf/nips/SongE19, DBLP:conf/icml/Sohl-DicksteinW15, DBLP:conf/cvpr/RombachBLEO22}.
After training a DGM on specific concepts, it can be used to add synthetic samples to authentic datasets at scale \cite{DBLP:conf/iclr/GalAAPBCC23}.
The cost and burden of generating synthetic data are relatively small compared to those of collecting and annotating authentic data.

A key limitation of DGM is their tendency to internalize spurious correlations between target concepts and their respective attributes \cite{DBLP:journals/natmi/GeirhosJMZBBW20}.
For example, if during DGM training, the model encounters a particular bird (e.g., an Albatross) alongside specific backgrounds (e.g., ocean background), it may continue to generate that bird with those backgrounds, even when prompted with different attribute descriptions (e.g., forest backgrounds) \cite{DBLP:conf/mm/YuanCWQYS24}.
Generating concepts with novel attributes while maintaining the core characteristics of the concept is crucial for DML, as it increases the intra-class diversity. This diversity weakens the links between a concept and its attributes in the dataset.
Those links might be exploited by a downstream DML model, which might impact its performance when the intra-class diversity in the dataset is limited \cite{hse_pa_iccv23}.
Prior work shows that diffusion models can yield challenging classification examples via class or background mixing, often with soft labels to manage noisy labels \cite{wang_cvpr2024_diffmix, dong2025sgd}.
However, DML poses stricter demands because its training signals are sensitive to noisy labels and distribution shifts between authentic and synthetic data \cite{DBLP:conf/eccv/MusgraveBL20}.
While previous work maintains class characteristics by pasting segmented foregrounds into novel backgrounds \cite{dong2025sgd}, the resulting images consist of unnatural pose characteristics.

We introduce \textbf{\blendr}, a sampling method designed to increase intra-class diversity by coupling two complementary controls in the sampling phase of T2I diffusion models.
First, we interpolate text embeddings during denoising to steer hidden states toward desired attributes.
Second, we compose denoising residuals computed for prompts sharing a latent through set-theory style union and intersection  operations, which consistently amplify or suppress attribute directions across timesteps. The motivation of \blendr is shown in Figure \ref{fig:blendr_teaser}.

\blendr is tailored to deep metric learning, where text embedding interpolation adds initial attribute directions. Residual unions inject new attribute directions that may be underrepresented in the training data throughout the denoising process, while residual intersections extract directions common across prompts. Together, these operations allow \blendr to synthesize a wide range of intra-class variations, which strengthen the training signal and improve metric learning performance.



\paragraph{Contributions:}
\begin{itemize}
    \item We introduce residual set operations, a method grounded in set-theoretic principles. \blendr composes denoising residuals using union and intersection operations, and provides a mathematical framework for these operations within diffusion models.
    \item We demonstrate that \blendr can synthesize images to significantly increase intra-class diversity in a controllable way. This allows specification of attributes such as pose or background, enabling more robust deep metric learning (DML).
    \item We show empirical evidence that synthetic data generated by \blendr consistently improves recall metrics over state-of-the-art DML methods on CUB-200-2011 and Cars-196 datasets, evaluated with multiple backbone architectures. Specifically, using a ResNet-50 backbone, Recall@1 on CUB-200 increased by $3.7\%$, and on Cars-196 by $1.8\%$.
\end{itemize}
\section{Related Work}
\label{sec:2_related_work}

\paragraph{Deep Metric Learning:}
Early DML methods utilized tuple based objectives such as contrastive \cite{chopra2005learning} and triplet loss \cite{facenet}.
Mining strategies are used to focus learning on informative pairs \cite{xuan_easy, deepranking, harwood2017smart, Manmatha2017SamplingMI, Yuan2016HardAwareDC}.
Subsequent works propose structured objectives that reduce the dependence on hard mining while preserving strong separation \cite{Sohn2016ImprovedDM, multisimilarity_dml}.
As the amount of triplets explode with increased dataset sizes, proxy based losses became popular.
Instead of focusing on pair to pair relations, these losses focus on learnable embeddings called proxies, that represent the class centers \cite{probab_proxy, proxy_anchor, proxy_gml, proxy_ncapp}.
Hyperbolic-based metrics have also been considered as alternatives to Euclidean distance metrics \cite{ermolov2022hyperbolic, kim2023hier}.
Recent work adopts potential fields and models attraction and repulsion forces across proxies and samples of same and different classes \cite{potential_fields}.
Hybrid Species Embedding (HSE) \cite{hse_pa_iccv23} uses augmentations between images in the batch in conjunction with an auxiliary loss to create hard samples.
HSE's CutMix operations suffer from two limitations: visible artifacts with discontinuities in background, pose, and semantics at cut boundaries, and the concept (class) remaining anchored to its original attributes due to per-class image sampling.
\blendr overcomes these issues by synthesizing the concept with novel attributes in coherent scenes that exhibit the attribute, e.g., natural pose and background variations, while maintaining semantic stability.

\paragraph{Generative Augmentation for Inter- and Intra-Class Diversity:}
Recent work adapts deep generative models to synthesize training data for classification.
I2I models add noise to randomly selected dataset samples, which are then denoised to a target class, producing hard intra- and inter-class samples with noisy labels \cite{wang_cvpr2024_diffmix}.
Saliency-guided mixing combined with I2I models improves label clarity and background diversity \cite{dong2025sgd}.
DDIM inversion with inversion circle interpolation augments images within a class for dataset diversity \cite{wang2025inversion}.
However, these approaches have the following limitations: I2I methods rely on a strength parameter to balance novel attribute consistency against class consistency, often producing images that closely resemble existing dataset samples.
The saliency-guided approach pastes objects into backgrounds rather than synthesizing them cohesively.
DDIM inversion methods offer limited control over attribute generation.
\blendr addresses these limitations by synthesizing learned concepts with novel attributes in a controlled manner, generating images that maintain coherent appearance between concept and attribute.

\paragraph{Diffusion Models and Controllable Attribute Composition:}
Text conditioned latent diffusion enables fast synthesis of high resolution images and has become the default backbone for personalization and editing \cite{DBLP:conf/iclr/SongME21, DBLP:conf/nips/HoJA20, DBLP:conf/nips/SongE19, DBLP:conf/icml/Sohl-DicksteinW15, DBLP:conf/cvpr/RombachBLEO22}.
Classifier-free guidance improves alignment but can limit diversity at large scale \cite{DBLP:journals/corr/abs-2207-12598}.
Composable diffusion models combine multiple text conditions by product of experts to form images that satisfy several prompts \cite{DBLP:conf/eccv/LiuLDTT22}.
While these methods can be conditioned using text, attribute control is limited when the base model struggles with generating a target concept. \blendr builds on these foundations, but acts directly in the model output space at inference time, composing residuals from several prompts that share a latent to inject or suppress attribute directions during denoising.

\begin{figure*}[htbp]
    \centering
    \begin{adjustbox}{max width=\textwidth}
        \includegraphics{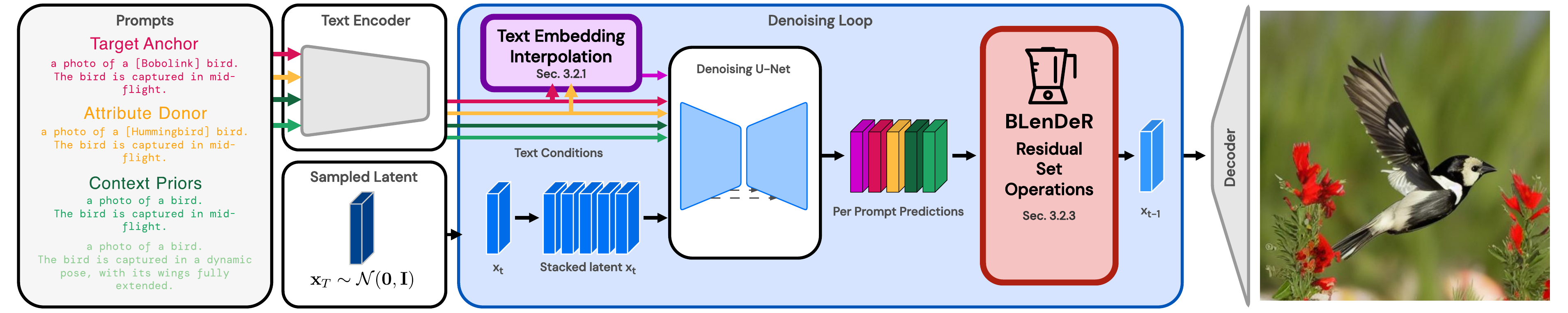}
    \end{adjustbox}
    \caption{Overview of the proposed \blendr approach. \blendr uses multiple text prompts: a target anchor prompt, an attribute donor prompt, and multiple context prior prompts.
    The attribute donor prompt is used with Text Embedding Interpolation to pre-align latents with the target attribute.
    \blendr utilizes the noise predictions from each text embedding in the proposed residual set operations, which use context priors to steer the denoising process toward the target attribute and concept specified in the target anchor prompt.}
    \label{fig:blendr_operations}
\end{figure*}

\section{Methodology}
\label{sec:methodology}

\blendr's core novelty is residual set operations (\opblendrop) that utilize context prior prompts to guide denoising trajectories toward target attributes.
Additionally, \blendr leverages text embedding interpolation (\opembmix) as a method to pre-align latents with target attributes.
Together, these operations enhance intra-class diversity while preserving class consistency.
An overview of our proposed \blendr framework is presented in Figure \ref{fig:blendr_operations}.
This Figure presents T2I latent diffusion (Section \ref{subsec:preliminaries}), \opembmix (Section \ref{subsubsec:methodology_textembinterp}),  and the proposed \blendr \opblendrop (Section \ref{subsec:rso}).

\subsection{Preliminaries}
\label{subsec:preliminaries}
\paragraph{T2I Latent Diffusion Models:}
Let $x_0 \in \R^D$ denote the image latent representation of a pre-trained autoencoder.
A forward diffusion process is defined by a Markov chain $q(x_t \mid x_{t-1})=\mathcal{N}(\sqrt{\alpha_t}x_{t-1}, (1-\alpha_t) I)$ over $T$ steps \cite{lai2025principles}.
The reverse process seeks a model $p_\theta(x_{t-1}\mid x_t)$ that progressively denoises $x_t$ back to a sample $\hat x_0$.
The denoising diffusion probabilistic model trains a noise predictor $\varepsilon_\theta(x_t,t,E(c))$ that minimizes
\begin{equation}
\mathcal{L}_{\varepsilon} = \E\Big[\norm{\varepsilon - \varepsilon_\theta(x_t,t,E(c))}_2^2\Big],
\label{eq:epsloss}
\end{equation}
where $E(c)\in \R^{S}$ is a condition, e.g. a text embedding of a prompt $c$ produced by a frozen text encoder \cite{lai2025principles}. 
$E(c)$ denotes optional conditioning through a text prompt.
During inference, the process begins with a latent variable $x_T$ sampled from a Gaussian distribution. At each step, $\varepsilon_\theta(x_t,t,E(c))$ predicts the noise present in $x_t$, which is then subtracted iteratively \cite{DBLP:conf/nips/HoJA20}.
This continues until a final latent $x_0$ is produced, which is expected to exhibit the attribute encoded by $E(c)$.

\paragraph{T2I Personalization:}
Diffusion model personalization enables generating images of specific concepts, such as classes in a dataset. Textual Inversion (TI) \cite{DBLP:conf/iclr/GalAAPBCC23} learns a unique text embedding for a target concept from example images, associating it with a chosen phrase $[V_i]$. Prompting the model with $[V_i]$ generates images of the personalized concept. We refer to $[V_i]$ as the target concept, representing class $i$ in the dataset along with its encoded attributes.

\paragraph{T2I with Attribute Annotations:}
Input prompts $c$ follow the template ``a photo of a $[V_i]$ \texttt{[metaclass]}. $a .$``, where $[V_i]$ is the target concept, \texttt{[metaclass]} specifies the object type (e.g., \texttt{bird} or \texttt{car}), and $a$ denotes a target attribute description.

\paragraph{Text Conditioning and Classifier-Free Guidance:}
Classifier-Free Guidance (CFG) \cite{DBLP:journals/corr/abs-2207-12598} evaluates the model on two inputs that share the same latent but differ in conditioning.
Denote by $c$ a conditioned prompt and by $\varnothing$ the empty prompt.
With a noise predictor the two outputs are
\begin{equation}
\varepsilon_{\mathrm{cond}} = \varepsilon_\theta(x_t,t,E(c)),
\varepsilon_{\varnothing} = \varepsilon_\theta(x_t,t,E(\varnothing)).
\end{equation}
The guided residual is
\begin{equation}
r_{\mathrm{cfg}} := \varepsilon_{\mathrm{cond}} - \varepsilon_{\varnothing},
\end{equation}
and the guidance adjusted output is
\begin{equation}
\widehat{\varepsilon} = \varepsilon_{\varnothing} + w_{\mathrm{cfg}}\, r_{\mathrm{cfg}},
\label{eq:cfg}
\end{equation}
with a scale $w_{\mathrm{cfg}}\ge 0$. CFG increases alignment with the prompt at the cost of reduced diversity \cite{DBLP:journals/corr/abs-2207-12598}.

\subsection{\blendr: Embedding Interpolation and Residual Set Operations}
\label{subsec:prompts}
We consider a latent sample $x_t\in \R^{C\times H\times W}$ at a time step $t$, a collection of $n$ prompts $\{c_i\}_{i=1}^n$, their text embeddings $h_i=E(c_i)$, and $h_{\varnothing}=E(\varnothing)$ the text embedding of the empty prompt $\varnothing$.
We instantiate the prompts as follows:
\begin{itemize}[leftmargin=1.5em]
\item $c_1$ (target anchor): a prompt that \emph{always} contains the target concept $[V_i]$ together with the \emph{novel} target attribute phrase $a$ we wish to imprint.
\item $c_2$ (attribute donor): a prompt that uses a \emph{different but related} concept $[V_j]$ that is known to co-occur with the same target attribute phrase $a$. This donor stabilizes the attribute direction in the model output space.
\item $c_3,\ldots,c_n$ (context priors): prompts that use either $[V_i]$, other concepts $[V_m]$ or only the \texttt{[metaclass]} name (such as \texttt{bird} or \texttt{car}) paired with attribute phrases $a'$ that semantically relate to the target attribute $a$ (e.g., paraphrases or similar attributes). These context prompts establish a semantic subspace around the target attribute, which is leveraged in residual set operations to guide generation toward the desired attribute while providing robustness to variations in attribute phrasing.
\end{itemize}

\subsubsection{Text Embedding Interpolation}
\label{subsubsec:methodology_textembinterp}

Text embedding interpolation biases the early latent toward target attribute $a$ while preserving concept $[V_i]$. A short early schedule adds the attribute donor $[V_j]$ from prompt $c_2$, then decays to anchor $[V_i]$ from prompt $c_1$ by cut time $t_\star$.
At each step we choose weights $\alpha_i(t)\ge 0$ with $\sum_{i=1}^n \alpha_i(t)=1$ and define
\begin{equation}
h_{\mathrm{mix}}(t) = \sum_{i=1}^n \alpha_i(t)\, h_i.
\label{eq:mixembed}
\end{equation}

Concretely, we set $\alpha_1(t)=1-\gamma(t)$, $\alpha_2(t)=\gamma(t)$, $\alpha_{k>2}(t)=0$, with a cosine ramp $\gamma(t)$ that starts at $1.0$ and decays to $0.0$ by $t_\star$.
This is considered as an \emph{early donor injection}.

\subsubsection{Latent Stacking}
\label{subsubsec:methodology_latents}

\blendr stacks a single latent $x_t$ $n+2$ times and evaluates the U-Net on the following hidden states
\begin{equation}
\{h_{\varnothing},\; h_{\mathrm{mix}},\; h_1,\ldots,h_n\}.
\label{eq:stack}
\end{equation}
From the U-Net we obtain predicted noise per hidden state
\begin{align}
\varepsilon_{\varnothing} &= \varepsilon_{\theta}(x_t,t,h_{\varnothing}), \nonumber\\
\varepsilon_{\mathrm{mix}} &= \varepsilon_{\theta}(x_t,t,h_{\mathrm{mix}}), \nonumber\\
\varepsilon_i &= \varepsilon_{\theta}(x_t,t,h_i)\quad i=1,\ldots,n, \label{eq:stackouts}
\end{align}
where the guidance residual is defined as $r_{\mathrm{cfg}}=\varepsilon_{\mathrm{mix}}-\varepsilon_{\varnothing}$.
We follow a similar approach to CFG and define per-prompt residuals relative to $\varepsilon_{\mathrm{mix}}$
\begin{equation}
r_i := \varepsilon_i - \varepsilon_{\mathrm{mix}}, \qquad i=1,\ldots,n,
\label{eq:perclassres}
\end{equation}
which represents the semantic offset in prediction space to achieve the respective prompt target.

\subsubsection{Residual Space Composition by Set Operations}
\label{subsec:rso}

Residuals $r$ live in a high dimensional Euclidean space $\R^{C\cdot H\cdot W} = \R^{D}, D=C \cdot H \cdot W$. We flatten them when useful. We now define two operations that mirror set union and intersection.
The operators are designed to be stable during denoising where the scale of $\varepsilon_\theta$ varies with $t$.

\paragraph{Union:} The union residual encourages any attribute present in at least one of the prompts
\begin{equation}
R_{\cup} := \sum_{r_i\in \mathcal{I}_{\cup}} r_i / (\norm{r_i}_2 + \delta),
\label{eq:union}
\end{equation}
where $\mathcal{I}_{\cup}\subset\{r_1,\ldots, r_n\}$ collects residuals of prompts that inject the attribute we want to add.
Normalization prevents a single large residual from dominating.
Parameter $\delta > 0$ is a small constant for numerical stability.
The operation is robust to phrasing because we can average across several paraphrases. 

\paragraph{Intersection:} The intersection seeks common directions across a set of residuals $r_i \in \mathcal{I}_{\cap}\subset\{r_1,\ldots,r_n\}$.
Let $M\in\R^{|\mathcal{I}_{\cap}|\times D}$ stack the flattened residuals.
The first principal component (PC-1) $v_1 = \text{argmax}_{||v||_{2}=1}||Mv||^{2}_{2}$ is recovered efficiently using singular value decomposition, selecting the top eigenpair. With mean residual $\mu = \frac{1}{|\mathcal{I}_{\cap}|} \sum_{r' \in \mathcal{I}_{\cap}} r'$ we define
\begin{equation}
R_{\cap} := \text{Proj}_{\text{span}(v_1)}(\mu) = \inner{\mu}{v_1} v_1.
\label{eq:inter}
\end{equation}
The PC-1 is the unit direction where the projection of the used residuals have maximal variance.
It ensures we capture how strongly the residuals collectively push in that direction.

\multilinecomment{
\paragraph{Difference.} Let $k$ index a target residual and $\mathcal{B} \in \R^{|\mathcal{I}_{\setminus}| \times D}$ index a set of basis residuals $r_i \in \mathcal{I}_{\setminus}\subset\{r_1,\ldots,r_n\}$ that we wish to remove, such as common backgrounds.
With $B = \begin{bmatrix} r_b \end{bmatrix}_{b \in \mathcal{B}}$ the orthogonal projector onto $\text{span}(B)$ is $\text{P}_{B} = B^T(BB^T + \lambda I)^{-1}B$ with a small regularization $\lambda > 0$ for numerical stability.
We define the difference operations as the orthogonal complement of the projected part of $r_k$ onto $\text{P}_B$:
\begin{equation}
R_{\setminus} := r_k - \text{P}_{B}r_k.
\label{eq:diff}
\end{equation}
This removes the components of $r_k$ explained by the basis and keeps the orthogonal part, which isolates features unique to the target attribute.
If $r_k$ can be explained by $\text{span}(B)$, then $R_{\setminus}=0$, signaling that it has no unique attribute.
For metric learning this constructs samples that are informative for a specific decision boundary by keeping only the discriminative cues.
}

\subsubsection{The \blendr Denoiser}

Using time varying weight functions $\beta(t)$, the combined residual is
\begin{equation}
R_{\mathrm{\blendr}} = \beta_{\cup}(t)\, R_{\cup} + \beta_{\cap}(t)\, R_{\cap}. 
\label{eq:Radd}
\end{equation}
We optionally remove the component parallel to $r_{\mathrm{cfg}}$:
\begin{equation}
R_{\mathrm{\blendr}} \leftarrow R_{\mathrm{\blendr}} - \frac{\inner{R_{\mathrm{\blendr}}}{r_{\mathrm{cfg}}}}{\norm{r_{\mathrm{cfg}}}^2} r_{\mathrm{cfg}},
\label{eq:orth}
\end{equation}
which preserves directions not already enforced by guidance and prevents over steering in directions of the guidance.
We finally clamp its norm relative to the guidance norm
\begin{equation}
\norm{R_{\mathrm{\blendr}}} \le \tau \norm{r_{\mathrm{cfg}}},
\label{eq:clamp}
\end{equation}
with a pre-defined norm scale $\tau$.
Let $w_{\text{cfg}}(t)$ be the guidance schedule.
\blendr forms the adjusted model output as
\begin{equation}
\widehat{\varepsilon}(t) = \varepsilon_{\varnothing} + w_{\mathrm{cfg}}(t)\, r_{\mathrm{cfg}} + R_{\mathrm{\blendr}}.
\label{eq:epshat}
\end{equation}


\newcommand{\groupw}{0.48\textwidth}     
\newcommand{\innerw}{0.24\linewidth}     
\newcommand{\groupgap}{1.0em}            
\newcommand{\rowgap}{1.0\baselineskip}   

\newcommand{\imgcell}[2]{%
  \begin{minipage}[t]{\innerw}%
    \vspace{0pt}\centering
    \includegraphics[width=\linewidth]{#1}%
    {\captionsetup{skip=2pt,belowskip=0pt}\caption*{#2}}%
  \end{minipage}%
}

\begin{figure*}[t]
  \centering
  \captionsetup[subfigure]{labelformat=simple,labelsep=space,font=small}

  \begin{subfigure}[t]{\groupw}
    \vspace{0pt}\centering
    \imgcell{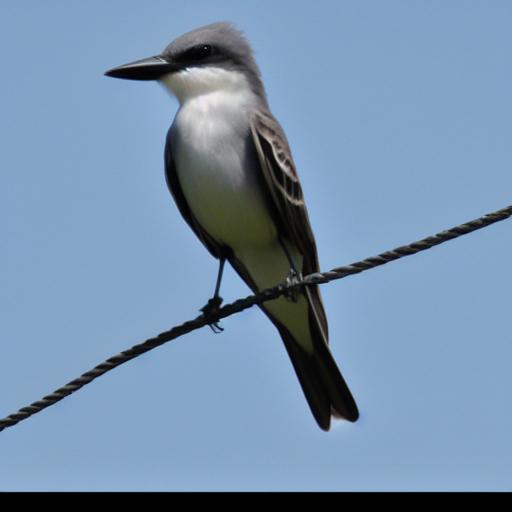}{\opprompt}\hfill
    \imgcell{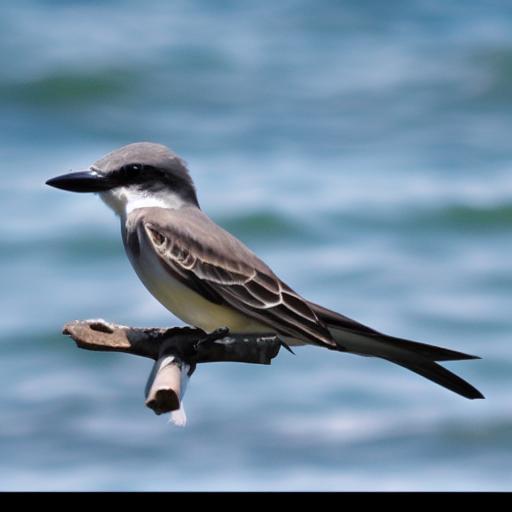}{\opembmix }\hfill
    \imgcell{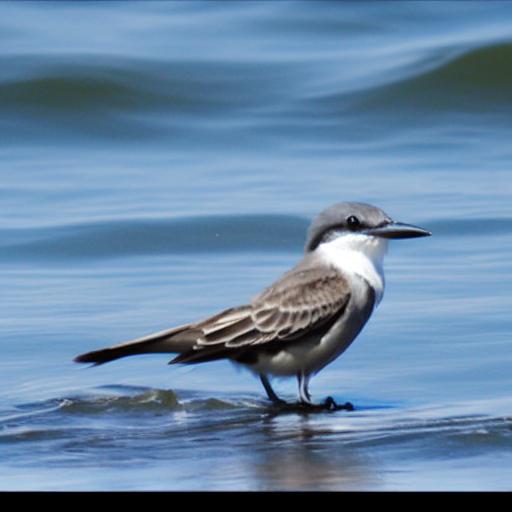}{\opblendrop ($\cup$)}\hfill
    \imgcell{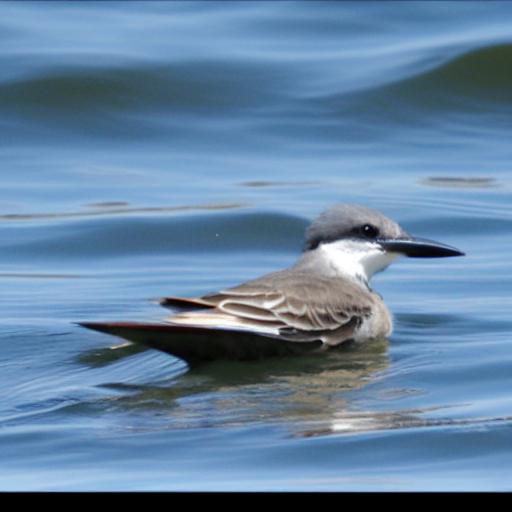}{\opblendr}
    \caption{\footnotesize Target attribute $a$: \texttt{The background of the image features a vast expanse of deep blue ocean, which stretches out to the horizon.}}\label{fig:blendr:g1}
  \end{subfigure}\hspace{\groupgap}
  \begin{subfigure}[t]{\groupw}
    \vspace{0pt}\centering
    \imgcell{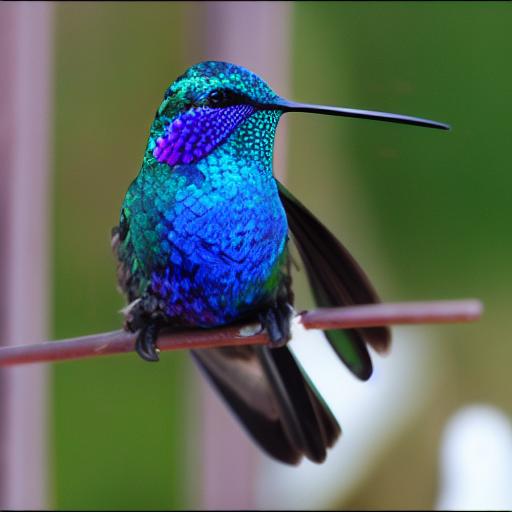}{\opprompt}\hfill
    \imgcell{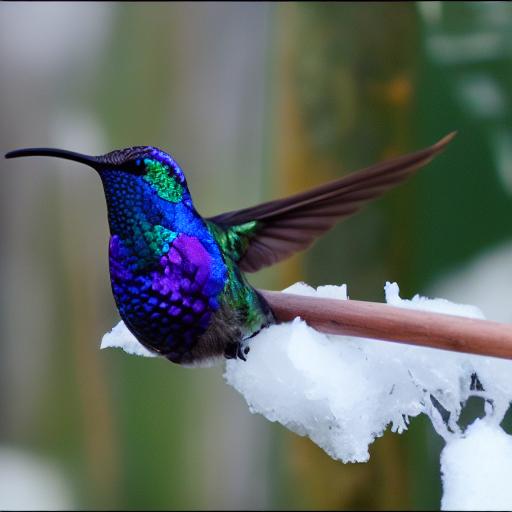}{\opembmix }\hfill
    \imgcell{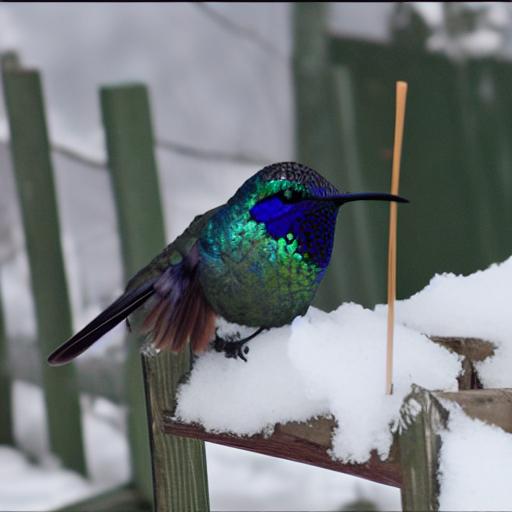}{\opblendrop ($\cup$)}\hfill
    \imgcell{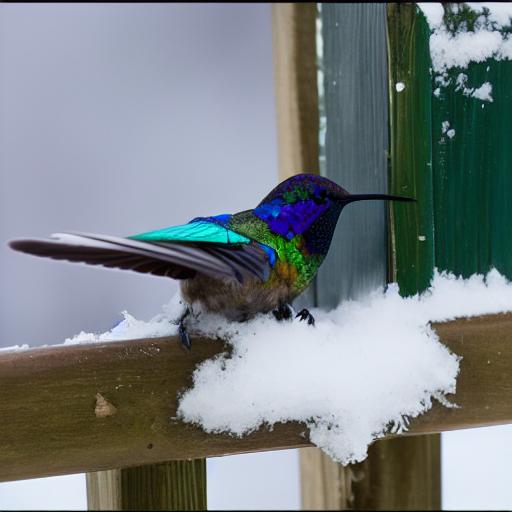}{\opblendr}
    \caption{\footnotesize Target attribute $a$: \texttt{The background of the image features a close-up view of a wooden fence partially covered with snow.}}\label{fig:blendr:g2}
  \end{subfigure}

  \par\vspace{\rowgap}

  \begin{subfigure}[t]{\groupw}
    \vspace{0pt}\centering
    \imgcell{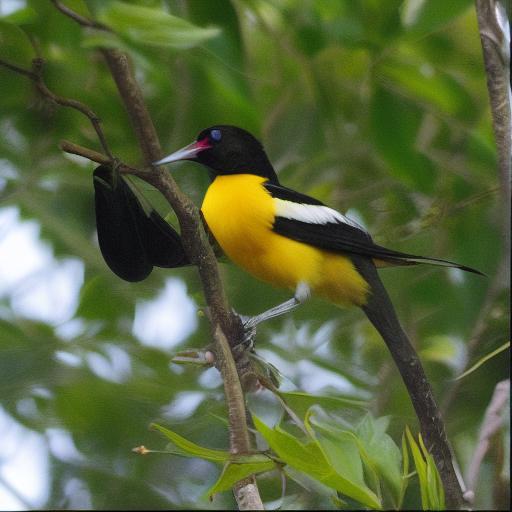}{\opprompt}\hfill
    \imgcell{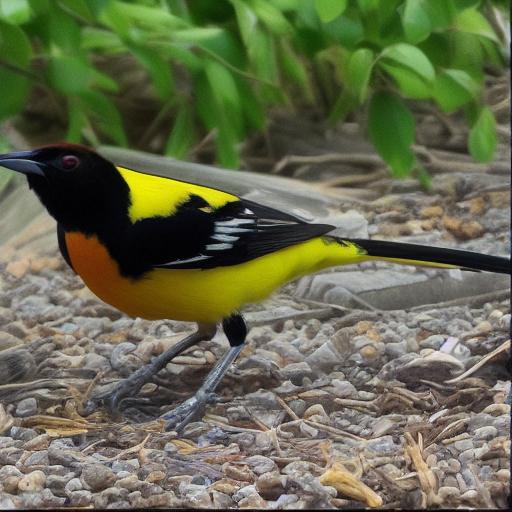}{\opembmix}\hfill
    \imgcell{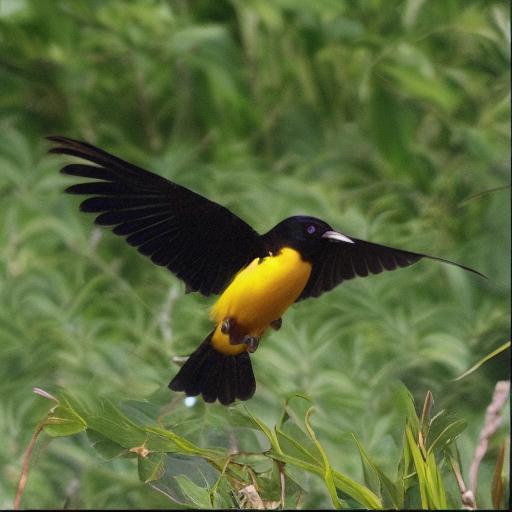}{\opblendrop ($\cap$)}\hfill
    \imgcell{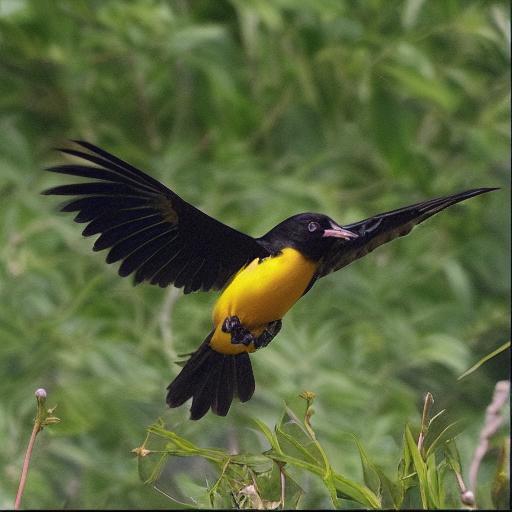}{\opblendr}
    \caption{\footnotesize Target attribute $a$: \texttt{The bird is captured in a dynamic pose, with its wings fully extended, showcasing its impressive wingspan.}}\label{fig:blendr:g3}
  \end{subfigure}\hspace{\groupgap}
  \begin{subfigure}[t]{\groupw}
    \vspace{0pt}\centering
    \imgcell{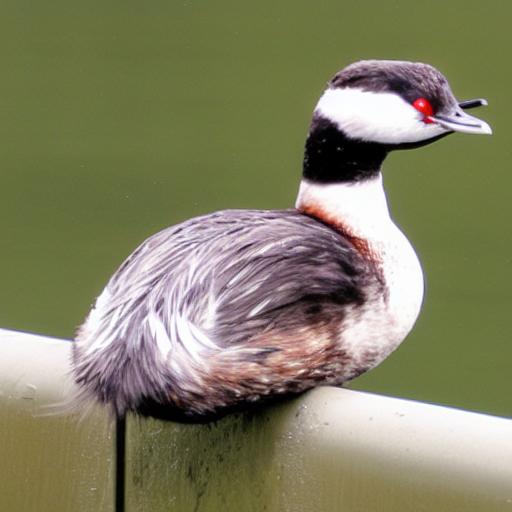}{\opprompt}\hfill
    \imgcell{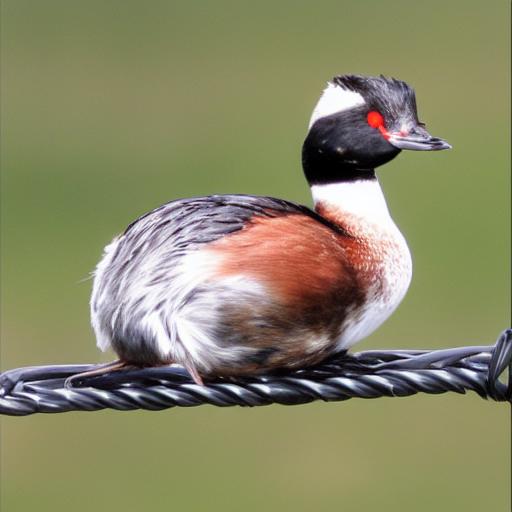}{\opembmix}\hfill
    \imgcell{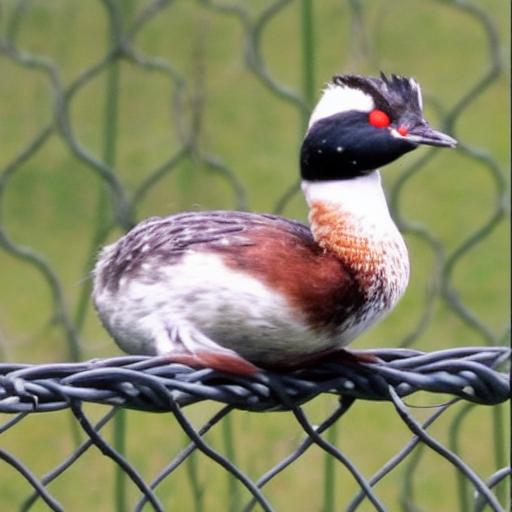}{\opblendrop ($\cap$)}\hfill
    \imgcell{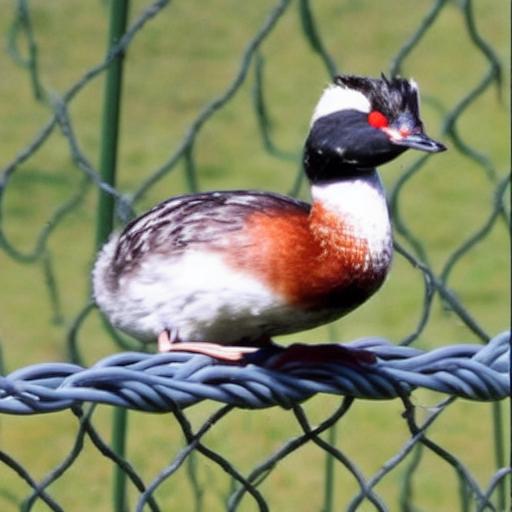}{\opblendr}
    \caption{\footnotesize Target attribute $a$: \texttt{The bird is perched on a barbed wire fence.}}\label{fig:blendr:g4}
  \end{subfigure}

  \caption{Visual demonstration when using different approaches for generating novel background and poses.
  Prompts used in generation are consisting of two parts, the invoking part that contains the target concept $[V_i]$, ``\texttt{A photo of a $[V_i]$ bird.}``, followed by the target attribute description $a$.
  For generation, either only the Target Anchor prompt $c_1$ is used as Baseline (\opprompt, Sec. \ref{subsec:prompts}), or $c_1$ with Text Embedding Interpolation with attribute donor prompt $c_2$ (\opembmix, Eq. \ref{eq:mixembed}), or $c_1$ with \blendr Residual Space Operation Union ($\cup$) or Intersection ($\cap$) (\opblendrop, Eq. \ref{eq:union} and Eq. \ref{eq:inter}), or full \blendr operation which combines \opembmix and \opblendrop.
  As can be seen, when the base model is not able to generate the concept with the targeted attribute, \blendr is able to synthesize the concept in conjunction with the targeted attribute.}
  \label{fig:blendr_grouped_rows}
\end{figure*}

\section{Experimental Setup}
\label{sec:experimental_setup}

This section presents the datasets used in this work, \blendr training and sampling, as well as DML protocols.
Additional implementation notes are deferred to the Appendix.

\textbf{Datasets:}
Experiments cover CUB-200-2011 \cite{welinder2010caltech} (CUB; 11{,}788 images, 200 bird species), and Cars-196 \cite{krause20133d} (Cars; 16{,}185 images, 196 car models).
Common DML train/test splits are adopted from \cite{DBLP:conf/icml/RothMSGOC20}.

\subsection{\blendr Training}
\label{subsec:blendr_training}
\quad \textbf{DGM Backbone:}
We use Stable Diffusion~1.5 \cite{DBLP:conf/cvpr/RombachBLEO22} with a DDPM scheduler \cite{DBLP:conf/nips/HoJA20}, following \cite{wang_cvpr2024_diffmix}.
LoRa \cite{DBLP:conf/iclr/HuSWALWWC22} is used as parameter-efficient fine tuning method.
One LoRA+TI model is trained per object category (birds, cars) to match the statistics of each dataset \cite{wang_cvpr2024_diffmix}.

\textbf{Data Preprocessing:}
Image descriptions are extracted with LLaVa-Next \cite{liu_arxiv2024_llavanext}, which is prompted to produce sentences that isolate foreground, background, pose, or camera viewpoint attributes (Appendix~\ref{supp:image_descriptions}).
Foreground masks follow the language-guided Segment Anything procedure of prior work (Appendix~\ref{supp:image_alignment}).

\textbf{Model Training:}
Every class receives a dedicated TI token $[V_i]$ \cite{wang_cvpr2024_diffmix, wang2025inversion}.
LoRAs \cite{DBLP:conf/iclr/HuSWALWWC22} of rank $r=10$ are inserted into the U-Net attention and linear layers while the base weights stay frozen.
The Text Encoder stays frozen and only the TI tokens are optimized \cite{wang_cvpr2024_diffmix, wang2025inversion}.
Training uses AdamW with learning rate $5\times 10^{-5}$, batch size 8, $512\times512$ image size, and 20k steps per category.
Objects are centered in the crop with a random border to regularize spatial context (Appendix~\ref{supp:image_alignment}).

\textbf{Prompt Template:}
Each training prompt follows ``a photo of a $[V_i]$ \texttt{[metaclass]}. $a$.`` where $a$ is a description from LLaVa-Next (Appendix~\ref{supp:training_prompts}).
This template instantiates concept $[V_i]$ with attribute $a$, whose text embeddings condition the denoiser.

\subsection{\blendr Generation}
\label{subsec:blendr_generation}
\blendr augments each class $[V_i]$ with challenging intra-class samples that introduce novel attribute combinations while keeping the personalized concept intact.

\textbf{Preprocessing:}
All attribute descriptions are encoded with CLIP to compute similarity rankings per image attribute (Appendix~\ref{supp:clip_description_ranking}).
Class similarity rankings are derived from embeddings of a ProxyAnchor-pretrained ResNet-50 \cite{proxy_anchor}, enabling retrieval of donors with matching semantics to target anchor.

\textbf{Attribute Selection:}
For a given concept $[V_i]$, a representative attribute $a_{[V_i]}$ is chosen from its descriptions.
A novel target attribute $a_{\text{new}_1}$ is sampled from the $50\%$ most dissimilar descriptions in the CLIP ranking.
This $a_{\text{new}_1}$ serves as the novel attribute target for $[V_i]$.
Robustness to paraphrasing is achieved by adding the four nearest CLIP neighbors of $a_{\text{new}_1}$, yielding $\{a_{\text{new}_j}\}_{j=1}^{5}$.
For each $a_{\text{new}_j}$ the class $[V_{a_{\text{new}_j}}]$ closest to $[V_i]$ that naturally co-occurs with the attribute is selected, providing donors with similar semantics to $[V_i]$ in case they are used in an operation.

\textbf{Text Embedding Interpolation (\opembmix):}
Generation uses the target anchor prompt $c_1$ with $[V_i]$ and $a_{\text{new}_1}$;  and attribute donor prompt $c_2$ with $[V_{a_{\text{new}_1}}]$ and $a_{\text{new}_1}$ inside Eq.~\eqref{eq:mixembed}.
Weights start with $\alpha_2(0)=\gamma(0)=1$ to bias the early denoising trajectory toward the donor attribute.
A cosine ramp drives $\gamma(t)$ to zero by $t_\star=0.2$, so that $\alpha_2(t_\star)=0$ and the mixture reverts to the target anchor for the remaining $80\%$ of steps (Appendix~\ref{supp:embedding_space_interpolation}).

\textbf{Target Attributes:}
Novel backgrounds are synthesized across CUB, and Cars.
Additional CUB generations target bird poses (e.g., flying, swimming), while Cars additionally target camera angles representing viewpoints.

\textbf{\blendr Operations:}
All operations utilize target anchor and attribute donor prompts together with \opembmix (Eq. \ref{eq:mixembed}).
Both union and intersection operations combine target anchor and attribute donor with context prior prompts $c_{j+2}$ ($\texttt{[metaclass]}, a_{\text{new}_j}$) for $j \in {1,\dots, 5}$ that omit class-specific concepts $[V_{a_{\text{new}_j}}]$ to avoid concept leakage.
Weight schedules $\beta(t)$ for union and intersection operation are using $\beta(0)$ randomly sampled from $\{3.0, 4.0, 5.0, 6.0\}$, with a cosine ramp down to $0.0$ at step $t=0.8$.
Classifier-Free Guidance of $4.0$ is used.
$R_{\mathrm{\blendr}}$ is orthogonalized using Eq. \ref{eq:orth}.
A norm clamp of $\tau=3$ is used (Eq. \ref{eq:clamp}).
See Appendix~\ref{supp:operation_parameters} for detailed explanations and experiments.

\textbf{Data Generation:}
CUB and Cars each yield 150 samples per class and attribute.

\subsection{DML Application}
\label{subsec:exp_setup_dml_training}
We demonstrate the effectiveness of data generated by \blendr on augmenting the training dataset of downstream DML task.

\textbf{DML Backbones:}
ImageNet \cite{deng2009imagenet} pretrained ResNet-50 \cite{DBLP:conf/cvpr/HeZRS16}, ViT (variant ViT-S)   \cite{dosovitskiy2021an} and DINO \cite{caron2021emerging} models serve as backbones.
Their final output layers are replaced with linear layers mapping to the embedding dimension (Appendix~\ref{supp:dml_model_training}).
During training, images are resized to $256\times256$ and random-cropped to $224\times224$, following \cite{potential_fields, proxy_anchor}.

\textbf{Model Training:}
SOTA losses, including Potential Field (PF) \cite{potential_fields} and ProxyAnchor (PA) \cite{proxy_anchor} are used with their respective hyperparameters.
We follow PF training setup and use class-balanced batches: PF uses batch size 100, PA uses 180, each with 10 images per class \cite{potential_fields}.
As PF \cite{potential_fields} does not provide in-depth parameter choices for DINO and ViT, we follow \cite{kim2023hier} for hyperparameter selection.
Synthetic data is mixed by randomly sampling $2$, $6$, or $10$ generated images per class from the synthetic datasets and inserting them into each batch, yielding a synthetic to real (S2R) ratio of $0.2$, $0.6$, and $1$, respectively.
Optimizer settings follow PF/PA defaults with learning rate $5\times10^{-4}$ on the backbone and a $100\times$ larger rate on the proxy parameters.
More details about training parameters are provided in Appendix \ref{supp:dml_model_training}

\textbf{Evaluation:}
Recall@$K$ is measured on the respective test split using images resized to $256\times256$, followed by a $224\times224$ center crop. For CUB and Cars, $K \in \{1,2,4\}$ is used.

\begin{table}
\centering
\begin{adjustbox}{width=0.65\linewidth}
\begin{tabular}{l ccc}
\toprule
\textbf{Dataset / Target Attribute / Subset} & \textbf{\opembmix} & \textbf{\opblendrop} & \textbf{\opblendr} \\
\midrule
\multicolumn{4}{l}{\textbf{Cars-196 / Pose / Intersection}} \\
Full           & +0.6\%  & +2.7\%  & +3.0\%  \\
Bottom 50\%    & +3.6\%  & +7.5\%  & +8.3\%  \\
Bottom 20\%    & +6.2\%  & +12.4\% & +13.6\% \\
Bottom 5\%     & +10.3\% & +19.8\% & +22.1\% \\
\midrule
\multicolumn{4}{l}{\textbf{Cars-196 / Background / Union}} \\
Full           & +0.3\%  & +9.4\%  & +10.0\%  \\
Bottom 50\%    & +4.3\%  & +21.2\% & +22.1\%  \\
Bottom 20\%    & +10.1\% & +35.8\% & +37.8\%  \\
Bottom 5\%     & +21.8\% & +62.4\% & +72.8\%  \\
\midrule
\multicolumn{4}{l}{\textbf{CUB-2011-200 / Pose / Intersection}} \\
Full           & +1.9\%  & +5.0\%  & +5.6\%  \\
Bottom 50\%    & +5.2\%  & +9.9\%  & +10.9\% \\
Bottom 20\%    & +8.2\%  & +15.3\% & +16.8\% \\
Bottom 5\%     & +13.7\% & +24.5\% & +27.2\% \\
\midrule
\multicolumn{4}{l}{\textbf{CUB-2011-200 / Background / Union}} \\
Full           & +2.5\%  & +9.2\%  & +10.2\%  \\
Bottom 50\%    & +6.4\%  & +18.4\% & +20.5\%  \\
Bottom 20\%    & +13.7\% & +32.9\% & +37.3\%  \\
Bottom 10\%    & +19.6\% & +46.8\% & +51.7\%  \\
Bottom 5\%     & +29.7\% & +63.8\% & +69.0\%  \\
\bottomrule
\end{tabular}
\end{adjustbox}
\caption{
CLIP cosine similarity improvement [\%] between target attribute descriptions and generated images for \blendr variants vs. baseline, shown for all samples ("Full") and challenging cases ("Bottom X\%") where baseline similarity is lowest, demonstrating \blendr's ability to enhance generation in those challenging cases. 
See Supplementary Material \ref{supp:clip_score_evaluation} for detailed results.
}
\label{tab:clip_improvements}
\end{table}
\section{Results}
\label{sec:results}
\subsection{Qualitative Results}
We qualitatively assess how \blendr is able to generate a target concept $[V_i]$ with a novel target attribute $a$ (Fig. \ref{fig:blendr_grouped_rows}).
Each of the sub figures (a-d) shows the generated images using different generation approaches, using base model with target anchor prompt.


The baseline model, which uses only the target anchor prompt (\opprompt), produces high-fidelity images of the input class. However, it often fails to incorporate target attributes that are uncommon for the input class. When applying text embedding interpolation (\opembmix), the denoising trajectory is pre-aligned with the desired attribute, resulting that the target attribute is appearing in the scene to a certain degree (Fig. \ref{fig:blendr_grouped_rows} a,b,d).
Using only the residual set operations (\opblendrop), the target attribute is generated to a stronger degree than using only \opembmix, but impacts the overall composition, e.g. how natural the pose is (Fig. \ref{fig:blendr_grouped_rows} a, bird standing on the ocean).
Combining both \opembmix and \opblendrop (\opblendr) produces images that contain the target attribute with higher fidelity, while maintaining the characteristic appearance of the concept. This improvement is further supported by quantitative results in the next section.

\begin{table}

\centering

\begin{adjustbox}{width=0.65\linewidth}

\begin{tabular}{l c l ccc ccc}

\toprule

  \multirow{2}{*}{

  \begin{tabular}{@{}c@{}}

    \textbf{Backbone} \\

    \textbf{Loss Function}

  \end{tabular} } & \textbf{Operation} & \textbf{S2R} & \multicolumn{3}{c}{\textbf{CUB-200-2011}} & \multicolumn{3}{c}{\textbf{Cars-196}} \\

 & & & R@1 & R@2 & R@4 & R@1 & R@2 & R@4 \\

\midrule
\midrule


\multirow{7}{*}{

  \begin{tabular}{@{}c@{}}

    ResNet50 (512 dim) \\

    ProxyAnchor

  \end{tabular}

} 

 & Base & 0 & 72.7 & 82.1 & 88.7 & 90.5 & 94.5 & 96.8 \\
\cmidrule(lr){2-9}
 & \multirow{3}{*}{

  \begin{tabular}{@{}c@{}}

    Background \\

    (Union)

  \end{tabular}

} & $0.2$ & 74.0 & 83.7 & \textbf{89.9} & 91.9 & 95.4 & 97.3 \\

 & & $0.6$ & 74.1 & \textbf{83.7} & 89.7 & \textbf{92.3} & \textbf{95.6} & \textbf{97.3} \\

 & & $1.0$ & \textbf{74.6} & 83.7 & 89.7 & 92.1 & 95.6 & 97.3 \\
\cmidrule(lr){2-9}
  & \multirow{3}{*}{

  \begin{tabular}{@{}c@{}}

    Pose \\

    (Intersection)

  \end{tabular}

} & $0.2$ & 73.7 & 82.6 & 89.3 & 90.9 & 95.0 & 97.0 \\

 & & $0.6$ & 73.2 & 82.5 & 89.0 & 90.7 & 94.8 & 96.9 \\

 & & $1.0$ & 72.5 & 81.9 & 88.7 & 90.1 & 94.5 & 96.7 \\

\midrule
\midrule


\multirow{7}{*}{

  \begin{tabular}{@{}c@{}}

    ResNet50 (512 dim) \\

    Potential Field

  \end{tabular}

} 

& Base & 0 & 73.3 & 83.0 & 89.0 & 90.2 & 94.3 & 96.9 \\
\cmidrule(lr){2-9}

 & \multirow{3}{*}{

  \begin{tabular}{@{}c@{}}

    Background \\

    (Union)

  \end{tabular}

} & $0.2$ & 74.8 & 83.7 & 89.8 & 91.3 & 95.3 & 97.4 \\

 & & $0.6$ & 76.3 & 85.2 & 90.6 & \textbf{91.9} & 95.6 & \textbf{97.8} \\

 & & $1.0$ & 75.9 & 84.9 & 90.3 & 91.7 & \textbf{95.9} & 97.6 \\
\cmidrule(lr){2-9}
 & \multirow{3}{*}{

  \begin{tabular}{@{}c@{}}

    Pose \\

    (Intersection)

  \end{tabular}

} & $0.2$ & 75.4 & 84.3 & 90.4 & 91.5 & 95.2 & 97.3 \\

 & & $0.6$ & \textbf{77.0} & \textbf{85.7} & \textbf{91.1} & 91.2 & 95.2 & 97.3 \\

 & & $1.0$ & 76.6 & 85.1 & 90.9 & 90.6 & 94.7 & 97.0 \\

\midrule
\midrule


\multirow{7}{*}{

  \begin{tabular}{@{}c@{}}

    ViT (384 dim) \\

    ProxyAnchor

  \end{tabular}

} 

  & Base & 0 & 84.1 & 90.1 & 94.3 & 87.2 & 92.6 & 96.0 \\
\cmidrule(lr){2-9}
  & \multirow{3}{*}{

  \begin{tabular}{@{}c@{}}

    Background \\

    (Union)

  \end{tabular}

} & $0.2$ & 84.2 & 90.7 & 94.3 & 88.2 & 93.3 & 96.2 \\

 & & $0.6$ & 84.2 & 90.4 & 94.4 & 88.4 & \textbf{93.7} & \textbf{96.4} \\

 & & $1.0$ & 84.3 & \textbf{90.9} & \textbf{94.5} & \textbf{88.4} & 93.6 & \textbf{96.4} \\
\cmidrule(lr){2-9}
 & \multirow{3}{*}{

  \begin{tabular}{@{}c@{}}

    Pose \\

    (Intersection)

  \end{tabular}

} & $0.2$ & 84.2 & 90.6 & 94.3 & 87.7 & 93.1 & 96.1 \\

 & & $0.6$ & \textbf{84.4} & 90.7 & 94.2 & 87.7 & 93.2 & 96.3 \\

 & & $1.0$ & 84.1 & 90.8 & 94.5 & 87.9 & 93.2 & 96.2 \\

\midrule
\midrule


\multirow{7}{*}{

  \begin{tabular}{@{}c@{}}

    ViT (384 dim) \\

    Potential Field

  \end{tabular}

} 

  & Base & 0 & 83.1 & 90.0 & 94.0 & 86.3 & 92.4 & 96.1 \\
\cmidrule(lr){2-9}
  & \multirow{3}{*}{

  \begin{tabular}{@{}c@{}}

    Background \\

    (Union)

  \end{tabular}

} & $0.2$ & 83.6 & 90.4 & 93.9 & 86.8 & 92.8 & 96.2 \\

 & & $0.6$ & \textbf{84.0} & \textbf{90.5} & \textbf{94.1} & \textbf{87.7} & \textbf{93.3} & 96.4 \\

 & & $1.0$ & 83.7 & 90.0 & 94.0 & 87.4 & 93.2 & \textbf{96.4} \\
\cmidrule(lr){2-9}
 & \multirow{3}{*}{

  \begin{tabular}{@{}c@{}}

    Pose \\

    (Intersection)

  \end{tabular}

} & $0.2$ & 83.6 & 90.4 & 94.0 & 87.2 & 93.1 & 96.3 \\

 & & $0.6$ & 83.9 & 90.3 & 94.0 & 86.8 & 92.7 & 96.2 \\

 & & $1.0$ & 83.7 & 90.3 & 93.9 & 86.7 & 92.7 & 96.0 \\

\bottomrule

\end{tabular}

\end{adjustbox}

\caption{Comparison of the Recall@$K$ (\%) achieved by our \blendr novel attribute datasets on the CUB and Cars datasets, using Potential Field and ProxyAnchor loss functions and using different synthetic to real (S2R) image ratio. A S2R of $0.2$ means that $2$ synthetic images per $10$ authentic image per class are added to the batch.
Boldfaced values indicate the highest metric within each subcolumn, corresponding to a specific combination of dataset, metric, loss function, and backbone.}

\label{tab:blendr_attribute_performance}

\end{table}
\subsection{Quantitative Results}
\label{subsec:results_quantitative_results}
We quantify target attribute adherence using CLIP similarity between target attribute description and images generated using same seed under the following settings: prompt only baseline (target concept $[V_i]$ with a novel target attribute $a$, labeled \opprompt), \opembmix, \opblendrop, and \opblendr.
The cosine similarity is calculated between CLIP embeddings of target attribute description and the respective image extracted using CLIP text and image encoders respectively.
A higher similarity score indicates better target attribute adherence in generated data.
Table \ref{tab:clip_improvements} shows the improvement (in percentage) of using \opembmix, \opblendrop, and \opblendr over using only prompt for image generation.
Across both benchmarks (CUB and Cars), \blendr consistently outperforms the baseline model, particularly in challenging scenarios where the baseline model struggles to generate target attribute with the target concept.
For samples with the lowest baseline similarity (e.g. bottom $20\%$ and $5\%$), \blendr achieves even better results with gains up to $69\%$.
These results indicate that \blendr is especially effective when the base model fails (e.g., extreme case of Bottom $5\%$) to render the target attribute.

We also evaluate the downstream DML performance of the \blendr generated datasets on ResNet-50 and ViT using ProxyAnchor (PA) and Potential Field (PF) loss, using their reported hyperparameter and under PF training settings (Sec. \ref{subsec:exp_setup_dml_training}).
The results are shown in Table \ref{tab:blendr_attribute_performance}.
Our replicated PA outperforms the published results of PA under PF settings (Tbl. \ref{tab:blendr_attribute_performance}).
We vary the ratio of synthetic images per authentic images in the batch by $0.2, 0.6,$ and $1.0$.

Both versions of \blendr (Union and Intersection) significantly outperform the baseline models across both datasets and a variety of backbone and loss function combinations. Results in Table~\ref{tab:blendr_attribute_performance} show that the Union operation generally results in better DML performance compared to Intersection. We attribute this to the importance of background invariance in DML tasks. Increasing S2R from $0.2$ to $0.6$ consistently improves Recall performance in most cases; however, further increasing it to $1.0$ can sometimes reduce performance. This suggests that the optimal real-to-synthetic ratio for \blendr-based training is $1:0.6$.



\subsection{Comparison to SOTA}

To put our results in broader perspective,  we present in Table \ref{tab:eval_cub_cars_sota} the achieved results by recent SOTA approaches and our best performing \blendr setup. 
As mentioned in Section \ref{subsec:results_quantitative_results}, we replicated PA and PF under PF settings, labeled with * in Table \ref{tab:eval_cub_cars_sota}.
While PA achieved improved performance across all benchmarks with ResNet-50, replicated PF achieved near similar performance with PF on CUB, but lower values on Cars.
Difference on other backbones might be due to different parameter choices not reported in \cite{potential_fields}.

In comparison to PA and PF and under the exact training setups, \blendr outperforms the baselines across all benchmarks and backbones for PA and most configurations for PF, which was configured using parameters from the original paper.

\blendr demonstrates improved performance compared to HSE \cite{hse_pa_iccv23}, which uses CutMix-based image augmentations and an auxiliary loss on top of PA to improve DML performance.
HSE with PA loss achieved $70.6\%$ on CUB, while \blendr with PA achieved $74.6\%$.
This $4$ percentage point improvement indicates that \blendr-generated images provide more challenging training samples while maintaining semantic and visual coherence across background and pose variations, unlike CutMix augmentations which introduce artificial discontinuities and mismatched backgrounds due to different source images.

Compared to PA, PF receives a smaller performance boost from synthetic data augmentation. This reduced improvement may be attributed to PF's piecewise potential formulation with hard margin constraints, which creates tighter decision boundaries and exhibits higher sensitivity to the distributional gap between synthetic and authentic samples.


\begin{table}
\centering
\begin{adjustbox}{width=0.8\linewidth}
\begin{tabular}{lcccccc}
\toprule
\multicolumn{1}{c}{Methods} & \multicolumn{6}{c}{Benchmarks} \\ 
\cmidrule(lr){1-1}\cmidrule(lr){2-7}
 & \multicolumn{3}{c}{\textbf{CUB-200-2011}} & \multicolumn{3}{c}{\textbf{Cars-196}} \\ 
\cmidrule(lr){2-4}\cmidrule(lr){5-7}
 & R@1 & R@2 & R@4 & R@1 & R@2 & R@4 \\ 
\midrule
\midrule
\textbf{ResNet50 (512 dim)} &  &  &  &  &  &  \\
ESPHN~\cite{xuan_easy} & 64.9 & 75.3 & 83.5 & 82.7 & 89.3 & 93.0 \\
N.Softmax~\cite{norm_softmax} & 61.3 & 73.9 & 83.5 & 84.2 & 90.4 & 94.4 \\
DiVA~\cite{diva_dml} & 69.2 & 79.3 & - & 87.6 & 92.9 & - \\
Proxy NCA++~\cite{proxy_ncapp} & 64.7 & - & - & - & 85.1 & - \\
Proxy Anchor~\cite{proxy_anchor} (PA) & 69.7 & 80.0 & 87.0 & 87.7 & 92.9 & 95.8 \\
DCML-MDW~\cite{DCML} & 68.4 & 77.9 & 86.1 & 85.2 & 91.8 & 96.0 \\
MS+DAS~\cite{das_eccv22} & 69.2 & 79.2 & 87.1 & 87.8 & 93.1 & 96.0 \\
HIST~\cite{lim2022hypergraph} & 71.4 & 81.1 & 88.1 & 89.6 & 93.9 & 96.4 \\
HIER\cite{kim2023hier} & 70.1 & 79.4 & 86.9 & 88.2 & 93.0 & 95.6 \\
HSE-PA~\cite{hse_pa_iccv23} & 70.6 & 80.1 & 87.1 & 89.6 & 93.8 & 96.0 \\
Potential Field$^\dag$ \cite{potential_fields} (PF) & 73.4 & 82.4 & 88.8 & 92.7 & 95.5 & 97.6 \\
\midrule
Proxy Anchor$^*$ (PA) & 72.7 & 82.1 & 88.7 & 90.5 & 94.5 & 96.8 \\
\blendr (Ours) - PA & 74.6 & 83.7 & 89.7 & \textbf{92.3} & \textbf{95.6} & 97.3 \\
Potential Field$^*$ (PF) & 73.3 & 83.0 & 89.0 & 90.2 & 94.3 & 96.9 \\
\blendr (Ours) - PF & \textbf{77.0} & \textbf{85.7} & \textbf{91.1} & 91.9 & 95.6 & \textbf{97.7} \\ 
\midrule
\midrule
\textbf{DINO (384 dim)} &  &  &  &  &  &  \\
DINO \cite{caron2021emerging} & 70.8 & 81.1 & 88.8 & 42.9 & 53.9 & 64.2 \\
Hyp \cite{ermolov2022hyperbolic} & 80.9 & 87.6 & 92.4 & 89.2 & 94.1 & 96.7 \\
HIER \cite{kim2023hier} & 81.1 & 88.2 & 93.3 & 91.3 & 95.2 & 97.1 \\
Potential Field$^\dag$ \cite{potential_fields} (PF)& 83.1 & 89.3 & 94.2 & 94.7 & 96.5 & 97.8 \\
\midrule
Proxy Anchor$^*$ (PA) & 81.2 & 88.8 & \textbf{93.1} & 92.1 & 95.6 & 97.6 \\
\blendr (Ours) - PA & \textbf{81.8} & \textbf{88.9} & 93.0 & \textbf{93.0} & \textbf{96.2} & \textbf{97.8} \\
Potential Field$^*$ (PF) & 80.9 & 88.2 & 92.7 & 90.8 & 95.0 & 97.3 \\
\blendr (Ours) - PF & 81.4 & 88.1 & 92.8 & 91.5 & 95.6 & 97.6 \\
\midrule
\midrule
\textbf{ViT (384 dim)} &  &  &  &  &  &  \\
ViT-S \cite{dosovitskiy2021an} & 83.1 & 90.4 & 94.4 & 47.8 & 60.2 & 72.2 \\
Hyp \cite{ermolov2022hyperbolic} & 85.6 & 91.4 & 94.8 & 86.5 & 92.1 & 95.3 \\
HIER \cite{kim2023hier} & 85.7 & 91.3 & 94.4 & 88.3 & 93.2 & 96.1 \\
Potential Field$^\dag$ \cite{potential_fields} (PF) & 87.8 & 92.6 & 95.7 & 91.5 & 95.2 & 97.4 \\
\midrule
Proxy Anchor$^*$ (PA) & 84.1 & 90.1 & \textbf{94.3} & 87.2 & 92.6 & 96.0 \\
\blendr (Ours) - PA & \textbf{84.4} & \textbf{90.7} & 94.2 & \textbf{88.4} & \textbf{93.6} & 96.4 \\
Potential Field$^*$ (PF) & 83.1 & 90.0 & 94.0 & 86.3 & 92.4 & 96.1 \\
\blendr (Ours) - PF & 84.0 & 90.5 & 94.1 & 87.7 & 93.3 & \textbf{96.4} \\
\midrule
\bottomrule
\end{tabular}
\end{adjustbox}
\caption{Comparison on Recall@K on CUB and Cars dataset. 
Potential Field results marked with $^\dag$ are as reported in their original paper but could not be replicated by us despite following the published training settings. 
Results with $^*$ are replicated using training settings of Potential Field \cite{potential_fields} reported in their paper (See Section \ref{sec:experimental_setup}). Boldface indicates the best result among methods with reproducible results. Result of our \blendr are reported on top of Potential Field and Proxy Anchor, noted as \blendr - PA and \blendr - PF, respectively.}
\label{tab:eval_cub_cars_sota}
\end{table}

\section{Limitations}
\label{sec:limitations}

While the target attribute adherence improves with \blendr (Tab. \ref{tab:clip_improvements}), adding the target attribute can not be fully isolated, as \blendr is not an image inpainting method.
For example, targeting a specific background description can also impact the pose of the concept (Fig. \ref{fig:blendr:g3}).
Additionally, \blendr demonstrates the capabilities to generate a concept together with novel target attributes, but this, theoretically, might impact the class characteristics to a certain degree.
\section{Conclusion}
\label{sec:conclusion}

This paper introduced \blendr, a novel diffusion sampling method that generates task aligned synthetic images to improve intra-class diversity of Deep Metric Learning training datasets.
Using two complementary controls, text embedding interpolation and residual composition using Union and Intersection, \blendr is able to generate learned concepts with novel, targeted attributes.
Across a diverse set of benchmarks, backbones and evaluation setups, \blendr demonstrated that it improves adherence to targeted attributes compared to prompt only text to image generation. 
We also showed that \blendr is effective in improving the performance of DML models significantly.

\clearpage
\newpage
\bibliographystyle{assets/plainnat}
\bibliography{paper}

\clearpage
\newpage
\beginappendix

%
%
This Appendix complements the main paper by providing detailed explanations, additional experimental results, and comprehensive implementation details:

\begin{itemize}
    \item Section \ref{subsec:blendr_training} and Section \ref{subsec:blendr_generation} of the main paper explained the Stable Diffusion model training and generation. Appendix Section \ref{supp:image_descriptions} provides detailed explanation of how attribute annotations for the datasets were obtained using LLaVa-Next, including the specific prompts used and example annotations alongside images.
    \item Section \ref{subsec:blendr_generation} introduced how novel attribute descriptions as attribute targets were obtained through CLIP-based similarity rankings. Appendix Section \ref{supp:clip_description_ranking} explains in detail the hard negative mining strategy, including how attribute descriptions are ranked in terms of nearest neighbors and furthest, and how the top-4 most similar descriptions form semantic pairs with the selected attribute annotation for robust generation.
    \item Section \ref{subsec:blendr_training} briefly mentioned that training images were aligned. Appendix Section \ref{supp:image_alignment} provides comprehensive explanation of how images are foreground-aligned using foreground masks obtained via language-guided Segment Anything, including the context parameter and padding strategies.
    \item Section \ref{subsec:blendr_training} of the main paper referred to the usage of training prompts. Appendix Section \ref{supp:training_prompts} outlines in detail the creation of input training prompts, including the template structure, metaclass specifications, and the random context sampling strategy used during preprocessing.
    \item Section \ref{subsubsec:methodology_textembinterp} introduced Text Embedding Interpolation, whose usage was described in Section \ref{subsec:blendr_generation} in the main paper. Appendix Section \ref{supp:embedding_space_interpolation} further explains the mathematical formulation of Text Embedding Interpolation and the principles of the cosine decay scheduling function used to transition from donor to anchor prompts. Additionally, ablation results are shown used to select the Text Embedding Interpolation parameters.
    \item Section~\ref{subsec:prompts} introduced the \blendr approach with Residual Set Operations (RSO) (union and intersection), alongside orthogonalization and norm clamping mechanisms for controlling the trade-off between target attribute adherence and target class preservation. Appendix Section~\ref{supp:operation_parameters} provides a comprehensive ablation study on RSO weights and clamping parameters, demonstrating their effectiveness and offering practical guidelines for parameter selection.
    \item Section \ref{subsec:exp_setup_dml_training} outlined the general parameters for DML training. Appendix Section \ref{supp:dml_model_training} provides comprehensive hyperparameter tables for all backbone architectures (ResNet50, ViT, DINO) across datasets (CUB, Cars) and loss functions (Proxy Anchor, Potential Field).
    \item Section \ref{subsec:results_quantitative_results} of the main paper presented CLIP similarity improvements as relative percentages. Appendix Sections \ref{supp:clip_score_evaluation} and \ref{supp:clip_image_evaluation} provide expanded quantitative analysis including absolute CLIP similarity scores and CLIP image-based diversity metrics.
    \item The main paper presented visual comparisons of generation strategies. Appendix Section \ref{supp:example_images_per_ablation} provides additional visual examples across different generation types (prompt-only, \opembmix, \opblendrop, and \opblendr) for both background and pose attributes, while Appendix Section \ref{supp:example_images_blendr_training_datasets} displays randomly sampled images from the complete synthetic training datasets.
    \item Additionally, a single denoising step is provided as an algorithmic overview in Algorithm \ref{alg:blendr}.
\end{itemize}

\begin{algorithm}[H]
\caption{One denoising step of \blendr at time $t$}
\label{alg:blendr}
\begin{algorithmic}[1]
\Require latent $x_t$, prompts $\{c_i\}_{i=1}^n$, weights $\alpha_i(t)$, residual schedules $\beta_{\cup}(t),\beta_{\cap}(t)$, clamp $\tau$, guidance $w_{\mathrm{cfg}}(t)$
\State $h_i \gets E(c_i)$, $h_{\varnothing}\gets E(\varnothing)$, $h_{\mathrm{mix}}\gets \sum_i \alpha_i(t) h_i$
\State $\varepsilon_{\varnothing} \gets \varepsilon_{\theta}(x_t,t,h_{\varnothing})$
\State $\varepsilon_{\mathrm{mix}} \gets \varepsilon_{\theta}(x_t,t,h_{\mathrm{mix}})$
\State $\varepsilon_i \gets \varepsilon_{\theta}(x_t,t,h_i)\quad i=1,\ldots,n$ 
\State $r_{\mathrm{cfg}}\gets \varepsilon_{\mathrm{mix}}-\varepsilon_{\varnothing}$, $r_i\gets \varepsilon_i-\varepsilon_{\mathrm{mix}}$
\State Build $R_{\cup}$ by \eqref{eq:union}, $R_{\cap}$ by \eqref{eq:inter}
\State $R_{\blendr}\gets \beta_{\cup}(t) R_{\cup} + \beta_{\cap}(t) R_{\cap}$
\State Optional orthogonalize: $R_{\blendr}\gets R_{\blendr} - \Proj_{r_{\mathrm{cfg}}}(R_{\blendr})$
\State Clamp by \eqref{eq:clamp}
\State $\widehat{\varepsilon}\gets \varepsilon_{\varnothing} + w_{\mathrm{cfg}} r_{\mathrm{cfg}} + R_{\blendr}$
\State Step scheduler with $\widehat{\varepsilon}$ to obtain $x_{t-1}$
\end{algorithmic}
\end{algorithm}

\section{Extracting Image Descriptions}
\label{supp:image_descriptions}

Section~\ref{subsec:blendr_training} mentioned that attribute descriptions are extracted with LLaVa-Next \cite{liu_arxiv2024_llavanext}, but did not detail the specific prompting strategies. This section provides the complete LLaVa-Next prompts for each attribute type and demonstrates how careful prompt design obtains clean, focused annotations essential for reproducibility.

To adapt the Stable Diffusion backbone using LoRA and Textual Inversion (TI), we use input prompts of the form ``a photo of a $[V_i]$ \texttt{[metaclass]}. $a$.''
Here, $a$ denotes the attribute description, which is utilized for both Stable Diffusion training and image generation (see Supplementary Section~\ref{supp:training_prompts} for details).

Attribute descriptions are extracted using the LLaVa-Next Image-Text to Text model \cite{liu_arxiv2024_llavanext}.
The model is prompted with an input image and tasked to describe the target attribute.
We observed that LLaVa-Next may include information about image regions not specified in the prompt (e.g., foreground details when targeting background attributes).
To address this, prompts are designed to focus on specific attributes and exclude unrelated descriptions.

We extract attribute descriptions for both background and foreground objects, such as birds (CUB) and cars (Cars).
Additionally, we obtain detailed annotations for bird pose (CUB) and camera/viewing angle (Cars).

\begin{figure}[H]
    \centering
    \begin{subfigure}[h]{0.24\textwidth}
        \centering
        \includegraphics[width=\textwidth,height=0.15\textheight,keepaspectratio]{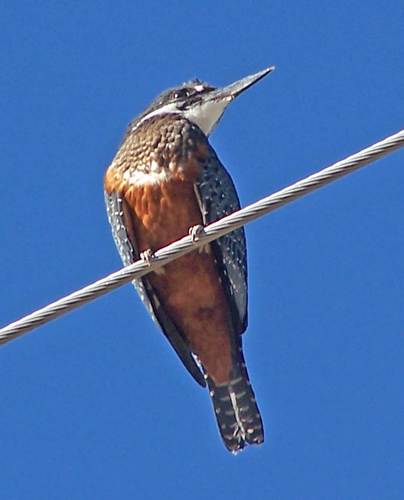}
        \caption{\textbf{Background:} \texttt{The background of the image is a clear blue sky with no visible clouds. The sky appears to be bright and sunny, suggesting good weather conditions.} \\ \textbf{Pose:} \texttt{The bird is perched on a wire, facing to the right. It has its head slightly tilted upwards, and its tail is slightly raised.} \\ \textbf{Foreground:} \texttt{The bird has a brown and white body, a black and white tail, and a white and black head.}}
    \end{subfigure}
    \hfill
    \begin{subfigure}[h]{0.24\textwidth}
        \centering
        \includegraphics[width=\textwidth,height=0.25\textheight,keepaspectratio]{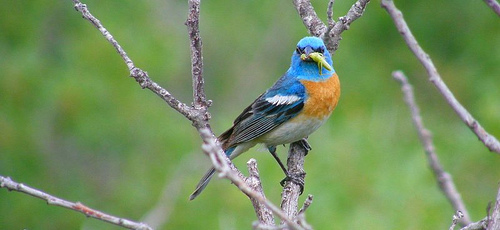}
        \caption{\textbf{Background:} \texttt{The background of the image features a natural setting with a variety of green hues. There are several branches with thin twigs extending from them, suggesting a dense vegetation.} \\ \textbf{Pose:} \texttt{The bird is perched on a thin, bare branch. It is facing slightly to the left with its head turned towards the camera. The bird's tail is extended, and its wings are folded neatly at its sides.} \\ \textbf{Foreground:} \texttt{The bird has a blue head, orange chest, and black and white stripes on its wings.}}
    \end{subfigure}
    \hfill
    \begin{subfigure}[h]{0.24\textwidth}
        \centering
        \includegraphics[width=\textwidth,height=0.25\textheight,keepaspectratio]{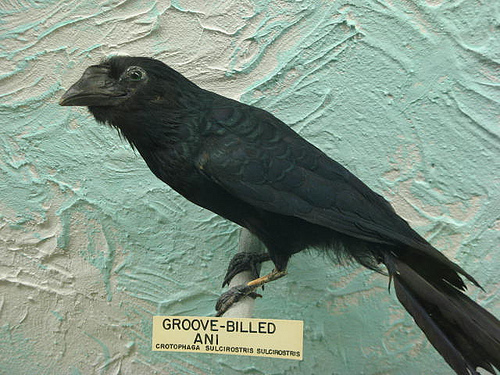}
        \caption{\textbf{Background:} \texttt{The background of the image features a textured wall with a light blue color. The wall has a rough, uneven surface with visible lines and cracks, giving it a somewhat aged appearance.} \\ \textbf{Pose:} \texttt{The bird is perched on a branch, facing to the left with its head slightly tilted downwards. Its wings are folded at its sides, and its tail feathers are spread out behind it.} \\ \textbf{Foreground:} \texttt{The bird has a black feathered body, a long beak, and a yellow label with black text.}}
    \end{subfigure}
    \hfill
    \begin{subfigure}[h]{0.24\textwidth}
        \centering
        \includegraphics[width=\textwidth,height=0.25\textheight,keepaspectratio]{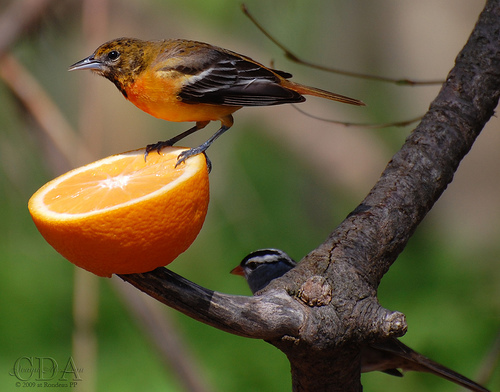}
        \caption{\textbf{Background:} \texttt{The background of the image features a blurred natural setting. There are green leaves and branches, suggesting a forest or woodland environment.} \\ \textbf{Pose:} \texttt{The bird is perched on a branch, which is part of a tree. The bird is facing towards the right side of the image, with its body oriented slightly downwards.} \\ \textbf{Foreground:} \texttt{The bird has black and orange feathers.}}
    \end{subfigure}
    \\
    \begin{subfigure}[h]{0.24\textwidth}
        \centering
        \includegraphics[width=\textwidth,height=0.25\textheight,keepaspectratio]{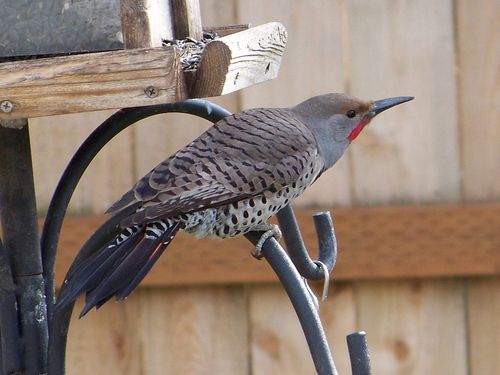}
        \caption{\textbf{Background:} \texttt{The background of the image features a wooden fence with horizontal slats. The fence appears to be weathered, suggesting it has been exposed to the elements for some time.} \\ \textbf{Pose:} \texttt{The bird is perched on a metal structure, which appears to be part of a bird feeder or a similar type of bird-friendly equipment. The bird is facing to the right, with its head turned slightly towards the camera.} \\ \textbf{Foreground:} \texttt{The bird has a gray body with black and white stripes on its tail.}}
    \end{subfigure}
    \hfill
    \begin{subfigure}[h]{0.24\textwidth}
        \centering
        \includegraphics[width=\textwidth,height=0.25\textheight,keepaspectratio]{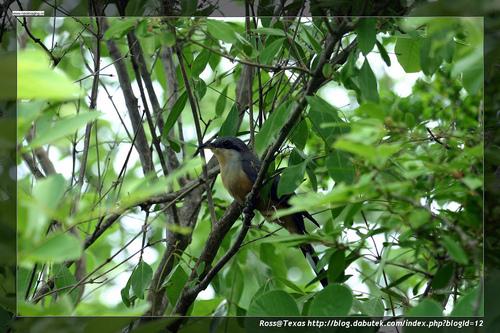}
        \caption{\textbf{Background:} \texttt{The background of the image features a dense canopy of green leaves, suggesting a lush, tropical or subtropical environment. The leaves are various shades of green, indicating a healthy and thriving plant life.} \\ \textbf{Pose:} \texttt{The bird is perched on a branch of a tree. It is facing to the left, with its head turned slightly towards the camera. The bird's body is oriented towards the right, and its tail is pointing downwards.} \\ \textbf{Foreground:} \texttt{The bird has a black head and a brown body.}}
    \end{subfigure}
    \hfill
    \begin{subfigure}[h]{0.24\textwidth}
        \centering
        \includegraphics[width=\textwidth,height=0.25\textheight,keepaspectratio]{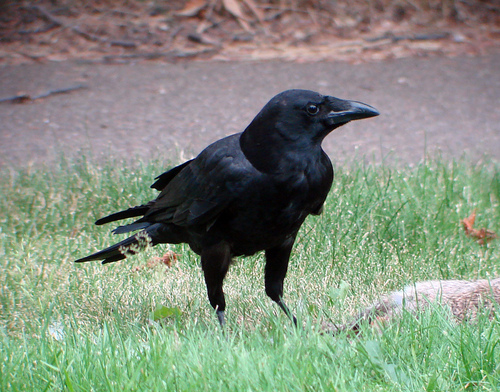}
        \caption{\textbf{Background:} \texttt{The background of the image features a natural setting with a grassy area. There is a path that appears to be made of gravel or small stones, leading towards the grass.} \\ \textbf{Pose:} \texttt{The bird is standing upright on one leg, with its body facing forward and its head turned to the side, giving the impression that it is looking to the right.} \\ \textbf{Foreground:} \texttt{The bird has black feathers.}}
    \end{subfigure}
    \hfill
    \begin{subfigure}[h]{0.24\textwidth}
        \centering
        \includegraphics[width=\textwidth,height=0.25\textheight,keepaspectratio]{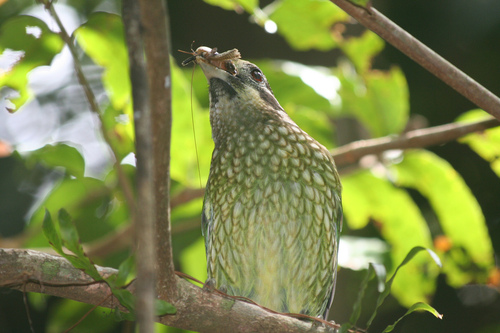}
        \caption{\textbf{Background:} \texttt{The background of the image features a lush green forest. The trees are dense with leaves, creating a canopy of green that fills the space behind the bird.} \\ \textbf{Pose:} \texttt{The bird is perched on a branch, facing to the right. It appears to be in a relaxed posture, with its head slightly tilted to the side and its beak closed.} \\ \textbf{Foreground:} \texttt{The bird has a spotted pattern on its body.}}
    \end{subfigure}
    \caption{Example images of birds with attribute descriptions.}
    \label{supp:fig:cub_attributes}
\end{figure}

Table~\ref{supp:tab:llava_next_prompts} lists the prompts used for attribute extraction.
Figure \ref{supp:fig:cub_attributes} presents example images of birds.

\begin{table}[H]
\centering
\begin{adjustbox}{width=1.0\linewidth}
\begin{tabular}{l p{0.8\linewidth}}
\toprule
\textbf{Attribute} & \textbf{Prompt} \\
\midrule
\textbf{CUB} & \\
Foreground & List distinct attributes or patterns of the foreground object, such as stripes, dots, text etc. One sentence answer. \\
Background & Describe only the background of the image in detail, do not mention any detail about any bird. \\
Pose / Camera Angle & Describe just the pose of the bird in detail, but do not describe the bird nor mention the bird species or type. \\
\midrule
\textbf{Cars} & \\
Foreground & List distinct attributes or patterns of the foreground object, such as stripes, dots, text etc. One sentence answer. \\
Background & Describe only the background of the image in detail, do not mention any detail about any car or any text that appears in the background. \\
Pose / Camera Angle & Describe how the camera is facing the car, and which parts of the car are visible. Do not mention any detail about the brand or model of the car. \\
\bottomrule
\end{tabular}
\end{adjustbox}
\caption{LLaVa-Next prompts for extracting attribute descriptions across different datasets, reformatted to a single column structure.}
\label{supp:tab:llava_next_prompts}
\end{table}

\section{CLIP Image Description Similarity Ranking}
\label{supp:clip_description_ranking}
Section~\ref{subsec:blendr_generation} introduced selecting novel attributes from the least similar 50\% and retrieving top-4 neighbors, but did not explain the underlying hard negative mining strategy. This section provides the complete overview of the similarity ranking and selection approach.

To generate samples that increase intra-class diversity, we first compute similarity rankings between all LLaVA-Next-generated attribute descriptions in the dataset.
Specifically, we encode each unique description per attribute (foreground, background, pose/camera angle) using the CLIP ViT-L/14 \cite{Radford2021LearningTV} text encoder, obtaining L2-normalized embeddings in the CLIP text feature space.
We then compute pairwise cosine similarities between all embeddings, and for each description, maintain a ranked list of all other descriptions sorted by similarity score in descending order.
During synthetic image generation, for each target class instance, we perform hard negative mining by randomly selecting an attribute description $a$ from the least similar $50\%$ of the similarity ranking, relative to the instance's ground-truth attribute.
This ensures that the selected attribute is sufficiently dissimilar to introduce diversity.
To maintain semantic coherence and avoid unrealistic combinations, we then retrieve the top-4 most similar description to this hard sample description $a$, forming a tight semantic cluster of five related yet challenging attribute descriptions. These five prompts guide the \blendr generation process through \opblendrop operations Union and Intersection, creating synthetic images that exhibit novel attributes while maintaining class characteristics to a large degree, which is demonstrated by improved DML performance (Table~\ref{tab:blendr_attribute_performance} in the main paper) and CLIP Image scores (Appendix Section \ref{supp:clip_image_evaluation}).

\section{Image Alignment}
\label{supp:image_alignment}

Section~\ref{subsec:blendr_training} briefly mentioned object centering but did not explain the procedure. This section details our foreground-aligned square cropping using Segment Anything masks, the context parameter $c$, and zero-padding strategy, demonstrating that proper alignment prevents truncated object generation.

\begin{figure}[H]
    \centering
    \begin{subfigure}[h]{0.24\linewidth}
        \centering
        \includegraphics[width=\linewidth]{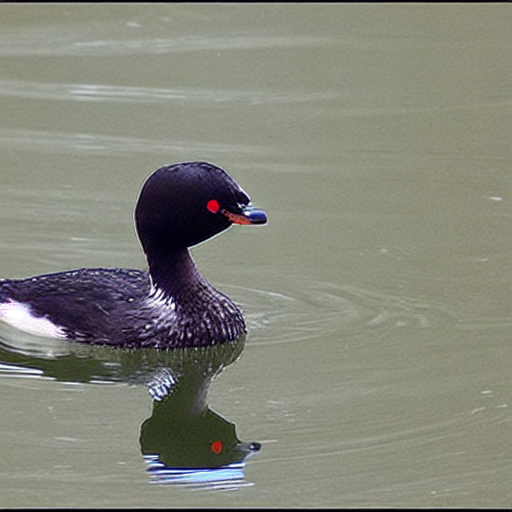}
        \caption{Example 1: \emph{Un}aligned training images}
        \label{supp:fig:alignment_comparison_cutoff1}
    \end{subfigure}
    \hfill
    \begin{subfigure}[h]{0.24\linewidth}
        \centering
        \includegraphics[width=\linewidth]{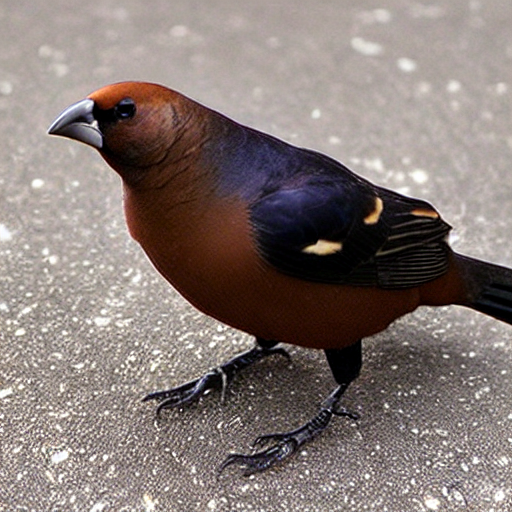}
        \caption{Example 2: \emph{Un}aligned training images}
        \label{supp:fig:alignment_comparison_cutoff2}
    \end{subfigure}
    \hfill
    \begin{subfigure}[h]{0.24\linewidth}
        \centering
        \includegraphics[width=\linewidth]{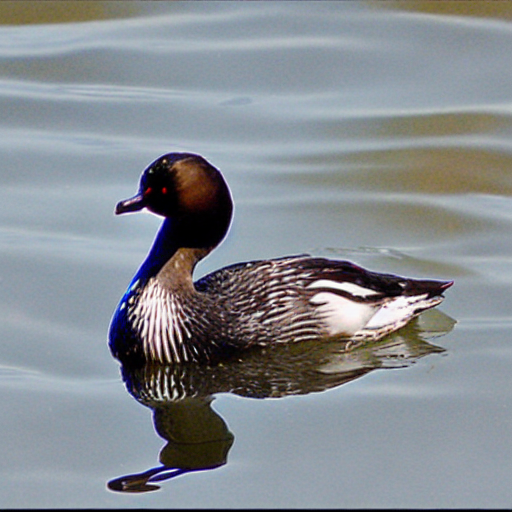}
        \caption{Example 1: \emph{Aligned} training images}
        \label{supp:fig:alignment_comparison_aligned1}
    \end{subfigure}
    \hfill
    \begin{subfigure}[h]{0.24\linewidth}
        \centering
        \includegraphics[width=\linewidth]{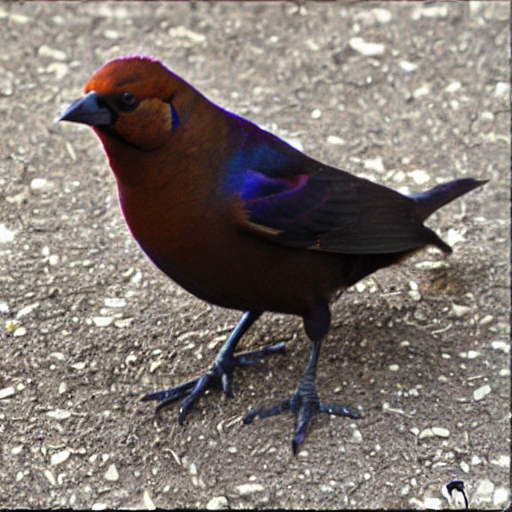}
        \caption{Example 2: \emph{Aligned} training images}
        \label{supp:fig:alignment_comparison_aligned2}
    \end{subfigure}
    
    \caption{Comparison of images generated by Stable Diffusion models trained on unaligned versus aligned datasets.}
    \label{supp:fig:alignment_comparison}
\end{figure}

Training generative models on unaligned datasets, where objects appear at arbitrary positions and scales, presents significant challenges.
Random or center crops frequently truncate objects, causing the generator to also generate truncated target objects, as seen in Figures~\ref{supp:fig:alignment_comparison_cutoff1} and \ref{supp:fig:alignment_comparison_cutoff2}.

To address this, we introduce a foreground-aligned square cropping procedure that centers objects within the training images while preserving their complete structure.
We first extract foreground segmentation masks for all training images using language-guided Segment Anything\footnote{\url{https://github.com/luca-medeiros/lang-segment-anything}}, following the protocol of~\cite{DBLP:journals/corr/abs-2411-02592}.

\begin{figure}[H]
    \centering
    \begin{subfigure}[h]{0.24\linewidth}
        \centering
        \includegraphics[width=\linewidth]{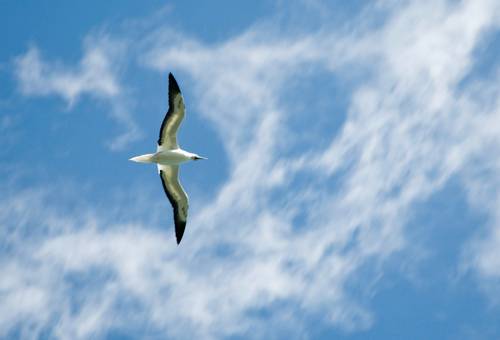}
        \caption{Original image}
        \label{supp:fig:alignment_original_image}
    \end{subfigure}
    \hfill
    \begin{subfigure}[h]{0.24\linewidth}
        \centering
        \includegraphics[width=\linewidth]{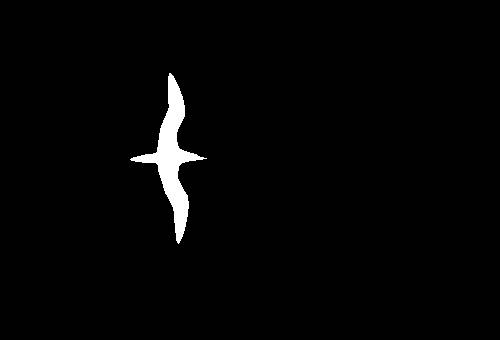}
        \caption{Foreground mask}
        \label{supp:fig:alignment_mask}
    \end{subfigure}
    \hfill
    \begin{subfigure}[h]{0.24\linewidth}
        \centering
        \includegraphics[width=\linewidth]{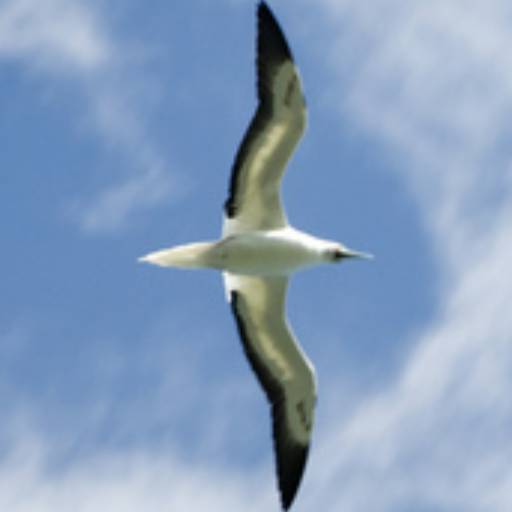}
        \caption{Tight crop}
        \label{supp:fig:alignment_tight_crop}
    \end{subfigure}
    \hfill
    \begin{subfigure}[h]{0.24\linewidth}
        \centering
        \includegraphics[width=\linewidth]{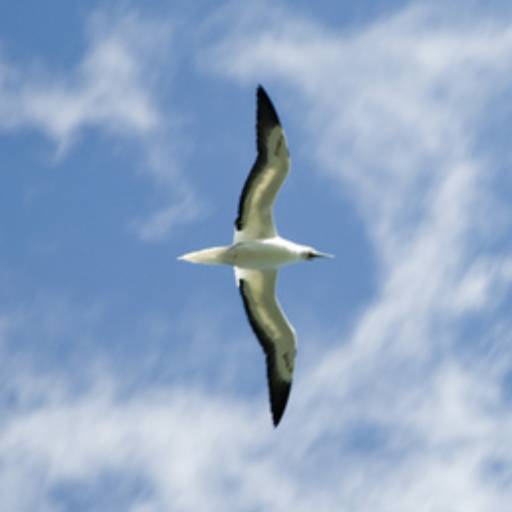}
        \caption{Loose crop}
        \label{supp:fig:alignment_loose_crop}
    \end{subfigure}
    
    \caption{Visualization of the square cropping image alignment approach. Given the original image (\ref{supp:fig:alignment_original_image}), a foreground mask extracted using language-guided Segment Anything (\ref{supp:fig:alignment_mask}) is used to crop the foreground object into a square image with a specified context margin $c$. The context parameter $c$ controls the border between the object's bounding box and the image edge: a value of $c=0.0$ produces a tight crop with no margin (\ref{supp:fig:alignment_tight_crop}), while larger values preserve more background context (\ref{supp:fig:alignment_loose_crop}, $c=0.5$). All crops are resized to $512 \times 512$ pixels with zero padding applied when the crop extends beyond the original image boundaries.}
    \label{supp:fig:alignment}
\end{figure}

Given an image and its corresponding mask, we compute the tight bounding box around the foreground object and extend it to a square crop based on the larger spatial dimension.
A context parameter $c$ controls the relative margin added around the object: $c=0$ produces a tight crop with the object edges touching the image border, while larger values (e.g., $c=0.1$ or $c=0.5$) preserve increasing amounts of background context. 
An example image with its mask and foreground-aligned square crops with different $c$ are shown in Figure~\ref{supp:fig:alignment}.
The square crop is extracted using affine grid sampling, resized to $512 \times 512$ resolution and zero-padded if it extends beyond image boundaries to keep the object centered.
This is common for rectangular objects like cars, which often can not be cropped without additional padding.
This alignment procedure ensures that Stable Diffusion model learns to generate complete, well-composed objects.
The advantage of the proposed approach is visible in Figures~\ref{supp:fig:alignment_comparison_aligned1} and \ref{supp:fig:alignment_comparison_aligned2}, which can be directly compared to images generated with Stable 
Diffusion trained on unaligned images.

\section{Input Prompts for Stable Diffusion training}
\label{supp:training_prompts}
As outlined in Section~\ref{subsec:blendr_training} in the main paper, we fine-tune Stable Diffusion using LoRA~\cite{DBLP:conf/iclr/HuSWALWWC22} and Textual Inversion~\cite{DBLP:conf/iclr/GalAAPBCC23}.
As we target to increase intra-class diversity using text descriptions, we need to enable the model to condition on attribute descriptions during generation.
Therefore, we incorporate attribute information into the training prompts. Specifically, for each training sample, we randomly select attribute descriptions $a$ from the image's metadata. During preprocessing, we apply foreground-aligned square cropping (Section~\ref{supp:image_alignment}) with randomly sampled context values $c \in \{0.2, 0.3, 0.4, 0.5\}$ to introduce scale variation and ensure sufficient background context.
Training prompts follow the template ``a photo of a $[V_i]$ \texttt{[metaclass]}. $a$.'', where $[V_i]$ denotes the Textual Inversion token for class $i$, $a$ are the sampled attribute descriptions, and \texttt{[metaclass]} is the dataset-specific category. For CUB it is \texttt{bird}, and for Cars it is \texttt{car}.

\section{Text Embedding Interpolation}
\label{supp:embedding_space_interpolation}

In Section~\ref{subsubsec:methodology_textembinterp} of the main paper, we introduced Text Embedding Interpolation (\opembmix), which blends text embeddings from a target anchor prompt $c_1$ (containing the target class $[V_i]$ and novel attribute $a$) and an attribute donor prompt $c_2$ (containing a donor class $[V_j]$ that exhibits attribute $a$) using time-dependent interpolation weights $\alpha_1(t)$ and $\alpha_2(t)$.
This interpolation pre-aligns the initial latent structure toward the desired attribute by leveraging the donor class during early denoising, then gradually transitions to the target class for final refinement.
In Section~\ref{subsec:blendr_generation}, we outlined the experimental configuration for \opembmix.
Here, we provide detailed information about the scheduling function that is used for interpolating weights across denoising timesteps with additional evaluation of target attribute adherence and authentic class prototype similarity.

The \opembmix weights $\alpha_1(t)$ and $\alpha_2(t)$ in Equation~\ref{eq:mixembed} are computed using a cosine decay schedule.
This schedule ensures smooth transitions between the donor and anchor prompts while maintaining the normalization constraint $\alpha_1(t) + \alpha_2(t) = 1$ at all denoising steps. The schedule is parameterized by a fade ratio $t_\star$ (set to $0.2$ in our experiments), defining the normalized timestep at which the donor prompt fully fades out. Specifically, the donor weight follows $\alpha_2(t) = \gamma(t)$, where $\gamma(t) = 0.5 \cdot (1 + \cos(\pi \cdot t_{\text{norm}}))$ for $t \in [0, t_\star]$ and $\gamma(t) = 0$ for $t > t_\star$. Here, $t_{\text{norm}} = t / t_\star$ normalizes the current denoising timestep ratio to the range $[0, 1]$ within the active interpolation window. The cosine formulation provides a smooth decay: at $t=0$ we have $\gamma(0) = 1.0$ (fully donor), at $t=t_\star$ we have $\gamma(t_\star) = 0.0$ (fully anchor), with a gradual transition in between. The anchor weight is computed as $\alpha_1(t) = 1 - \alpha_2(t)$ to maintain normalization.
By completing the interpolation within the first $t_\star$ of denoising steps, this early donor injection strategy biases the initial latent structure toward the novel attribute while allowing the remaining steps to refine details using the target anchor concept, ensuring both attribute novelty and intra-class consistency.

To evaluate the impact of \opembmix and $t_\star$ on both target attribute adherence and how closely the generated image resembles the target class, we generate images using \opembmix only with $t_\star$ values of $0.2$, $0.4$, $0.6$, and $0.8$.
For all experiments, $20$ images per class are generated.

We utilize the CLIP similarity metric defined in Section~\ref{subsec:results_quantitative_results} of the main paper to evaluate target attribute adherence, which computes the cosine similarity between CLIP embeddings of the target attribute description and the generated images. Additionally, we measure the cosine similarity between embeddings of the generated images and the mean embeddings of the target class, extracted from authentic training data using a pre-trained ResNet-50 DML model trained with ProxyAnchor loss.

The results are presented in Table~\ref{tab:text_embedding}.

As shown in the table, the CLIP score increases with larger $t_\star$ values, indicating improved target attribute adherence. This demonstrates that a longer pre-alignment of the latent through \opembmix does indeed enhance target attribute adherence.

As expected, the cosine similarity to the target class decreases with increasing $t_\star$. This occurs because fewer timesteps remain available to steer the denoising direction toward the target class, causing the model to generate features of the attribute donor class for a larger portion of the denoising process.

Since the goal of \blendr is to synthesize images of the target class with novel attributes, we aim to minimize the influence of the attribute donor class. The residual set operations used in \blendr serve as approach to steer the denoising trajectory toward the target attribute while preserving target class characteristics.

Therefore, we select $t_\star = 0.2$ as the default value, as it provides sufficient pre-alignment for the latent without excessively reducing target class similarity.

\begin{figure}[H]
    \centering
    \includegraphics[width=0.6 \columnwidth]{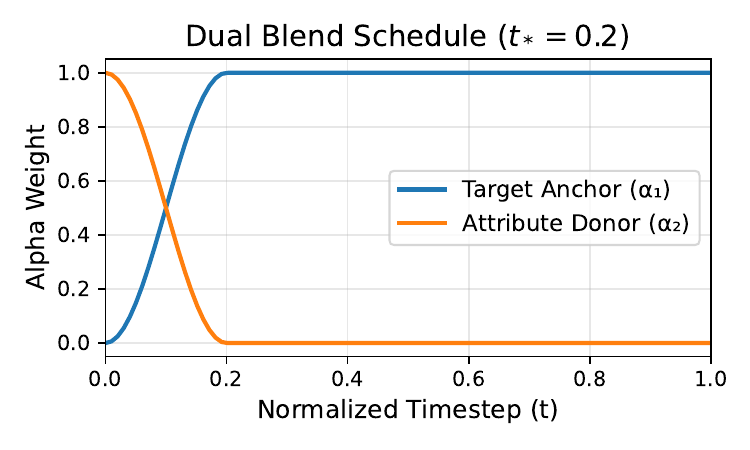}
    \vspace{-8mm}
    \caption{Text embedding interpolation weights $\alpha_1(t)$ (target anchor) and $\alpha_2(t)$ (attribute donor) across normalized timesteps. The cosine decay schedule transitions from donor-dominated ($\alpha_2 = 1$) to anchor-only ($\alpha_1 = 1$) by fade ratio $t_\star = 0.2$.}
    \label{supp:fig:tei_blending_schedule}
    \vspace{-4mm}
\end{figure}

The used \opembmix schedule is visualized in Figure~\ref{supp:fig:tei_blending_schedule}.

\begin{table}[H]
\centering
\caption{Impact of the Text Embedding Interpolation parameter $t_\star$ on target attribute adherence (CLIP similarity) and target class similarity (cosine similarity to class mean). Higher $t_\star$ values improve attribute adherence but reduce class similarity. We select $t_\star = 0.2$ to balance both objectives.}
\label{tab:text_embedding}
\begin{tabular}{ccc}
\toprule
$t$ & Cosine Similarity $\uparrow$ & CLIP $\uparrow$ \\
\midrule
0.2 & $0.8075 \pm 0.0940$ & $0.1676 \pm 0.0275$ \\
0.4 & $0.7743 \pm 0.1080$ & $0.1703 \pm 0.0270$ \\
0.6 & $0.7286 \pm 0.1222$ & $0.1715 \pm 0.0267$ \\
0.8 & $0.6798 \pm 0.1305$ & $0.1725 \pm 0.0261$ \\
\bottomrule
\end{tabular}
\end{table}

\section{Residual Set Operations}
\label{supp:operation_parameters}

As outlined in Section~\ref{subsec:prompts} of the main paper, \blendr uses time-varying weight functions $\beta(t)$ to modulate the contribution of the Residual Set Operations (RSO) across timesteps $t$.

To evaluate the influence of $\beta(t)$ on both union and intersection operations and to determine appropriate parameter values, we conduct experiments on the CUB dataset using background as the target attribute. We generate images using text embedding interpolation with $t_\star = 0.2$, varying $\beta(0)$ across values of $0.5$ and $1.0$--$9.0$, with a cosine ramp decaying to $0.0$ at $t = 0.8$, similar to the schedule used for Text Embedding Interpolation. We evaluate Classifier-Free Guidance (CFG) values of $2.0$, $4.0$, and $7.5$, following~\cite{gendataagent}. For each experiment, $20$ images per class are generated.

As shown in Table~\ref{tab:combined_main}, increasing $\beta(0)$ improves target attribute adherence as measured by CLIP similarity across all operations and CFG values. However, similar to the behavior observed with Text Embedding Interpolation, the cosine similarity between generated images and the mean embedding of the target class decreases as the weight for RSO increases, steering the denoising trajectory away from the target class.

With increasing CFG values, cosine similarity improves in most cases across all evaluated weight settings. This is expected, as CFG steers the denoising trajectory toward the target class, and higher CFG values amplify this effect.

Based on the overall performance, we select a CFG value of $4.0$ as it provides the best trade-off between target attribute adherence and target class similarity. Furthermore, as shown in the Table~\ref{tab:combined_main}, $\beta(0)$ values in the range of $3.0$--$6.0$ yield the highest target attribute adherence (measured via CLIP) across the majority of configurations, including all CFG values and both operations (union and intersection). We therefore select $\beta(0) \in \{3.0, 4.0, 5.0, 6.0\}$.

However, as evident in Table~\ref{tab:combined_main}, higher $\beta(0)$ values negatively impact cosine similarity, particularly for the union operation. To mitigate this issue, we employ the additional mechanisms introduced in Section~\ref{subsec:prompts}: orthogonalization with respect to the CFG residual to remove directions already captured by guidance, thereby preventing over steering (Eq.~\ref{eq:orth}), and norm clamping of the combined \blendr residual relative to the guidance norm (Eq.~\ref{eq:clamp}).

We evaluate norm clamping with $\tau$ values of $1.0$, $2.0$, and the respective $\beta(0)$ value used ($3.0$, $4.0$, $5.0$, and $6.0$). All experiments use a CFG value of $4.0$. The results for both intersection and union operations are presented in Table~\ref{tab:combined_clamp}.

As shown in the table, applying orthogonalization and clamping preserves target attribute adherence (measured via CLIP) to a large extent while substantially improving cosine similarity to the target class. For example, for the intersection operation with $\beta(0) = 6.0$, the cosine similarity increases from $0.650$ (unclamped) to $0.781$ (with $\tau = 2.0$). Similarly, for the union operation with $\beta(0) = 6.0$, the cosine similarity increases from $0.386$ (unclamped) to $0.797$ (with $\tau = 1.0$). These results demonstrate the effectiveness of the orthogonalization and clamping operations.

Based on the overall results, we utilize CFG value of $4.0$, randomly select $\beta(0)$ from $\{3.0, 4.0, 5.0, 6.0\}$, and use orthogonalization with norm clamping $\tau=3.0$ to prevent the impact across different target attributes and datasets.

Using this combination of parameters, \blendr is able to generate images that adhere to both the target attribute and the target class characteristics.
In summary, these results validate the effectiveness of \blendr's complementary control mechanisms. While increasing $\beta(0)$ enhances target attribute adherence, it can compromise target class similarity if left unconstrained.
The orthogonalization and norm clamping operations successfully mitigate this effect, preserving attribute adherence while substantially recovering target class similarity.
This demonstrates that \blendr provides fine-grained control over the attribute-class trade-off, enabling the reliable synthesis of images that simultaneously exhibit novel target attributes and maintain the semantic integrity of the target class.

\begin{table}[H]
\centering
\resizebox{\textwidth}{!}{%
\begin{tabular}{ccccccccccccc}
\toprule
 & \multicolumn{4}{c}{CFG=2.0} & \multicolumn{4}{c}{CFG=4.0} & \multicolumn{4}{c}{CFG=7.5} \\
\cmidrule(lr){2-5} \cmidrule(lr){6-9} \cmidrule(lr){10-13}
 & \multicolumn{2}{c}{Intersection} & \multicolumn{2}{c}{Union} & \multicolumn{2}{c}{Intersection} & \multicolumn{2}{c}{Union} & \multicolumn{2}{c}{Intersection} & \multicolumn{2}{c}{Union} \\
\cmidrule(lr){2-3} \cmidrule(lr){4-5} \cmidrule(lr){6-7} \cmidrule(lr){8-9} \cmidrule(lr){10-11} \cmidrule(lr){12-13}
$\beta(0)$ & CLIP $\uparrow$ & Cosine Similarity $\uparrow$ & CLIP $\uparrow$ & Cosine Similarity $\uparrow$ & CLIP $\uparrow$ & Cosine Similarity $\uparrow$ & CLIP $\uparrow$ & Cosine Similarity $\uparrow$ & CLIP $\uparrow$ & Cosine Similarity $\uparrow$ & CLIP $\uparrow$ & Cosine Similarity $\uparrow$ \\
\midrule
0.5 & 0.1714 & 0.7604 & 0.1751 & 0.7146 & 0.1687 & 0.8072 & 0.1708 & 0.8013 & 0.1707 & 0.7915 & 0.1721 & 0.7896 \\
1.0 & 0.1734 & 0.7344 & 0.1794 & 0.6135 & 0.1697 & 0.8037 & 0.1732 & 0.7752 & 0.1711 & 0.7901 & 0.1734 & 0.7883 \\
2.0 & 0.1779 & 0.6608 & 0.1824 & 0.4565 & 0.1716 & 0.7933 & 0.1777 & 0.6797 & 0.1719 & 0.7887 & 0.1762 & 0.7653 \\
3.0 & 0.1800 & 0.5790 & 0.1786 & 0.3941 & 0.1737 & 0.7739 & 0.1812 & 0.5602 & 0.1732 & 0.7869 & 0.1786 & 0.7206 \\
4.0 & 0.1821 & 0.5104 & 0.1752 & 0.3723 & 0.1755 & 0.7450 & 0.1810 & 0.4609 & 0.1741 & 0.7816 & 0.1813 & 0.6544 \\
5.0 & 0.1828 & 0.4602 & 0.1725 & 0.3604 & 0.1777 & 0.7043 & 0.1793 & 0.4104 & 0.1752 & 0.7753 & 0.1835 & 0.5800 \\
6.0 & 0.1824 & 0.4264 & 0.1709 & 0.3529 & 0.1793 & 0.6502 & 0.1781 & 0.3856 & 0.1765 & 0.7639 & 0.1840 & 0.5051 \\
7.0 & 0.1818 & 0.4049 & 0.1686 & 0.3469 & 0.1802 & 0.5961 & 0.1762 & 0.3712 & 0.1776 & 0.7426 & 0.1835 & 0.4547 \\
8.0 & 0.1807 & 0.3915 & 0.1679 & 0.3398 & 0.1809 & 0.5458 & 0.1734 & 0.3603 & 0.1786 & 0.7291 & 0.1820 & 0.4204 \\
9.0 & 0.1793 & 0.3828 & 0.1660 & 0.3354 & 0.1813 & 0.5017 & 0.1723 & 0.3514 & 0.1799 & 0.7033 & 0.1808 & 0.3971 \\
\bottomrule
\end{tabular}%
}
\caption{Effect of the weight parameter $\beta(0)$ on target attribute adherence (CLIP similarity) and target class similarity (cosine similarity) for intersection and union operations across different CFG values. Higher $\beta(0)$ values improve CLIP scores but reduce cosine similarity, reflecting the trade-off between attribute injection and class preservation. Experiments are conducted on the CUB dataset with background as the target attribute.}
\label{tab:combined_main}
\end{table}

\begin{table}[H]
\centering
\resizebox{\textwidth}{!}{%
\begin{tabular}{ccccccccccccccccc}
\toprule
 & \multicolumn{8}{c}{Intersection} & \multicolumn{8}{c}{Union} \\
\cmidrule(lr){2-9} \cmidrule(lr){10-17}
 & \multicolumn{2}{c}{$\tau=1$} & \multicolumn{2}{c}{$\tau=2$} & \multicolumn{2}{c}{$\tau=\beta(0)$} & \multicolumn{2}{c}{Unclamped} & \multicolumn{2}{c}{$\tau=1$} & \multicolumn{2}{c}{$\tau=2$} & \multicolumn{2}{c}{$\tau=\beta(0)$} & \multicolumn{2}{c}{Unclamped} \\
\cmidrule(lr){2-3} \cmidrule(lr){4-5} \cmidrule(lr){6-7} \cmidrule(lr){8-9} \cmidrule(lr){10-11} \cmidrule(lr){12-13} \cmidrule(lr){14-15} \cmidrule(lr){16-17}
$\beta(0)$ & CLIP $\uparrow$ & Cos. Sim. $\uparrow$  & CLIP $\uparrow$ & Cos. Sim. $\uparrow$  & CLIP $\uparrow$ & Cos. Sim. $\uparrow$  & CLIP $\uparrow$ & Cos. Sim. $\uparrow$  & CLIP $\uparrow$ & Cos. Sim. $\uparrow$  & CLIP $\uparrow$ & Cos. Sim. $\uparrow$  & CLIP $\uparrow$ & Cos. Sim. $\uparrow$  & CLIP $\uparrow$ & Cos. Sim. $\uparrow$  \\
\midrule
3 & 0.171 & 0.801 & 0.173 & 0.793 & 0.173 & 0.793 & 0.174 & 0.774 & 0.171 & 0.798 & 0.174 & 0.783 & 0.175 & 0.769 & 0.181 & 0.560 \\
4 & 0.172 & 0.799 & 0.174 & 0.787 & 0.174 & 0.783 & 0.176 & 0.745 & 0.171 & 0.798 & 0.174 & 0.781 & 0.177 & 0.745 & 0.181 & 0.461 \\
5 & 0.171 & 0.799 & 0.174 & 0.782 & 0.175 & 0.774 & 0.178 & 0.704 & 0.172 & 0.798 & 0.174 & 0.780 & 0.179 & 0.718 & 0.179 & 0.410 \\
6 & 0.172 & 0.799 & 0.174 & 0.781 & 0.176 & 0.763 & 0.179 & 0.650 & 0.172 & 0.797 & 0.174 & 0.779 & 0.180 & 0.688 & 0.178 & 0.386 \\
\bottomrule
\end{tabular}%
}
\caption{Effect of orthogonalization and norm clamping on target attribute adherence and target class similarity for intersection and union operations (CFG = 4.0). Clamping with $\tau \in \{1.0, 2.0, \beta(0)\}$ is compared against the unclamped baseline. Orthogonalization and clamping preserve CLIP scores while substantially improving cosine similarity, demonstrating the effectiveness of these operations in balancing attribute injection with class preservation.}
\label{tab:combined_clamp}
\vspace{1mm}
\end{table}

\section{DML Model Training}
\label{supp:dml_model_training}

Tables \ref{tab:hyperparams_resnet50}, \ref{tab:hyperparams_vit_small} and \ref{tab:hyperparams_dino_vits} show a comprehensive overview of the hyperparameters used for training DML models across different backbones and datasets for both Proxy Anchor (PA) \cite{proxy_anchor} and Potential Field (PF) \cite{potential_fields} losses.

\begin{table}[H]
\centering
\begin{adjustbox}{width=0.65\linewidth}
\begin{tabular}{@{}lcccccc@{}}
\toprule
\textbf{Parameter} & \multicolumn{2}{c}{\textbf{CUB}} & \multicolumn{2}{c}{\textbf{Cars}}\\
\cmidrule(lr){2-3} \cmidrule(lr){4-5}
& \textbf{PF} & \textbf{PA} & \textbf{PF} & \textbf{PA}\\
\midrule
Embedding Size & 512 & 512 & 512 & 512 \\
Learning Rate & 1e-4 & 1e-4 & 1e-4 & 1e-4\\
Weight Decay & 5e-4 & 5e-4 & 1e-4 & 5e-4\\
Batch Size & 100 & 180 & 100 & 180 \\
Epochs & 200 & 150 & 200 & 150 \\
LR Decay Step & 30 & 30 & 30 & 30 \\
LR Decay Gamma & 0.5 & 0.5 & 0.5 & 0.5 \\
Proxy LR Mult. & 100 & 100 & 100 & 100\\
Images/Class & 10 & 10 & 10 & 10 \\
Optimizer & AdamW & AdamW & AdamW & AdamW \\
\bottomrule
\end{tabular}
\end{adjustbox}
\caption{Hyperparameters for ResNet50 backbone across datasets and loss functions.}
\label{tab:hyperparams_resnet50}
\end{table}

\begin{table}[H]
\centering
\begin{adjustbox}{width=0.65\linewidth}
\begin{tabular}{@{}lcccccc@{}}
\toprule
\textbf{Parameter} & \multicolumn{2}{c}{\textbf{CUB}} & \multicolumn{2}{c}{\textbf{Cars}}\\
\cmidrule(lr){2-3} \cmidrule(lr){4-5} 
& \textbf{PF} & \textbf{PA} & \textbf{PF} & \textbf{PA}\\
\midrule
Embedding Size & 384 & 384 & 384 & 384  \\
Learning Rate & 1e-5 & 1e-5 & 1e-5 & 1e-5 \\
Weight Decay & 1e-2 & 1e-2 & 1e-2 & 1e-2\\
Batch Size & 100 & 180 & 100 & 180 \\
Epochs & 200 & 150 & 200 & 150  \\
LR Decay Step & 30 & 30 & 30 & 30 \\
LR Decay Gamma & 0.5 & 0.5 & 0.5 & 0.5 \\
Proxy LR Mult. & 100 & 100 & 100 & 100 \\
ViT FC LR Scale & -- & -- & -- & -- & \\
Images/Class & 10 & 10 & 10 & 10 & \\
Optimizer & AdamW & AdamW & AdamW & AdamW  \\
\bottomrule
\end{tabular}
\end{adjustbox}
\caption{Hyperparameters for ViT-Small backbone across datasets and loss functions.}
\label{tab:hyperparams_vit_small}
\end{table}

\begin{table}[H]
\centering
\begin{adjustbox}{width=0.65\linewidth}
\begin{tabular}{@{}lcccccc@{}}
\toprule
\textbf{Parameter} & \multicolumn{2}{c}{\textbf{CUB}} & \multicolumn{2}{c}{\textbf{Cars}} \\
\cmidrule(lr){2-3} \cmidrule(lr){4-5} 
& \textbf{PF} & \textbf{PA} & \textbf{PF} & \textbf{PA} \\
\midrule
Embedding Size & 384 & 384 & 384 & 384 \\
Learning Rate & 5e-6 & 5e-6 & 5e-6 & 5e-6  \\
Weight Decay & 1e-2 & 1e-2 & 1e-2 & 1e-2  \\
Batch Size & 100 & 180 & 100 & 180 \\
Epochs & 200 & 150 & 200 & 150  \\
LR Decay Step & 30 & 30 & 30 & 30 \\
LR Decay Gamma & 0.5 & 0.5 & 0.5 & 0.5  \\
Proxy LR Mult. & 100 & 100 & 100 & 100 \\
ViT FC LR Scale & 1 & 1 & 1 & 1  \\
Images/Class & 10 & 10 & 10 & 10  \\
Optimizer & AdamW & AdamW & AdamW & AdamW \\
\bottomrule
\end{tabular}
\end{adjustbox}
\caption{Hyperparameters for DINO-ViT-Small (dino\_vits16) backbone across datasets and loss functions.}
\label{tab:hyperparams_dino_vits}
\end{table}

\section{CLIP Score evaluation}
\label{supp:clip_score_evaluation}

In Section~\ref{subsec:results_quantitative_results} of the main paper, we presented quantitative results on target attribute adherence measured using CLIP similarity scores between generated images and their corresponding target attribute descriptions. Table~\ref{tab:clip_improvements} in the main paper reported the relative improvement (in percentage) of three generation strategies:
text embedding interpolation (\opembmix), residual set operations without text embedding interpolation (\opblendrop), and the full \blendr method combining both techniques.
Those techniques are compared to the baseline prompt-only approach (\opprompt). Here in the Supplementary Material, we provide
in Table~\ref{supp:tab:clip_improvements_full_absolute} all methods performance as absolute cosine similarity scores.
This expanded table enables direct comparison of both relative gains and absolute performance across all generation strategies.

\begin{table}[H]
\centering
\begin{adjustbox}{width=0.75\textwidth}
\begin{tabular}{l cccc}
\toprule
\textbf{Dataset / Target Attribute / Subset} & \textbf{\opprompt (Baseline)} & \textbf{\opembmix} & \textbf{\opblendrop} & \textbf{\opblendr} \\
\midrule
\multicolumn{5}{l}{\textbf{Cars-196 / Pose / Intersection}} \\
Full & 0.199 & 0.200 & 0.204 & 0.205 \\
Bottom 50\% & 0.175 & 0.181 & 0.188 & 0.189 \\
Bottom 25\% & 0.161 & 0.170 & 0.178 & 0.180 \\
Bottom 20\% & 0.156 & 0.166 & 0.176 & 0.178 \\
Bottom 10\% & 0.145 & 0.156 & 0.168 & 0.170 \\
Bottom 5\% & 0.134 & 0.148 & 0.160 & 0.163 \\
\midrule
\multicolumn{5}{l}{\textbf{Cars-196 / Background / Union}} \\
Full & 0.132 & 0.133 & 0.145 & 0.145 \\
Bottom 50\% & 0.097 & 0.102 & 0.118 & 0.119 \\
Bottom 25\% & 0.079 & 0.086 & 0.104 & 0.105 \\
Bottom 20\% & 0.074 & 0.081 & 0.100 & 0.102 \\
Bottom 10\% & 0.061 & 0.070 & 0.093 & 0.094 \\
Bottom 5\% & 0.049 & 0.060 & 0.080 & 0.086 \\
\midrule
\multicolumn{5}{l}{\textbf{CUB-2011-200 / Pose / Intersection}} \\
Full & 0.227 & 0.231 & 0.238 & 0.239 \\
Bottom 50\% & 0.204 & 0.215 & 0.225 & 0.227 \\
Bottom 25\% & 0.191 & 0.205 & 0.218 & 0.220 \\
Bottom 20\% & 0.187 & 0.202 & 0.216 & 0.218 \\
Bottom 10\% & 0.176 & 0.196 & 0.211 & 0.215 \\
Bottom 5\% & 0.166 & 0.189 & 0.207 & 0.211 \\
\midrule
\multicolumn{5}{l}{\textbf{CUB-2011-200 / Background / Union}} \\
Full & 0.172 & 0.176 & 0.188 & 0.190 \\
Bottom 50\% & 0.132 & 0.141 & 0.157 & 0.159 \\
Bottom 25\% & 0.108 & 0.120 & 0.139 & 0.143 \\
Bottom 20\% & 0.102 & 0.116 & 0.135 & 0.140 \\
Bottom 10\% & 0.086 & 0.103 & 0.126 & 0.130 \\
Bottom 5\% & 0.073 & 0.095 & 0.120 & 0.124 \\
\bottomrule
\end{tabular}
\end{adjustbox}
\caption{
Absolute CLIP cosine similarity scores between target attribute descriptions and generated images for \blendr variants and \opprompt baseline shown for all samples ("Full") and challenging cases ("Bottom X\%") where baseline similarity is lowest, demonstrating \blendr's ability to enhance generation in those challenging cases.}
\label{supp:tab:clip_improvements_full_absolute}
\end{table}

\section{CLIP Image Evaluation}
\label{supp:clip_image_evaluation}

While retrieval metrics effectively measure discriminative quality of learned representations, they do not directly assess the visual diversity and semantic fidelity of generated synthetic images. To complement our main evaluation, we analyze CLIP-based image similarity scores that quantify how well synthetic images match their class characteristics while maintaining realistic intra-class variation. This analysis is particularly important for deep metric learning, where training data diversity directly impacts the model's ability to generalize to unseen variations within each class.

For each class $c$, we compute a class prototype $\mathbf{p}_c$ as the normalized mean of CLIP ViT-L/14 image embeddings from authentic training images. For each image $\mathbf{x}$ (authentic or synthetic), we then compute the cosine similarity to the class prototype: $s(\mathbf{x}, c) = \mathbf{p}_c^T \cdot \text{CLIP}_{\text{img}}(\mathbf{x}) / ||\text{CLIP}_{\text{img}}(\mathbf{x})||_2$. Higher scores indicate stronger alignment with the class prototype. We analyze the distribution of these scores across all classes, computing mean $\mu$ (overall class alignment) and standard deviation $\sigma$ (intra-class diversity). An ideal synthetic dataset that adds intra-class variation should exhibit moderate mean similarity while having an increased standard deviation compared to authentic data.

Tables~\ref{tab:clip_similarity_cars} and~\ref{tab:clip_similarity_cub} present the CLIP similarity statistics for Cars and Cub. We evaluate two \blendr configurations: Background (Union) uses detailed background descriptions with union-based set operations, while Pose (Intersection) uses detailed pose descriptions with intersection-based operations.

\begin{table}[H]
\centering
\begin{adjustbox}{width=0.5\linewidth}
\begin{tabular}{@{}lccc@{}}
\toprule
Configuration & Authentic & Plain Prompt & \blendr \\
\midrule
\multicolumn{4}{l}{\textit{Background}} \\
Mean $\mu$ & 0.9033 & 0.8735 & 0.8579 \\
Std. Dev. $\sigma$ & 0.0437 & 0.0523 & 0.0590 \\
\midrule
\multicolumn{4}{l}{\textit{Pose}} \\
Mean $\mu$ & 0.9033 & 0.8823 & 0.8730 \\
Std. Dev. $\sigma$ & 0.0437 & 0.0470 & 0.0475 \\
\bottomrule
\end{tabular}
\end{adjustbox}
\caption{CLIP similarity statistics for Cars dataset. All scores are cosine similarities to class prototypes.}
\label{tab:clip_similarity_cars}
\end{table}

\begin{table}[H]
\centering
\begin{adjustbox}{width=0.5\linewidth}
\begin{tabular}{@{}lccc@{}}
\toprule
Configuration & Authentic & Plain Prompt & \blendr \\
\midrule
\multicolumn{4}{l}{\textit{Background}} \\
Mean $\mu$ & 0.9165 & 0.8823 & 0.8584 \\
Std. Dev. $\sigma$ & 0.0378 & 0.0374 & 0.0472 \\
\midrule
\multicolumn{4}{l}{\textit{Pose}} \\
Mean $\mu$ & 0.9165 & 0.8770 & 0.8550 \\
Std. Dev. $\sigma$ & 0.0378 & 0.0400 & 0.0477 \\
\bottomrule
\end{tabular}
\end{adjustbox}
\caption{CLIP similarity statistics for Cub dataset. All scores are cosine similarities to class prototypes.}
\label{tab:clip_similarity_cub}
\end{table}

\begin{figure}[H]
    \centering
    \begin{subfigure}[h]{0.48\columnwidth}
        \centering
        \includegraphics[width=\textwidth]{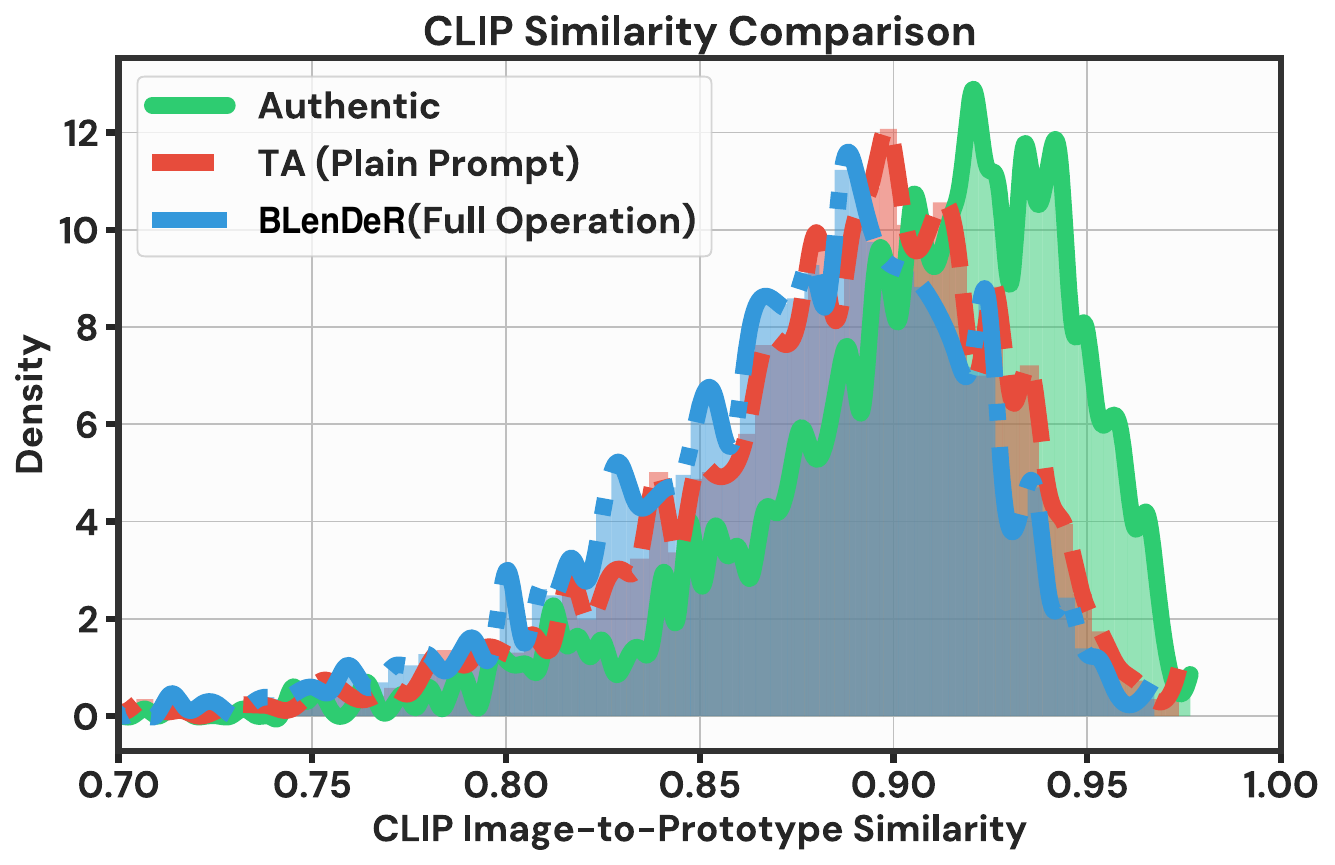}
        \caption{Cars - Pose}
        \label{supp:cossim_cars_pose}
    \end{subfigure}
    \hfill
    \begin{subfigure}[h]{0.48\columnwidth}
        \centering
        \includegraphics[width=\textwidth]{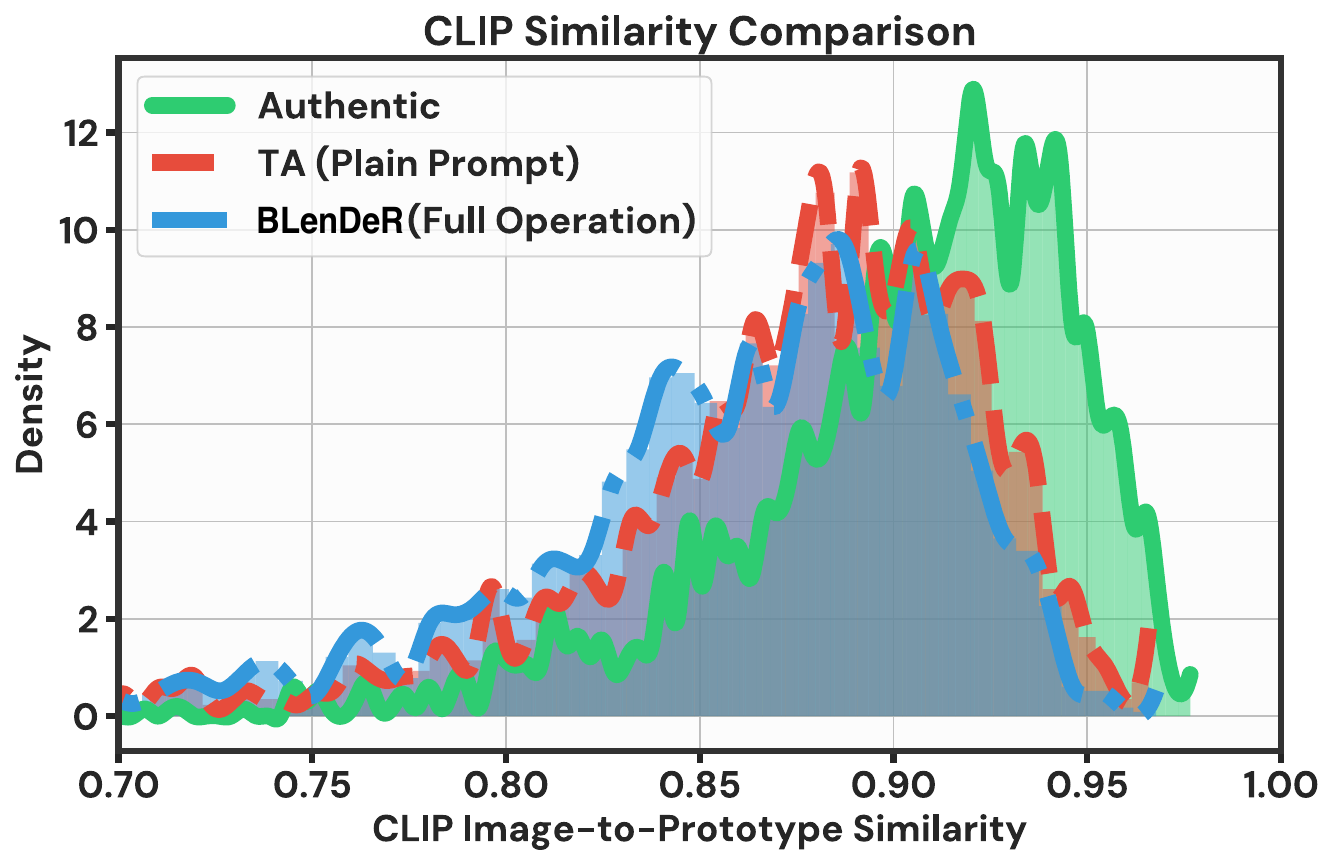}
        \caption{Cars - Background}
        \label{supp:cossim_cars_background}
    \end{subfigure}
    
    \vspace{0.5em}
    
    \begin{subfigure}[h]{0.48\columnwidth}
        \centering
        \includegraphics[width=\textwidth]{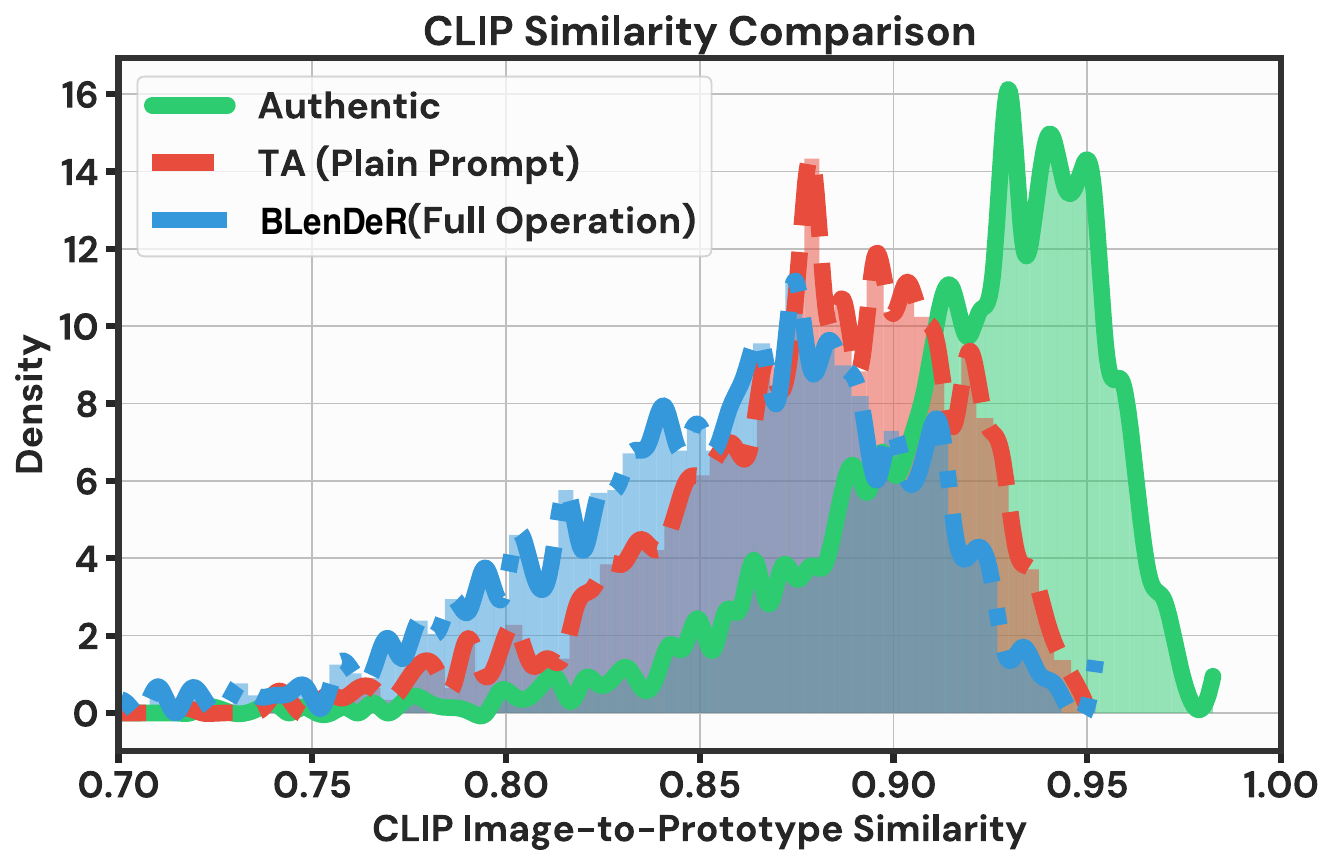}
        \caption{CUB - Pose}
        \label{supp:cossim_cub200_pose}
    \end{subfigure}
    \hfill
    \begin{subfigure}[h]{0.48\columnwidth}
        \centering
        \includegraphics[width=\textwidth]{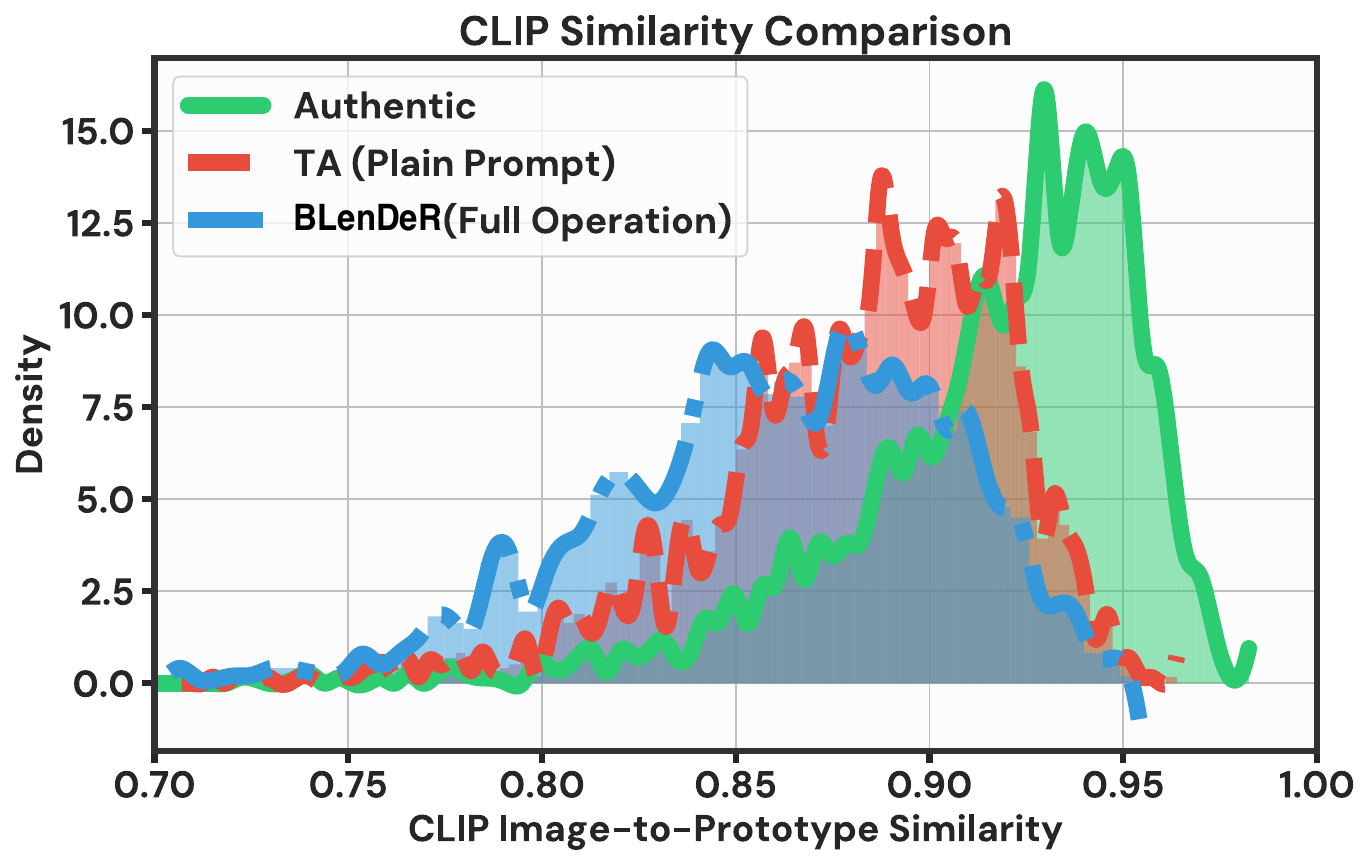}
        \caption{CUB - Background}
        \label{supp:cossim_cub200_background}
    \end{subfigure}
    
    \caption{CLIP similarity distribution comparison between Authentic, plain prompt baseline (\opprompt), and \blendr synthetic images across different datasets and attributes.}
    \label{fig:cossim_combined}
\end{figure}

Across all configurations, \blendr consistently exhibits higher standard deviation compared to plain prompt baselines: +12.8\% for Cars Background, +1.1\% for Cars Pose, +26.2\% for Cub Background, and +19.3\% for Cub Pose (Figures~\ref{supp:cossim_cars_background} and~\ref{supp:cossim_cub200_background}). This increased variance indicates greater visual diversity within each class, approaching the natural variation in authentic data. The lower mean similarity demonstrates that \blendr explores a broader semantic space rather than clustering around a single visual prototype.

Plain prompt baselines exhibit higher mean similarity and lower variance, indicating that standard text-to-image generation produces visually homogeneous images clustered tightly around class prototypes. \blendr's set-based embedding interpolation directly addresses this limitation: union operations broaden the semantic space by combining multiple visual attributes, while intersection operations refine specific attributes while preserving class coherence. The combined effect—lower mean similarity with higher standard deviation (Figures~\ref{supp:cossim_cars_pose} and~\ref{supp:cossim_cub200_pose})—demonstrates that \blendr introduces meaningful visual diversity while maintaining class identity.

Although \blendr's mean similarity is lower than authentic images (expected, since authentic images define the prototype), its standard deviation approaches authentic levels (Figure~\ref{fig:cossim_combined}). For Cars, \blendr Background achieves $\sigma=0.0590$ versus authentic $\sigma=0.0437$; for Cub, $\sigma=0.0472$ versus $\sigma=0.0378$. The similarity distributions in Figures~\ref{supp:cossim_cars_pose} and~\ref{supp:cossim_cub200_pose} further demonstrate that \blendr captures realistic intra-class variation, which is essential for deep metric learning where diverse within-class examples improve generalization.

These results confirm that \blendr addresses a fundamental limitation of standard text-to-image generation: insufficient visual diversity. Training on homogeneous synthetic data can lead to overfitting and reduced discriminative capacity. By introducing controlled diversity through set-based operations, \blendr produces synthetic training data with statistical properties closer to real-world distributions, directly contributing to the improved retrieval performance demonstrated in our main results.

\section{Example Images for different generation types}
\label{supp:example_images_per_ablation}

The main paper showed limited visual examples due to space constraints. This section provides extensive randomly selected examples across background and pose attributes, demonstrating \blendr's success in generating images with challenging attribute descriptions.

\begin{figure}[H]
    \centering
    \small
    \begin{subfigure}[t]{0.15\textwidth}
        \centering
        \includegraphics[width=\textwidth]{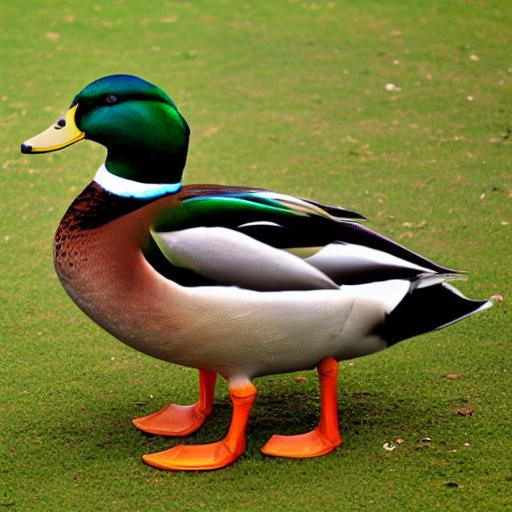}
        \caption{TA}
    \end{subfigure}
    \hfill
    \begin{subfigure}[t]{0.15\textwidth}
        \centering
        \includegraphics[width=\textwidth]{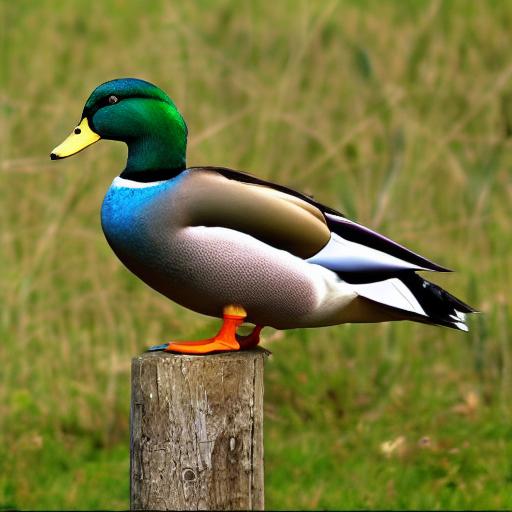}
        \caption{TEI}
    \end{subfigure}
    \hfill
    \begin{subfigure}[t]{0.15\textwidth}
        \centering
        \includegraphics[width=\textwidth]{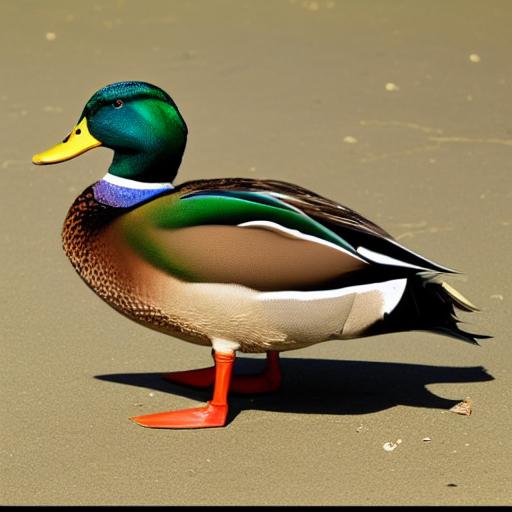}
        \caption{RSO}
    \end{subfigure}
    \hfill
    \begin{subfigure}[t]{0.15\textwidth}
        \centering
        \includegraphics[width=\textwidth]{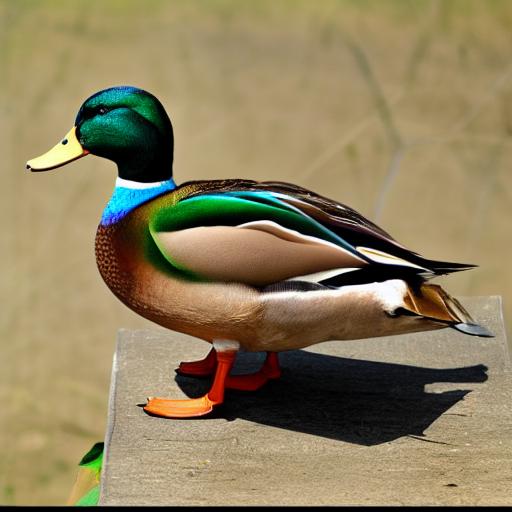}
        \caption{BLenDeR}
    \end{subfigure}
    \\[0.2em]
    {\footnotesize \textbf{Target:} \texttt{The background of the image is a soft, warm beige color, which provides a neutral backdrop that allows the bird to stand out prominently. The lighting in the background is soft and diffused, creating a gentle and serene atmosphere.}}
    \\[0.5em]
    \begin{subfigure}[t]{0.15\textwidth}
        \centering
        \includegraphics[width=\textwidth]{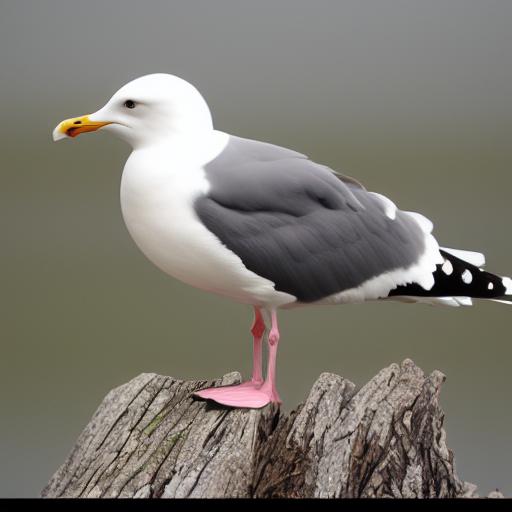}
        \caption*{TA}
    \end{subfigure}
    \hfill
    \begin{subfigure}[t]{0.15\textwidth}
        \centering
        \includegraphics[width=\textwidth]{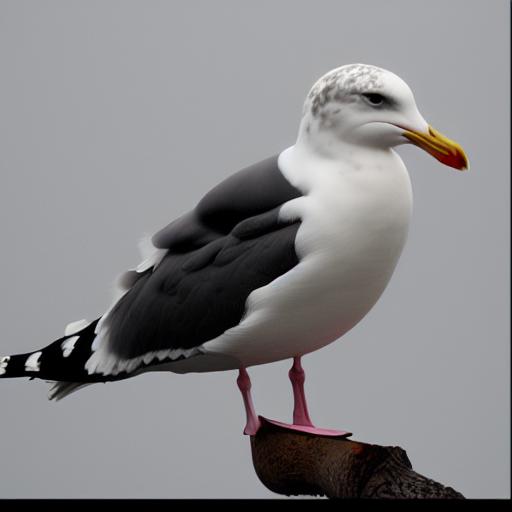}
        \caption*{TEI}
    \end{subfigure}
    \hfill
    \begin{subfigure}[t]{0.15\textwidth}
        \centering
        \includegraphics[width=\textwidth]{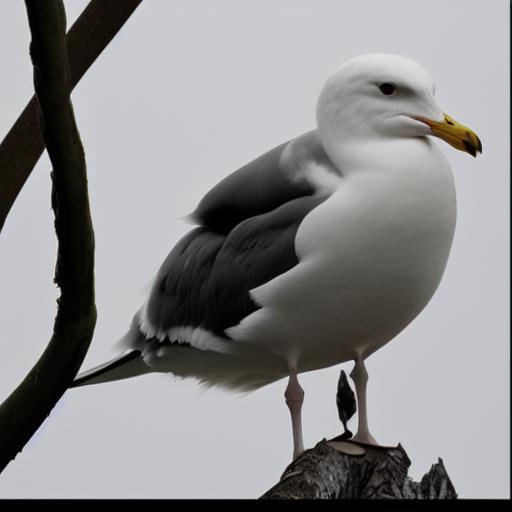}
        \caption*{RSO}
    \end{subfigure}
    \hfill
    \begin{subfigure}[t]{0.15\textwidth}
        \centering
        \includegraphics[width=\textwidth]{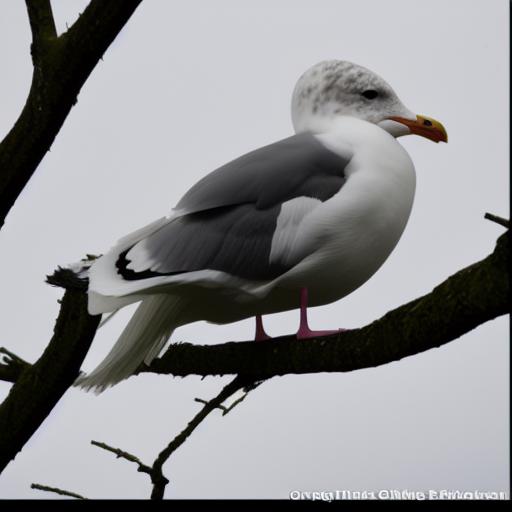}
        \caption*{BLenDeR}
    \end{subfigure}
    \\[0.2em]
    {\footnotesize \textbf{Target:} \texttt{The background of the image is a soft, overcast sky with a muted, diffused light. There are no distinct features or objects in the background, as the focus of the image is on the bird perched on the tree branch.}}
    \\[0.5em]
    \begin{subfigure}[t]{0.15\textwidth}
        \centering
        \includegraphics[width=\textwidth]{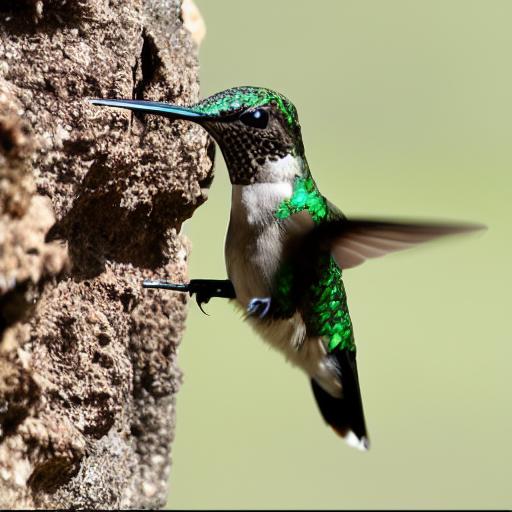}
        \caption*{TA}
    \end{subfigure}
    \hfill
    \begin{subfigure}[t]{0.15\textwidth}
        \centering
        \includegraphics[width=\textwidth]{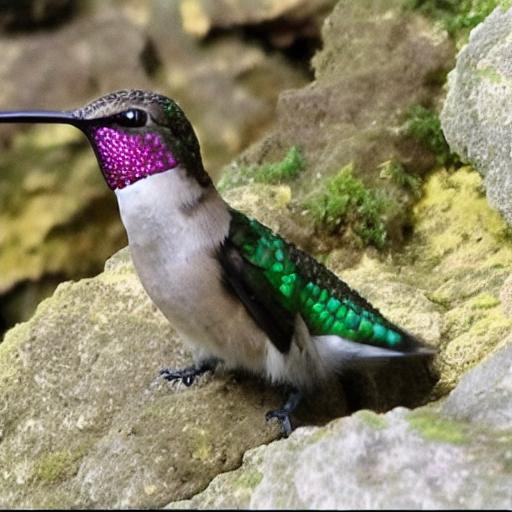}
        \caption*{TEI}
    \end{subfigure}
    \hfill
    \begin{subfigure}[t]{0.15\textwidth}
        \centering
        \includegraphics[width=\textwidth]{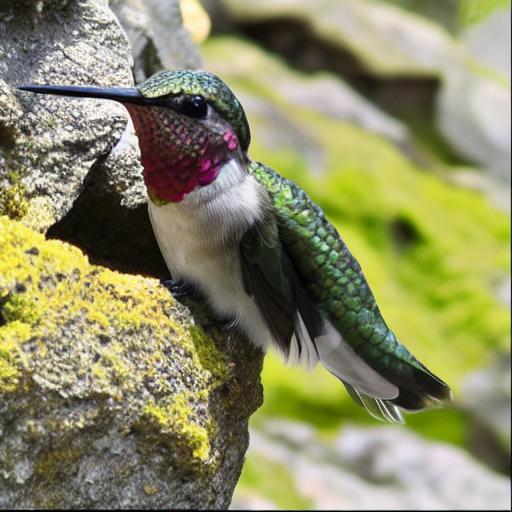}
        \caption*{RSO}
    \end{subfigure}
    \hfill
    \begin{subfigure}[t]{0.15\textwidth}
        \centering
        \includegraphics[width=\textwidth]{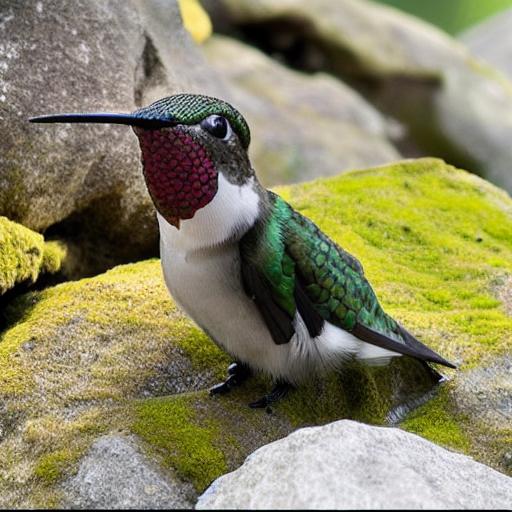}
        \caption*{BLenDeR}
    \end{subfigure}
    \\[0.2em]
    {\footnotesize \textbf{Target:} \texttt{The background of the image features a rocky cliff face with a variety of textures and colors. The rock appears to be weathered with patches of green moss and lichen, indicating some level of moisture or humidity in the environment.}}
    \\[0.5em]
    \begin{subfigure}[t]{0.15\textwidth}
        \centering
        \includegraphics[width=\textwidth]{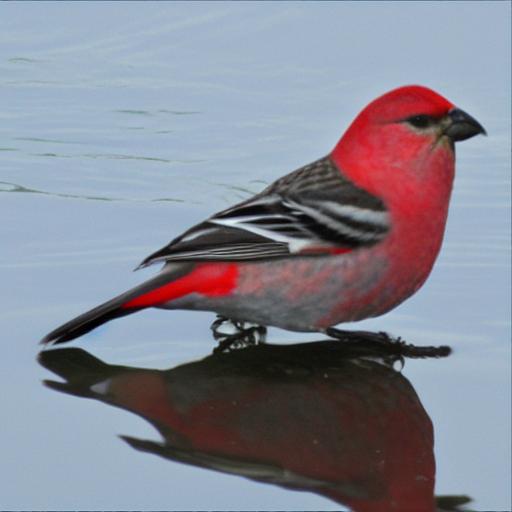}
        \caption*{TA}
    \end{subfigure}
    \hfill
    \begin{subfigure}[t]{0.15\textwidth}
        \centering
        \includegraphics[width=\textwidth]{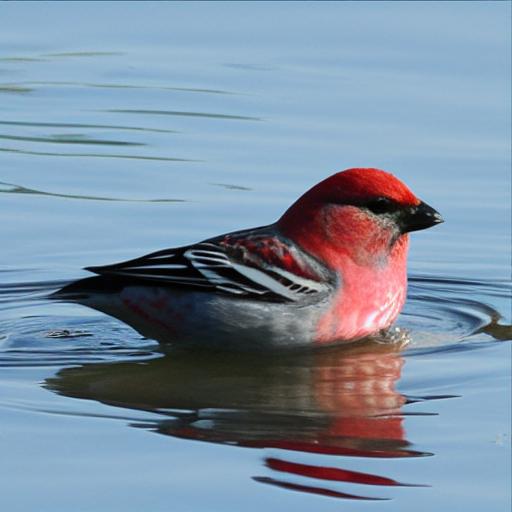}
        \caption*{TEI}
    \end{subfigure}
    \hfill
    \begin{subfigure}[t]{0.15\textwidth}
        \centering
        \includegraphics[width=\textwidth]{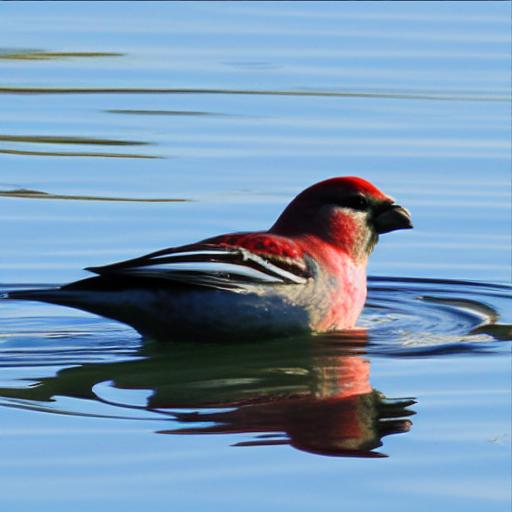}
        \caption*{RSO}
    \end{subfigure}
    \hfill
    \begin{subfigure}[t]{0.15\textwidth}
        \centering
        \includegraphics[width=\textwidth]{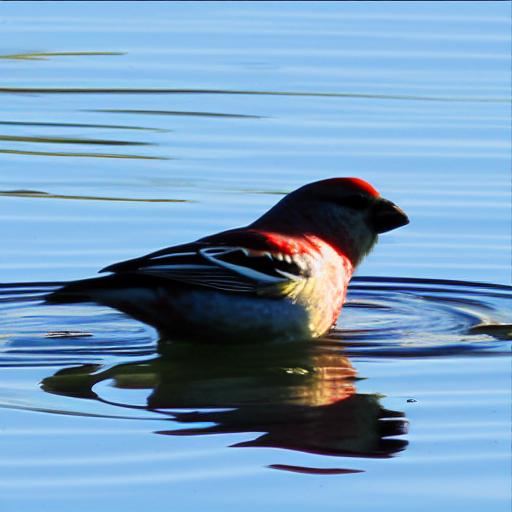}
        \caption*{BLenDeR}
    \end{subfigure}
    \\[0.2em]
    {\footnotesize \textbf{Target:} \texttt{The background of the image features a body of water that appears calm with gentle ripples. The water is a deep blue color, reflecting the light in a way that suggests it might be a sunny day.}}
    \\[0.5em]
    \begin{subfigure}[t]{0.15\textwidth}
        \centering
        \includegraphics[width=\textwidth]{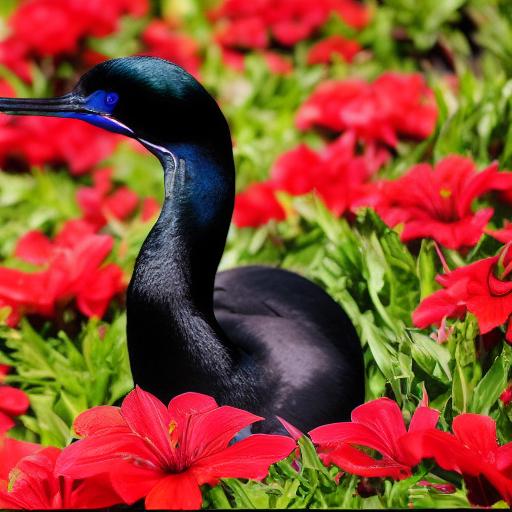}
        \caption*{TA}
    \end{subfigure}
    \hfill
    \begin{subfigure}[t]{0.15\textwidth}
        \centering
        \includegraphics[width=\textwidth]{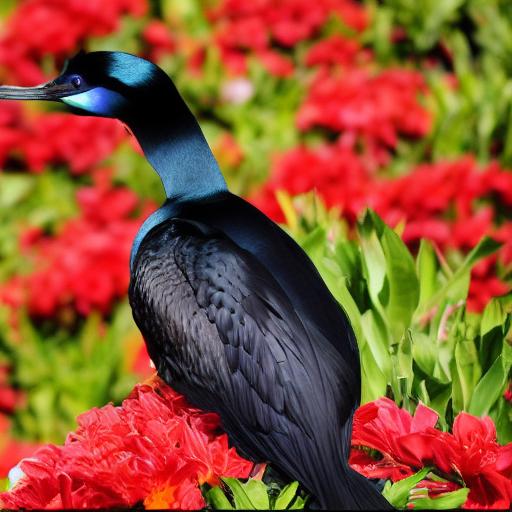}
        \caption*{TEI}
    \end{subfigure}
    \hfill
    \begin{subfigure}[t]{0.15\textwidth}
        \centering
        \includegraphics[width=\textwidth]{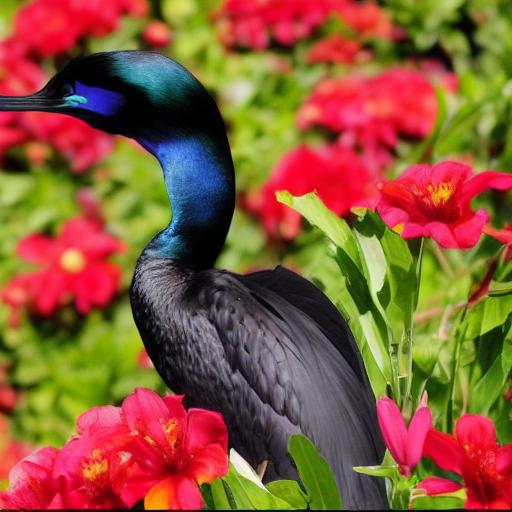}
        \caption*{RSO}
    \end{subfigure}
    \hfill
    \begin{subfigure}[t]{0.15\textwidth}
        \centering
        \includegraphics[width=\textwidth]{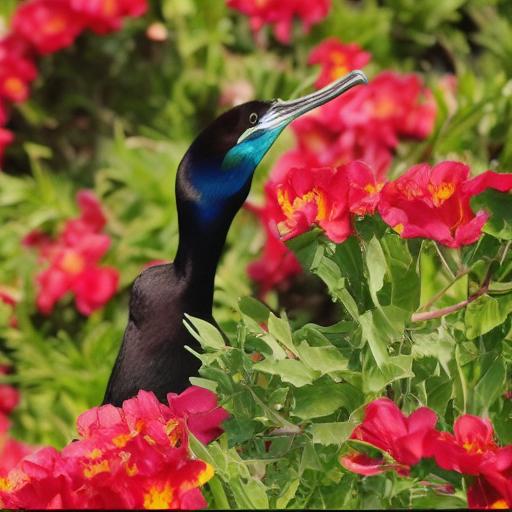}
        \caption*{BLenDeR}
    \end{subfigure}
    \\[0.2em]
    {\footnotesize \textbf{Target:} \texttt{The background of the image features a vibrant display of red flowers. These flowers are in full bloom, with their petals spread wide, revealing the intricate details of their stamens and pistils.}}
    \caption{Randomly selected samples displaying generation approaches for the background attribute on images of birds. Each row shows a different sample with four generation strategies: \opprompt (target anchor prompt only), \opembmix (text embedding interpolation), \opblendrop (residual set operations only), and \opblendr (full method with both TEI and RSO).}
    \label{supp:fig:overview_background_4}
\end{figure}

\begin{figure}[H]
    \centering
    \small
    \begin{subfigure}[t]{0.15\textwidth}
        \centering
        \includegraphics[width=\textwidth]{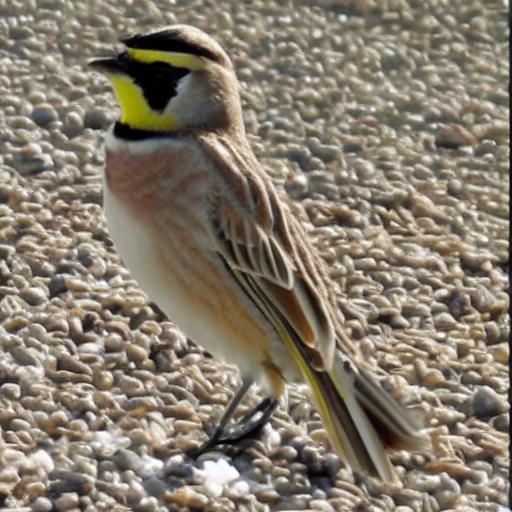}
        \caption{TA}
    \end{subfigure}
    \hfill
    \begin{subfigure}[t]{0.15\textwidth}
        \centering
        \includegraphics[width=\textwidth]{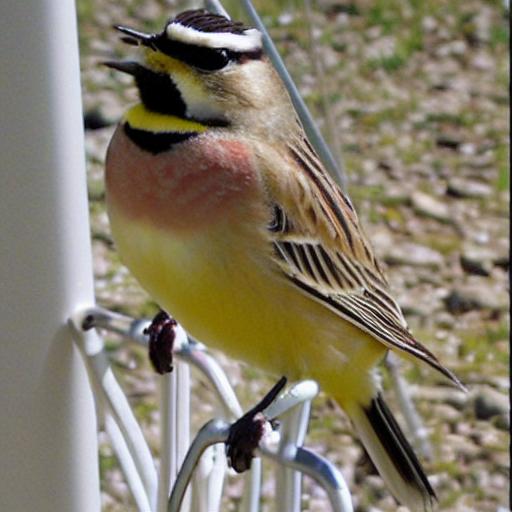}
        \caption{TEI}
    \end{subfigure}
    \hfill
    \begin{subfigure}[t]{0.15\textwidth}
        \centering
        \includegraphics[width=\textwidth]{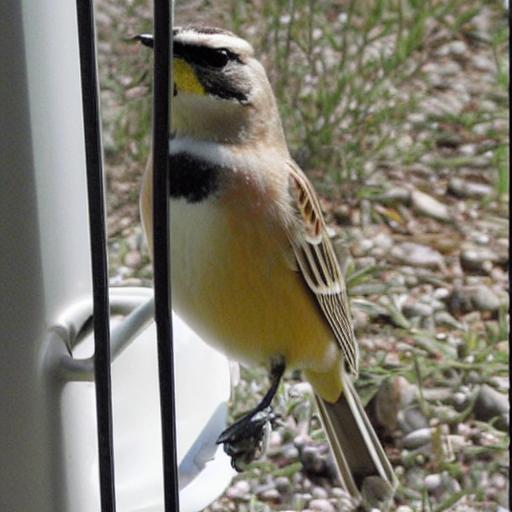}
        \caption{RSO}
    \end{subfigure}
    \hfill
    \begin{subfigure}[t]{0.15\textwidth}
        \centering
        \includegraphics[width=\textwidth]{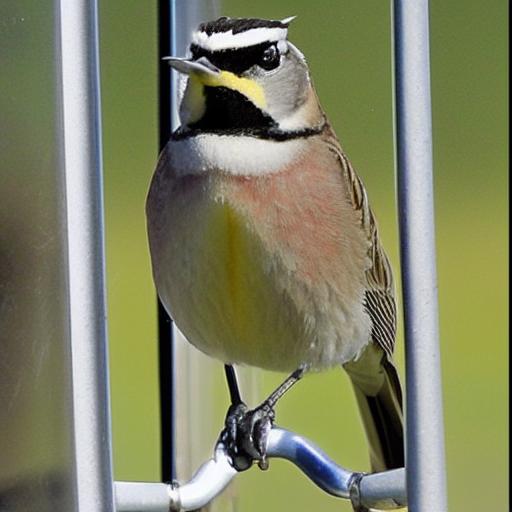}
        \caption{BLenDeR}
    \end{subfigure}
    \\[0.2em]
    {\footnotesize \textbf{Target:} \texttt{The bird is perched on a metal structure, which appears to be part of a bird feeder or a similar type of bird-friendly equipment. The bird is facing to the right, with its head turned slightly towards the camera.}}
    \\[0.5em]
    \begin{subfigure}[t]{0.15\textwidth}
        \centering
        \includegraphics[width=\textwidth]{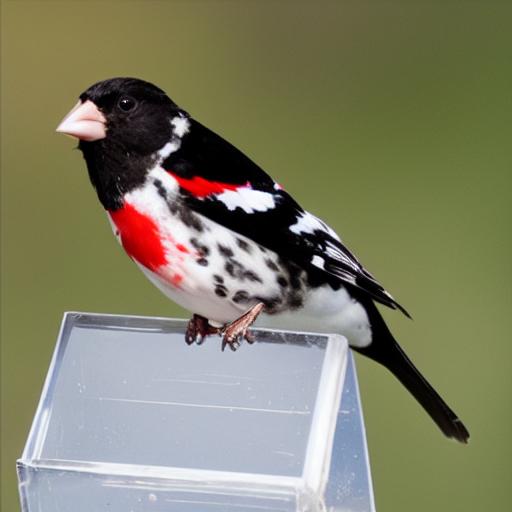}
        \caption*{TA}
    \end{subfigure}
    \hfill
    \begin{subfigure}[t]{0.15\textwidth}
        \centering
        \includegraphics[width=\textwidth]{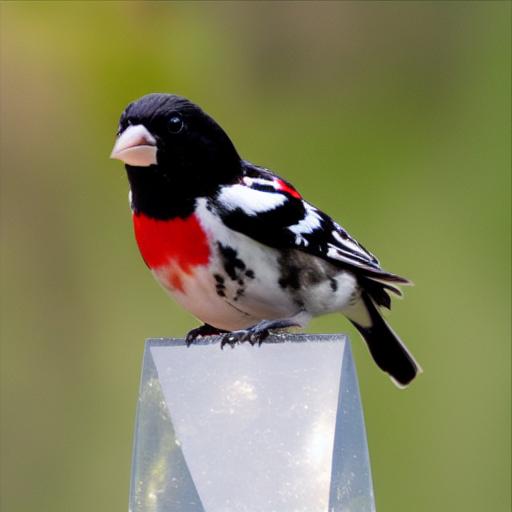}
        \caption*{TEI}
    \end{subfigure}
    \hfill
    \begin{subfigure}[t]{0.15\textwidth}
        \centering
        \includegraphics[width=\textwidth]{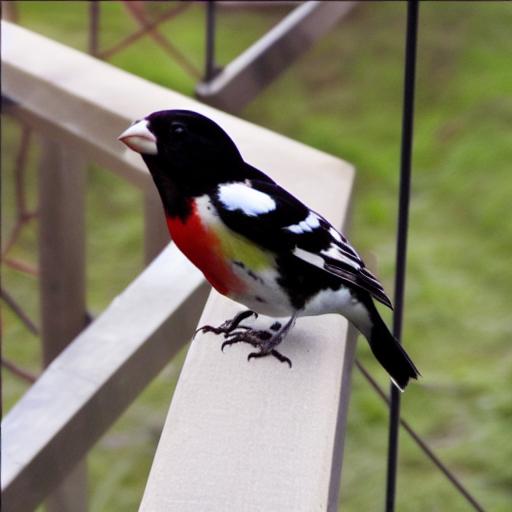}
        \caption*{RSO}
    \end{subfigure}
    \hfill
    \begin{subfigure}[t]{0.15\textwidth}
        \centering
        \includegraphics[width=\textwidth]{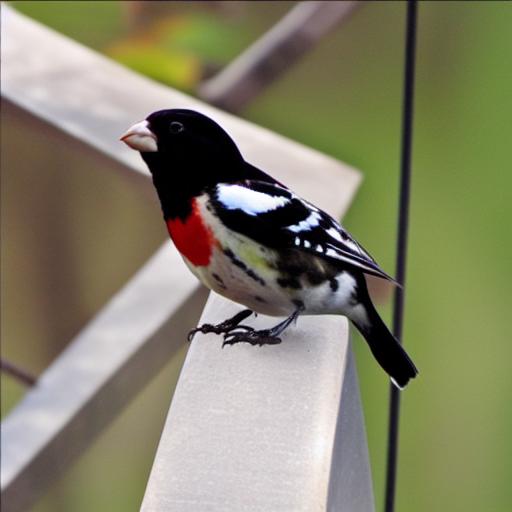}
        \caption*{BLenDeR}
    \end{subfigure}
    \\[0.2em]
    {\footnotesize \textbf{Target:} \texttt{The bird is perched on a triangular structure, which appears to be a piece of glass or a similar transparent material.}}
    \\[0.5em]
    \begin{subfigure}[t]{0.15\textwidth}
        \centering
        \includegraphics[width=\textwidth]{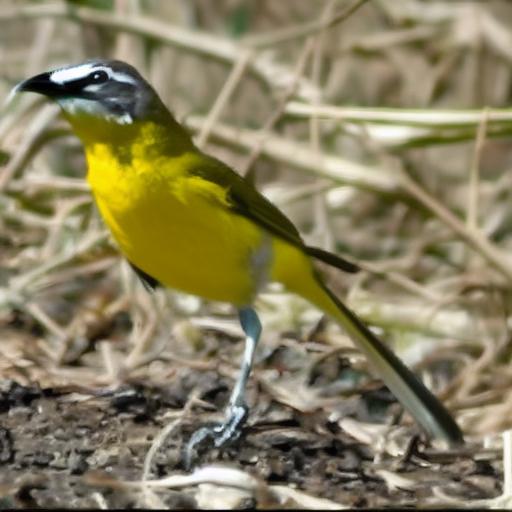}
        \caption*{TA}
    \end{subfigure}
    \hfill
    \begin{subfigure}[t]{0.15\textwidth}
        \centering
        \includegraphics[width=\textwidth]{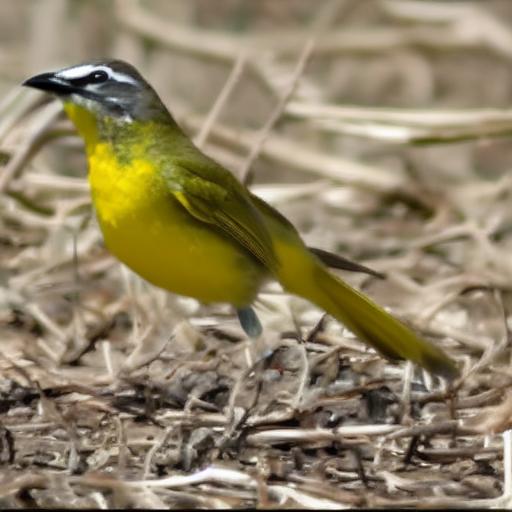}
        \caption*{TEI}
    \end{subfigure}
    \hfill
    \begin{subfigure}[t]{0.15\textwidth}
        \centering
        \includegraphics[width=\textwidth]{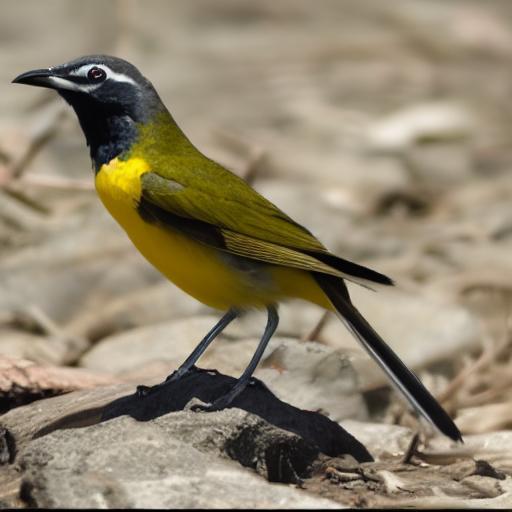}
        \caption*{RSO}
    \end{subfigure}
    \hfill
    \begin{subfigure}[t]{0.15\textwidth}
        \centering
        \includegraphics[width=\textwidth]{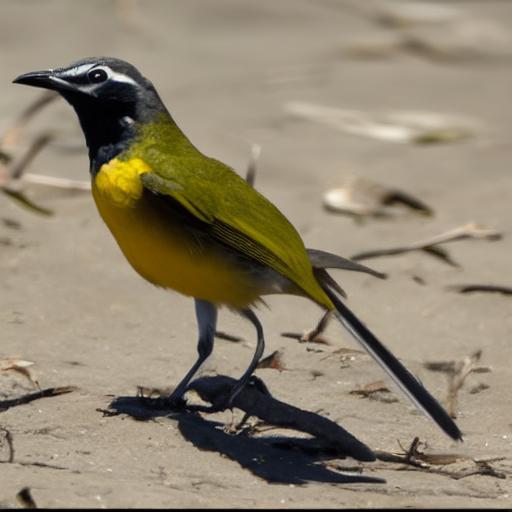}
        \caption*{BLenDeR}
    \end{subfigure}
    \\[0.2em]
    {\footnotesize \textbf{Target:} \texttt{The bird is captured in a dynamic pose, with its head tilted upwards and its beak wide open as if it is in the midst of a call or song.}}
    \\[0.5em]
    \begin{subfigure}[t]{0.15\textwidth}
        \centering
        \includegraphics[width=\textwidth]{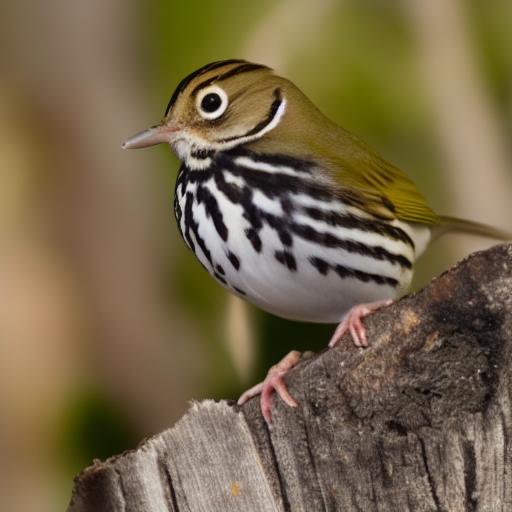}
        \caption*{TA}
    \end{subfigure}
    \hfill
    \begin{subfigure}[t]{0.15\textwidth}
        \centering
        \includegraphics[width=\textwidth]{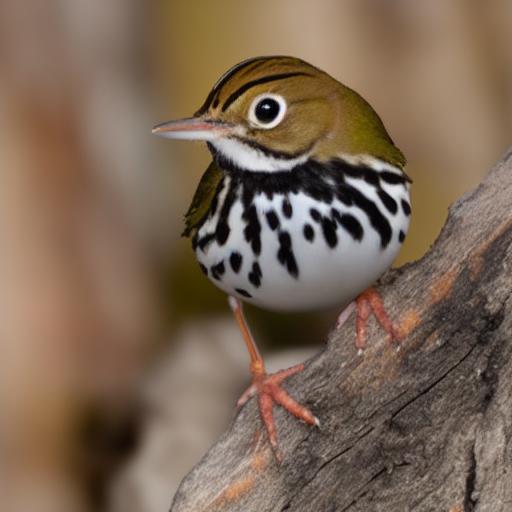}
        \caption*{TEI}
    \end{subfigure}
    \hfill
    \begin{subfigure}[t]{0.15\textwidth}
        \centering
        \includegraphics[width=\textwidth]{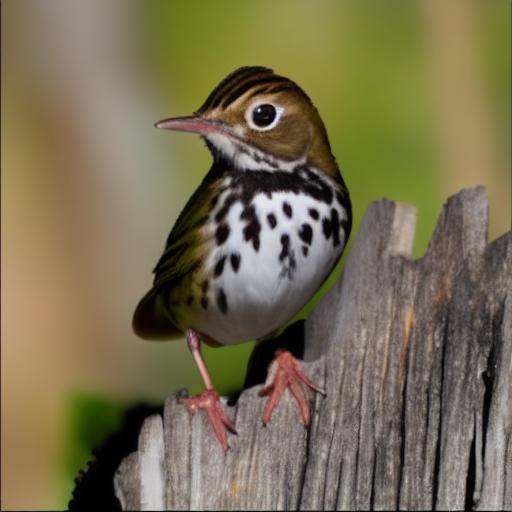}
        \caption*{RSO}
    \end{subfigure}
    \hfill
    \begin{subfigure}[t]{0.15\textwidth}
        \centering
        \includegraphics[width=\textwidth]{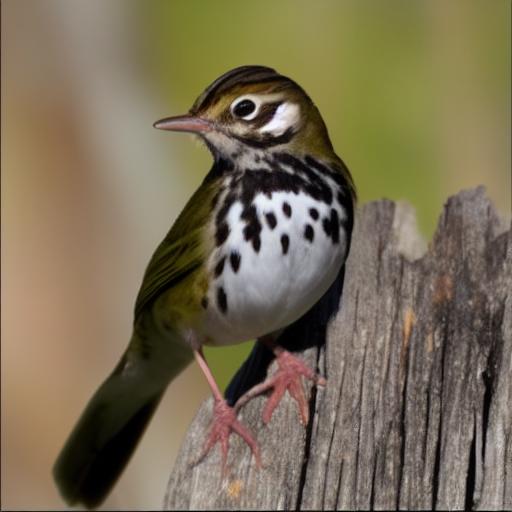}
        \caption*{BLenDeR}
    \end{subfigure}
    \\[0.2em]
    {\footnotesize \textbf{Target:} \texttt{The bird is perched on a wooden fence post. It is facing to the left with its head turned slightly towards the camera.}}
    \\[0.5em]
    \begin{subfigure}[t]{0.15\textwidth}
        \centering
        \includegraphics[width=\textwidth]{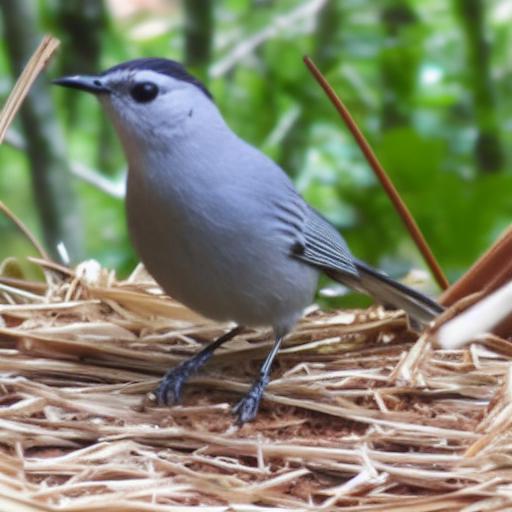}
        \caption*{TA}
    \end{subfigure}
    \hfill
    \begin{subfigure}[t]{0.15\textwidth}
        \centering
        \includegraphics[width=\textwidth]{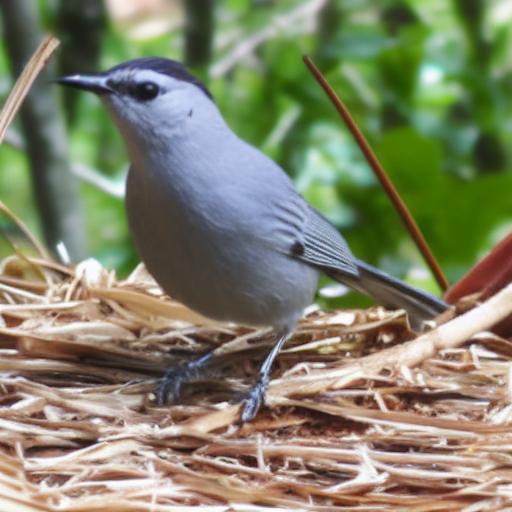}
        \caption*{TEI}
    \end{subfigure}
    \hfill
    \begin{subfigure}[t]{0.15\textwidth}
        \centering
        \includegraphics[width=\textwidth]{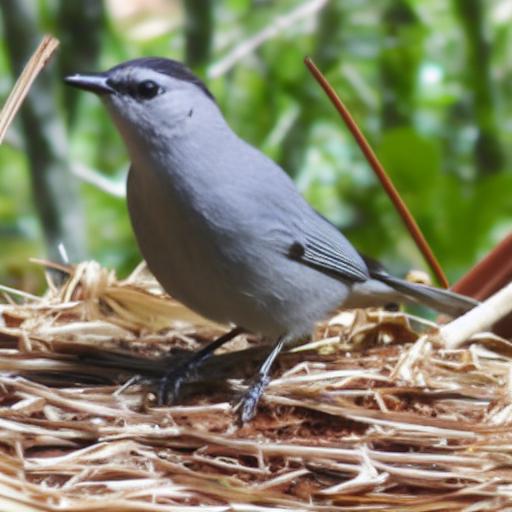}
        \caption*{RSO}
    \end{subfigure}
    \hfill
    \begin{subfigure}[t]{0.15\textwidth}
        \centering
        \includegraphics[width=\textwidth]{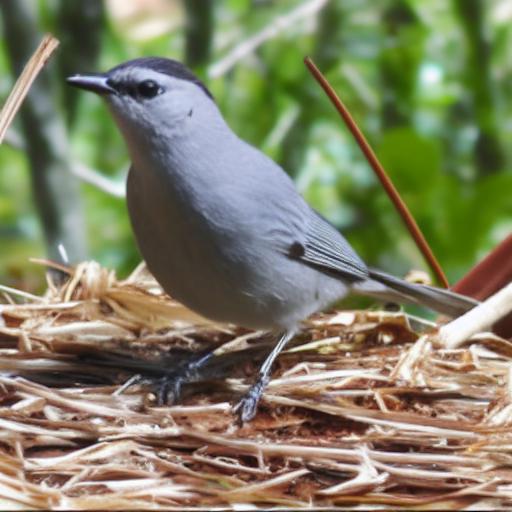}
        \caption*{BLenDeR}
    \end{subfigure}
    \\[0.2em]
    {\footnotesize \textbf{Target:} \texttt{The bird is perched upright on a bed of straw and twigs. Its head is slightly tilted to the left, and it appears to be looking downwards towards the ground.}}
    \caption{Randomly selected samples displaying generation approaches for the pose attribute on images of birds. Each row shows a different sample with four generation strategies: \opprompt (target anchor prompt only), \opembmix (text embedding interpolation), \opblendrop (residual set operations only), and \opblendr (full method with both \opembmix and \opblendrop).}
    \label{supp:fig:overview_pose_8}
\end{figure}

Figure ~\ref{supp:fig:overview_background_4} and Figure~\ref{supp:fig:overview_pose_8} present visual comparisons of the four generation strategies across background and pose attributes, respectively, on images of birds. Each figure displays five randomly selected samples from different bird classes, demonstrating the generalization of our approach across diverse species and attributes. Importantly, these samples were selected randomly and were not hand-picked or cherry-picked to showcase favorable results. The visualizations consistently demonstrate that when the baseline target anchor (\opprompt) approach struggles to achieve strong prompt adherence for challenging attribute descriptions, \blendr substantially improves alignment with the target attribute while maintaining class fidelity. This improvement is particularly evident in cases where the target attribute requires specific visual details that are difficult for the base model to generate from the prompt alone, such as complex backgrounds (e.g., snowy landscapes, ocean waves) or precise poses (e.g., birds in flight, specific perching positions).

\section{Example Images of \blendr Training Dataset}
\label{supp:example_images_blendr_training_datasets}

While Section~\ref{supp:example_images_per_ablation} compared generation strategies, this section shows random samples from the actual synthetic training datasets for bird images with background and pose attributes to assess overall generation quality and diversity.

\begin{figure}[H]
    \centering

    \begin{subfigure}[h]{\textwidth}
        \centering
        \includegraphics[width=0.12\textwidth]{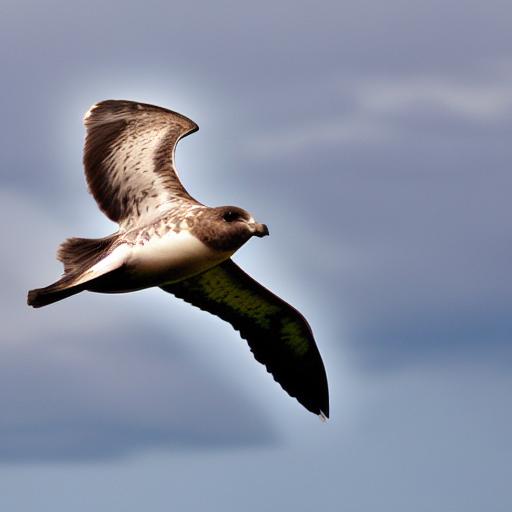}\hspace{0.5mm}%
        \includegraphics[width=0.12\textwidth]{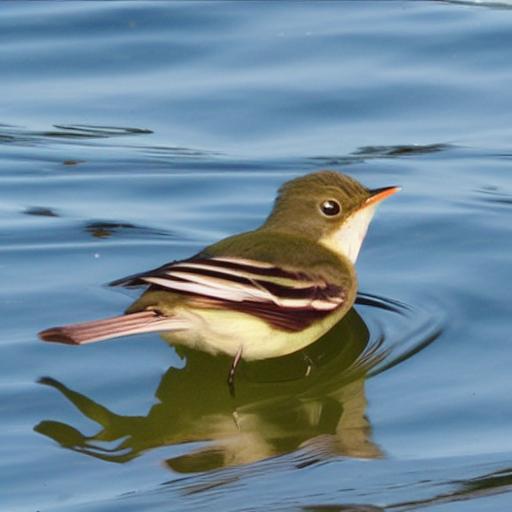}\hspace{0.5mm}%
        \includegraphics[width=0.12\textwidth]{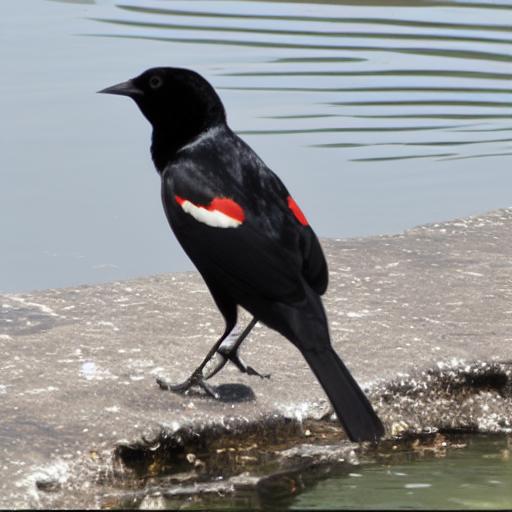}\hspace{0.5mm}%
        \includegraphics[width=0.12\textwidth]{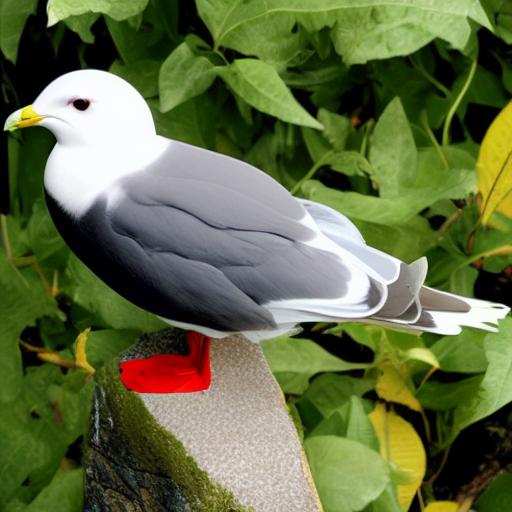}\hspace{0.5mm}%
        \includegraphics[width=0.12\textwidth]{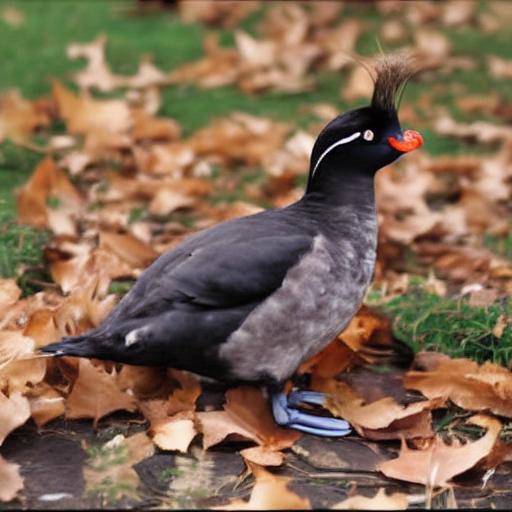}\hspace{0.5mm}%
        \includegraphics[width=0.12\textwidth]{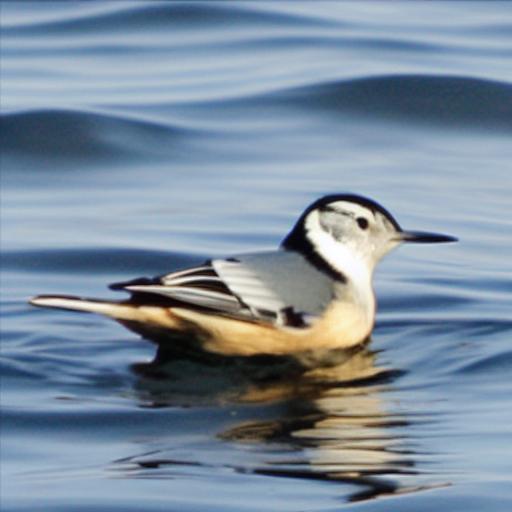}\\[1mm]
        \includegraphics[width=0.12\textwidth]{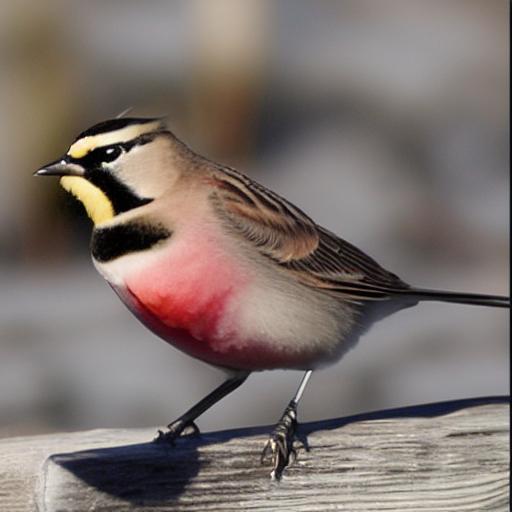}\hspace{0.5mm}%
        \includegraphics[width=0.12\textwidth]{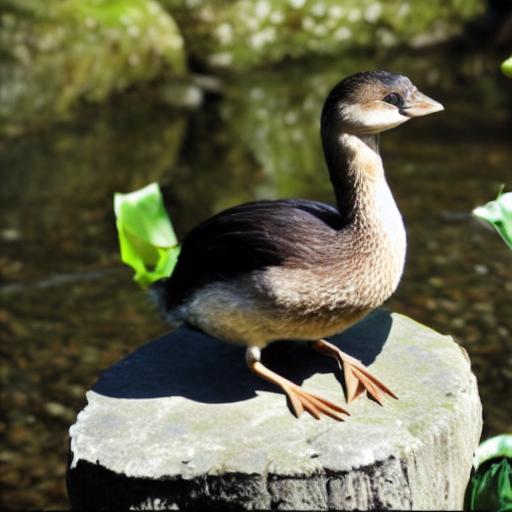}\hspace{0.5mm}%
        \includegraphics[width=0.12\textwidth]{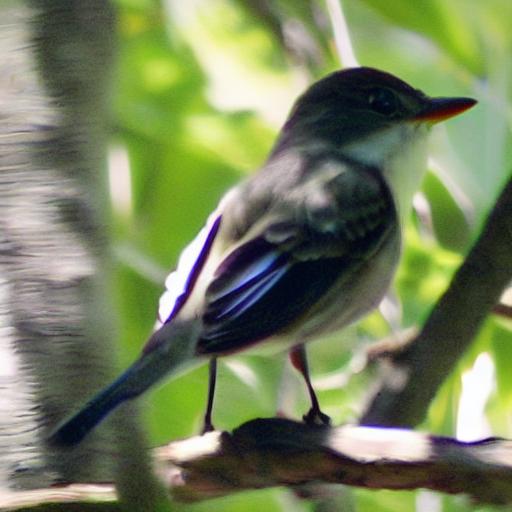}\hspace{0.5mm}%
        \includegraphics[width=0.12\textwidth]{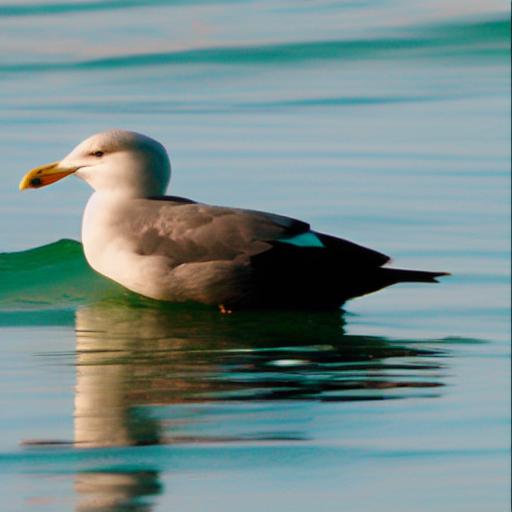}\hspace{0.5mm}%
        \includegraphics[width=0.12\textwidth]{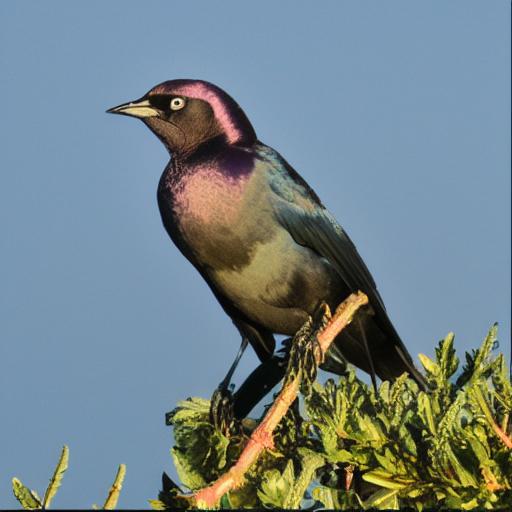}\hspace{0.5mm}%
        \includegraphics[width=0.12\textwidth]{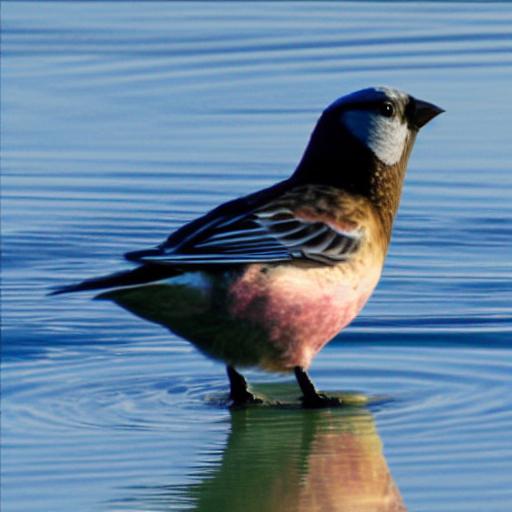}\\[1mm]
        \includegraphics[width=0.12\textwidth]{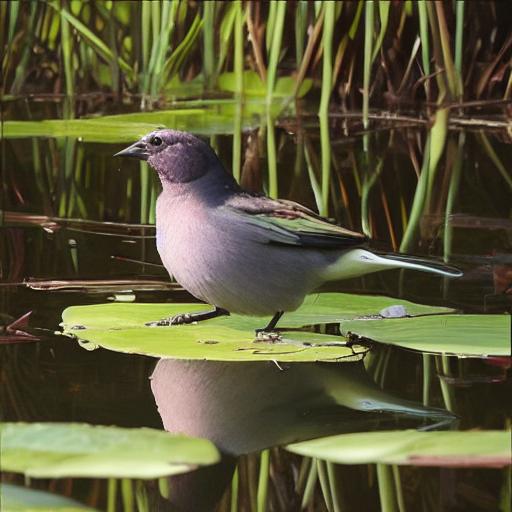}\hspace{0.5mm}%
        \includegraphics[width=0.12\textwidth]{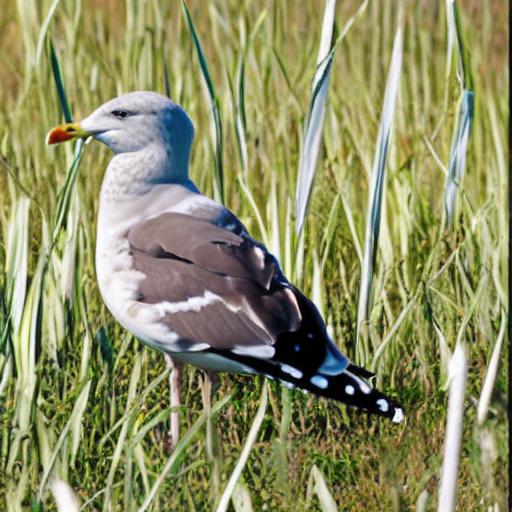}\hspace{0.5mm}%
        \includegraphics[width=0.12\textwidth]{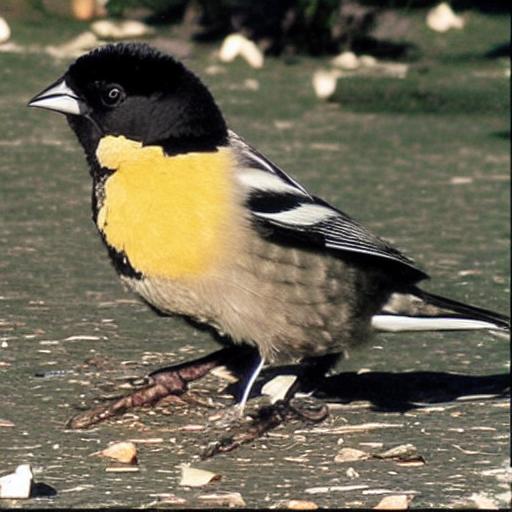}\hspace{0.5mm}%
        \includegraphics[width=0.12\textwidth]{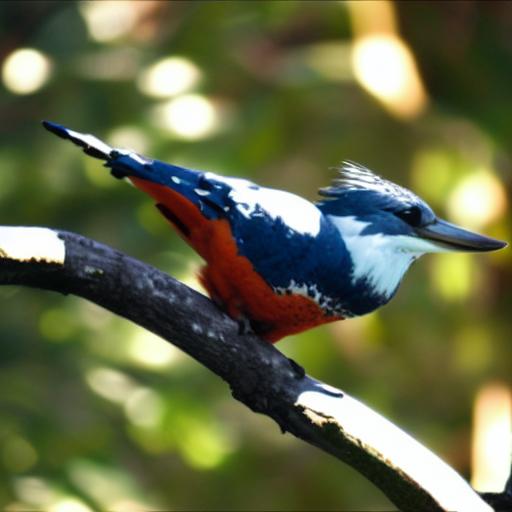}\hspace{0.5mm}%
        \includegraphics[width=0.12\textwidth]{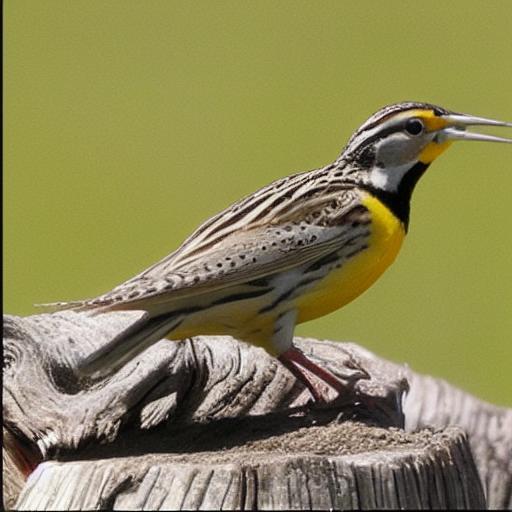}\hspace{0.5mm}%
        \includegraphics[width=0.12\textwidth]{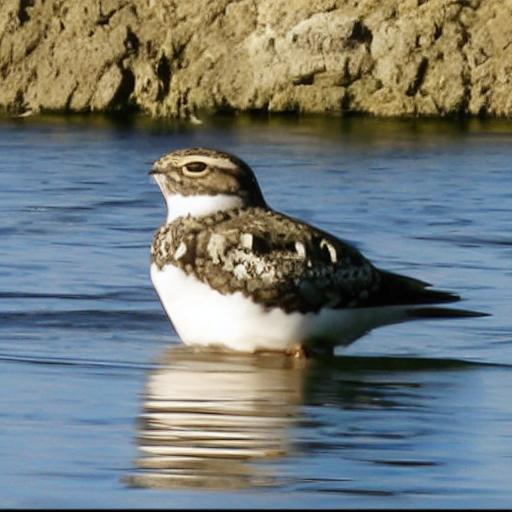}
        \caption{Birds - Background}
        \label{supp:fig:samples_cub200_background}
    \end{subfigure}

    \vspace{3mm}

    \begin{subfigure}[h]{\textwidth}
        \centering
        \includegraphics[width=0.12\textwidth]{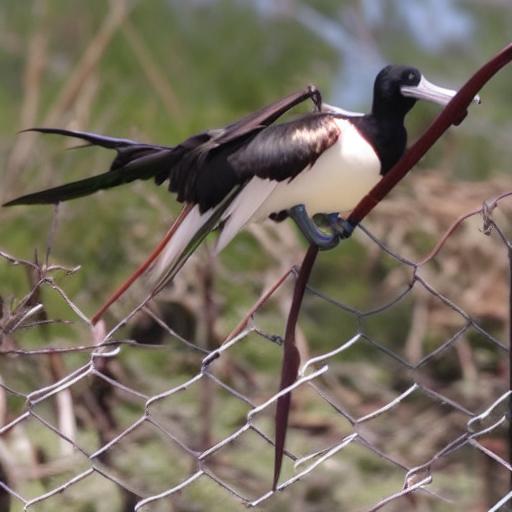}\hspace{0.5mm}%
        \includegraphics[width=0.12\textwidth]{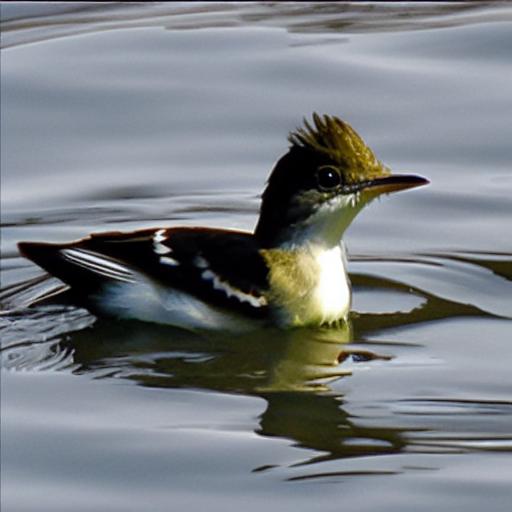}\hspace{0.5mm}%
        \includegraphics[width=0.12\textwidth]{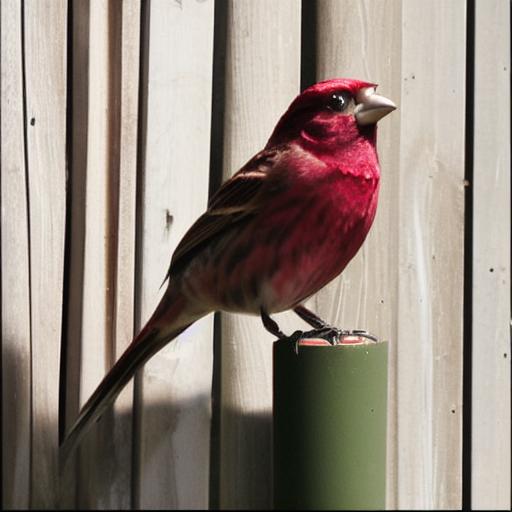}\hspace{0.5mm}%
        \includegraphics[width=0.12\textwidth]{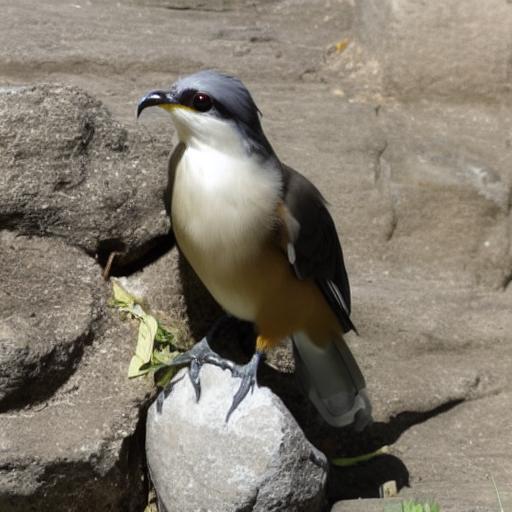}\hspace{0.5mm}%
        \includegraphics[width=0.12\textwidth]{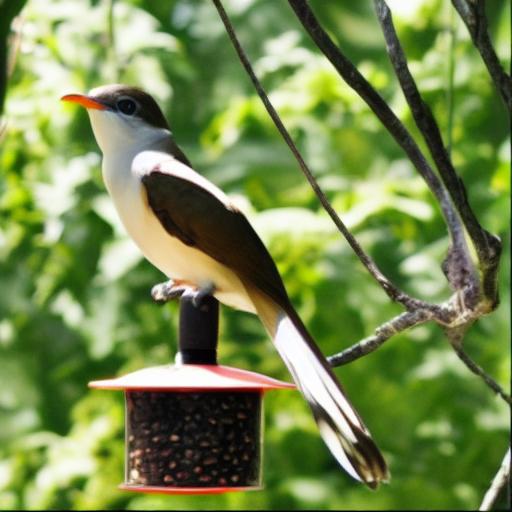}\hspace{0.5mm}%
        \includegraphics[width=0.12\textwidth]{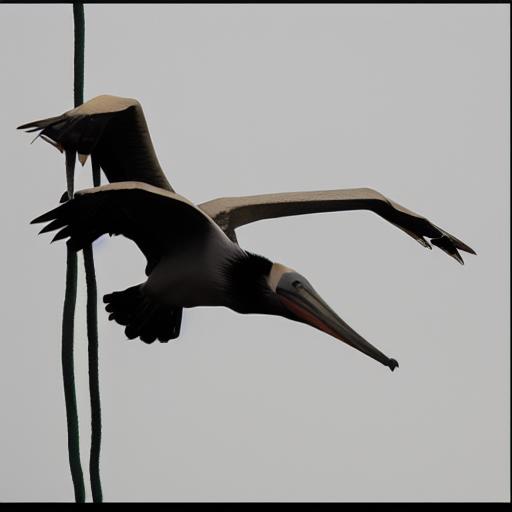}\\[1mm]
        \includegraphics[width=0.12\textwidth]{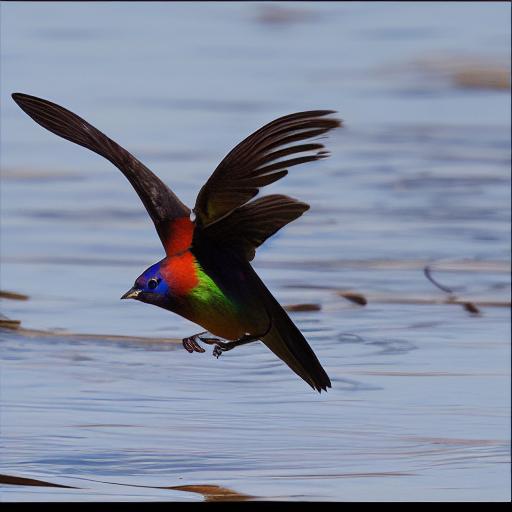}\hspace{0.5mm}%
        \includegraphics[width=0.12\textwidth]{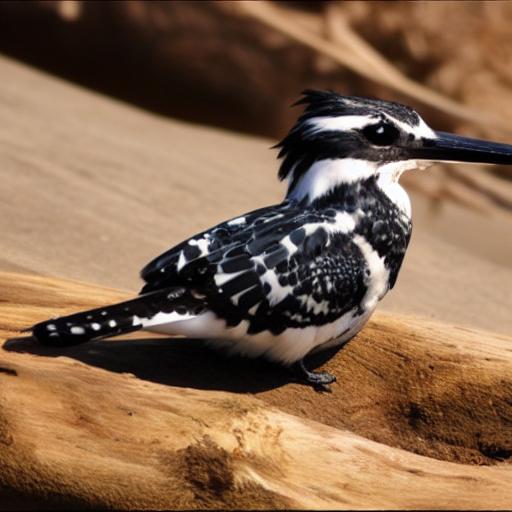}\hspace{0.5mm}%
        \includegraphics[width=0.12\textwidth]{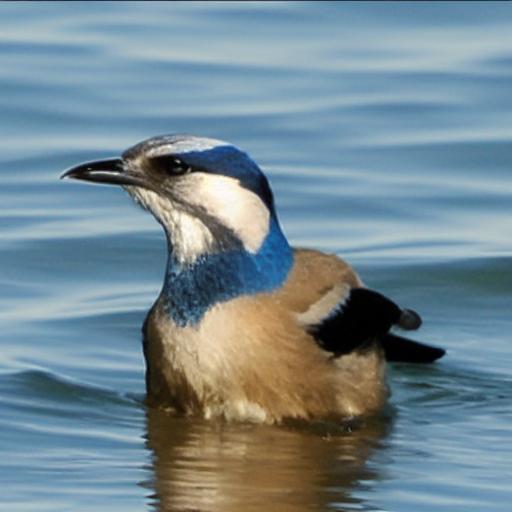}\hspace{0.5mm}%
        \includegraphics[width=0.12\textwidth]{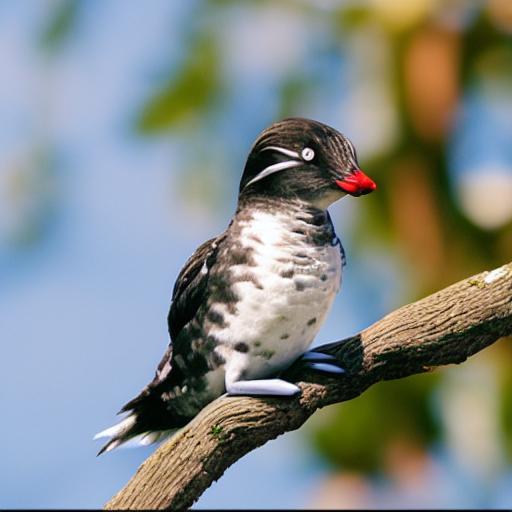}\hspace{0.5mm}%
        \includegraphics[width=0.12\textwidth]{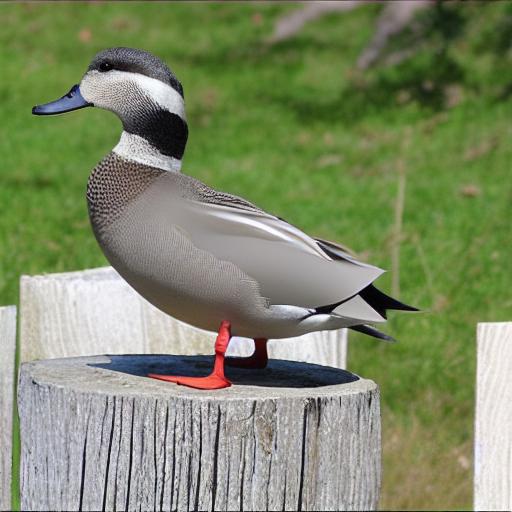}\hspace{0.5mm}%
        \includegraphics[width=0.12\textwidth]{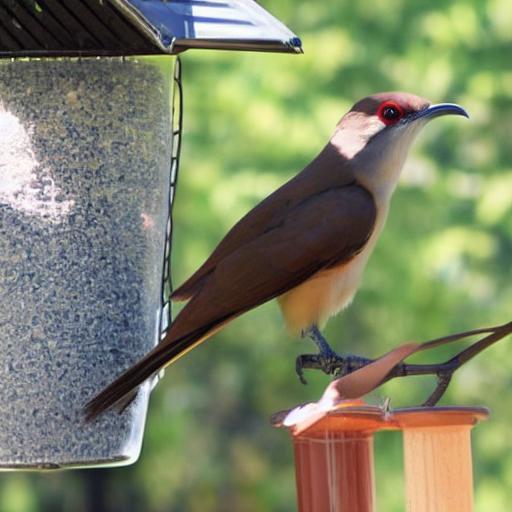}\\[1mm]
        \includegraphics[width=0.12\textwidth]{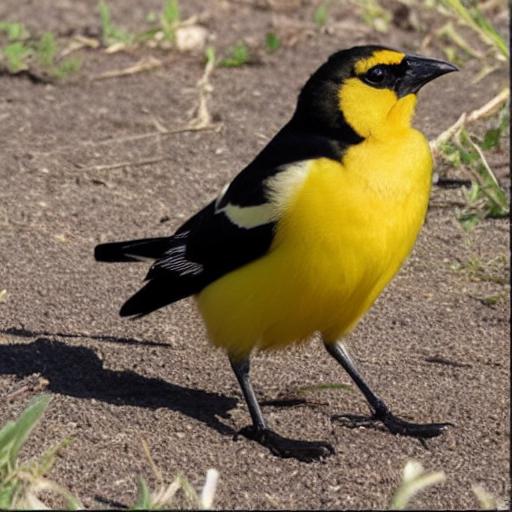}\hspace{0.5mm}%
        \includegraphics[width=0.12\textwidth]{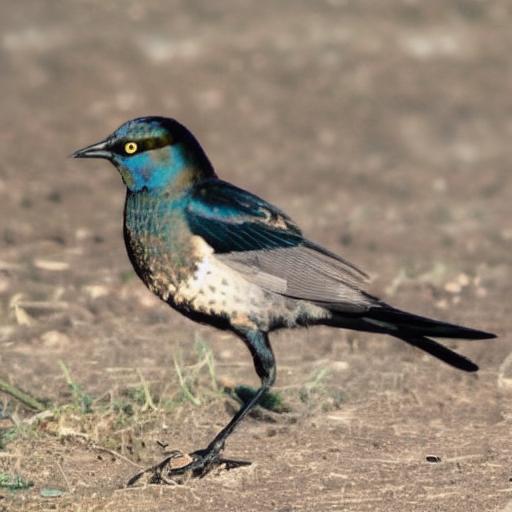}\hspace{0.5mm}%
        \includegraphics[width=0.12\textwidth]{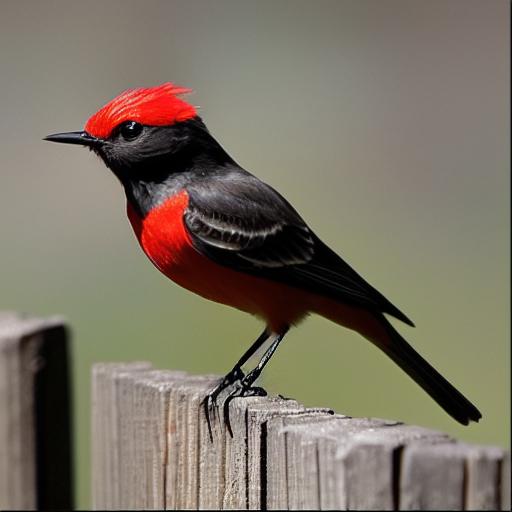}\hspace{0.5mm}%
        \includegraphics[width=0.12\textwidth]{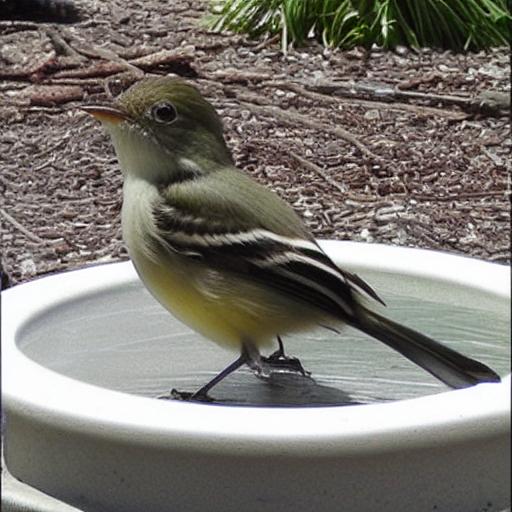}\hspace{0.5mm}%
        \includegraphics[width=0.12\textwidth]{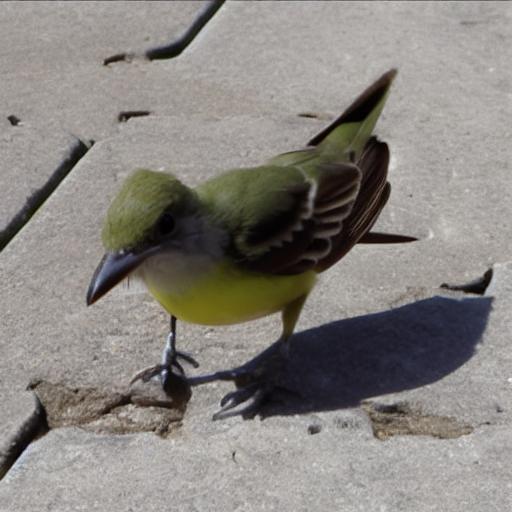}\hspace{0.5mm}%
        \includegraphics[width=0.12\textwidth]{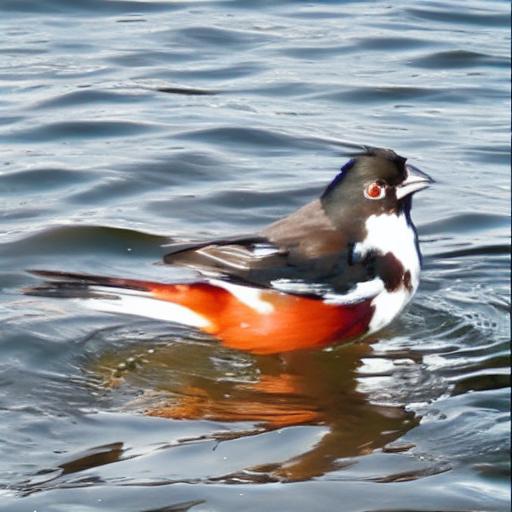}
        \caption{Birds - Pose}
        \label{supp:fig:samples_cub200_pose}
    \end{subfigure}

    \caption{Random sample images from \blendr synthetic datasets. Each row shows 6 images from different classes, from (a) images of birds  with background attribute manipulation, (b) images of birds  with pose attribute manipulation.}
    \label{supp:fig:sample_images_overview}
\end{figure}

Figure~\ref{supp:fig:sample_images_overview} displays randomly selected bird images from background (Fig.~\ref{supp:fig:samples_cub200_background}) and pose (Fig.~\ref{supp:fig:samples_cub200_pose}) target attributes

\end{document}